\documentclass[10pt,a4paper]{article}

\usepackage{amsmath,amsfonts,amssymb,amsthm,url}
\newcommand{\Bem}[1]{}
\newcommand{\NSTUFF}[1]{{#1}\/}
\newcommand{\figaddr}[1]{#1}

\newtheorem{theorem}{Theorem}
\newtheorem{definition}[theorem]{Definition}
\newtheorem{ax}[theorem]{Property}

\usepackage{graphicx}

\begin{document}

\title{Towards Continuous Consistency Axiom}

\author{  Mieczys{\l}aw A. K{\l}opotek
   \\ Institute of Computer Science\\
       ul. Jana Kazimierza 5\\
       01-248 Warszawa, Poland
     \and
        Robert A. K{\l}opotek
        \\Cardinal Stefan  Wyszy{\'n}ski University\\
       Faculty of Mathematics and Natural Sciences\\School of Exact Sciences\\
       ul. Dewajtis 5, 01-815 Warszawa, Poland
        }

%
%


\maketitle

\begin{abstract}
Development of new algorithms in the area of machine learning, especially clustering, comparative studies of such algorithms as well as testing according to software engineering principles requires availability of labeled data sets. While standard benchmarks are made available, a broader range of such data sets is necessary in order to avoid the problem of overfitting. 
In this context, theoretical works on axiomatization of clustering algorithms, especially axioms on clustering preserving transformations  are quite a cheap way to produce labeled data sets from existing ones. 
However, the frequently cited axiomatic system 
of of Klein{}berg \cite{Kleinberg:2002}, as we show in this paper, 
is not applicable for finite dimensional Euclidean spaces, in which many algorithms  like $k$-means, operate. In particular, the so-called outer-consistency axiom fails upon making small changes in data{}point positions and inner-consistency axiom is valid only for identity transformation in general settings.  

Hence we propose an alternative axiomatic system, in which Klein{}berg's inner consistency axiom is replaced by a centric consistency axiom and outer consistency axiom is replaced by motion consistency axiom. We demonstrate that the new system is satisfiable for a hierarchical version of $k$-means with auto-adjusted $k$, hence it is not contradictory. Additionally, as $k$-means creates convex clusters only, we demonstrate that it is possible to create a version detecting concave clusters and still the axiomatic system can be satisfied. 
The practical application area of such an axiomatic system may be the generation of new labeled test data from existent ones for clustering algorithm testing. 
\end{abstract}


\section{Introduction}

Development of data mining algorithms, in particular also of clustering algorithms (see Def.\ref{def:KleinbergClusteringFunction}),  requires a considerable body of labeled data. 
The data may be used for development of new algorithms, fine-tuning of algorithm parameters, testing of algorithm implementations, comparison of various brands of algorithms, investigating of algorithm properties like scaling in sample size, stability under perturbation and other.

There exist publicly available repositories of labeled data\footnote{See e.g. \url{https://ifcs.boku.ac.at/repository/}, \url{https://archive.ics.uci.edu/ml/datasets.php}}, there exist technologies for obtaining new labeled data sets like crowd{}sourcing \cite{Chang:2017}, as well as extending existent small bodies of labeled data into bigger ones (semi-supervised learning)
as well as various cluster generators \cite{Iglesias:2019,Mount:2005}. 
Nonetheless these resources may prove sparse for an extensive development and testing efforts because of risk of overfitting, risk of instability etc. 
Therefore efforts are made to provide the developers with abundant amount of training data. 

The development of axiomatic systems may be exploited for such purposes.
Theoretical works on axiomatization of clustering algorithms are quite a cheap way to produce labeled data sets from existing ones. 
These axiomatic systems (Def.\ref{def:axiomaticSystem}) include among others definitions of data set transformations under which some properties of the clustering algorithm will remain unchanged. The most interesting property is the partition of the data. 
The axiomatic systems serve of course other purposes, like deepening the understanding of the concept of clusters, the group similarity, partition and the clustering algorithm itself.

Various axiomatic frameworks have been proposed, e.g.
for unsharp partitioning by \cite{Wright:1973},
for graph clustering by \cite{vanLaarhoven:2014},
for cost function driven algorithms  by  \cite{Ben-David:2009}, 
 for linkage algorithms by  \cite{Ackerman:2010}, 
 for hierarchical algorithms by \cite{Carlsson:2010,Gower:1990,Thomann:2015}, 
 for multi-scale clustering by \cite{Carlsson:2008}.
for settings with increasing sample sizes by   \cite{Hopcroft:2012},
for community detection by \cite{Zeng:2016},
for pattern clustering by \cite{Shekar:1988}, 
and for distance based clustering 
\cite{Kleinberg:2002}, \cite{Ben-David:2009}, see also \cite{Ackerman:2010NIPS}.
Regrettably, the natural settings for many algorithms, like $k$-means, seem to have been ignored, that is (1) the embedding in the Euclidean space (Def.\ref{def:embedding}), (2) partition of not only the sample but of the sample space, and (3) the behavior under continuous transformations (Def.\ref{def:continuousTransformation}). 
%
It would also be a useful property if the clustering would be the same under some perturbation of the data within the range of error.
\NSTUFF{Perturbation $\mathfrak{p}(S)$ is an akin concept to continuous transformation in that for some small $\epsilon>0$ the distance between  $P \in S$ and $\mathfrak{p}(P)$ is below $\epsilon$. }
The importance of the perturbation robustness in clustering has been recognized years ago and was studied by  \cite{Awasthi:2012,Balcan:2016,Moore:2016} among others.

The candidate axiomatic system for this purpose seems to be the mentioned Klein{}berg's system \cite{Kleinberg:2002}, as there are no formal obstacles to apply the axioms under Euclidean continuous settings. 
As we show, however, one of the transformations implied  by Klein{}berg's axioms, the consistency axiom transformation (Property.\ref{ax:consistency})
turns out to be identity transformation in Euclidean space, if continuity is required (Def.\ref{def:continuousTransformation}.)
Furthermore,  its special case of inner-consistency (Def. \ref{def:innerConsistency}) turns out to be identity transformation in Euclidean space even without continuity requirement (Theorem    \ref{lem:noinnerconsistency2dim4clusters}  and Theorem \ref{thm:noinnerConsistency}).
The special case of outer-consistency (Def.\ref{def:outerConsistency}) suffers also from problems under continuous transformation (Theorem \ref{thm:noGeneralConvexOuterConsistency}).

The possibility to perform continuous $\Gamma$-transformations preserving consistency is very important from the point of view of clustering algorithm testing because it is a desirable property that small perturbations of data do not produce different clusterings. In case that a consistency transform reduces to identity transform, the transformation is useless from the point of view of algorithm testing (same data set obtained).

Therefore,  the axiom of consistency has to be replaced.  
We suggest to consider replacements for inner-consistency in terms of the centric consistency (Property.\ref{ax:centricconsistency}) and the outer-consistency should be replaced with motion consistency (Def.\ref{def:motionConsistency}). 
Both special types of consistency need replacements because each of them leads to contradictions, as shown in Section \ref{sec:consistencyProblems}

We will demonstrate that we repair in this way the basic deficiency of Kleinberg, that is the self-contradiction as well as transfer the axiomatic thinking into continuous domain (Theorem \ref{thm:noncontradictoryaxiomaticsystem}). 
 In Section \ref{sec:experRealData}, we will verify the validity of our basic cluster-preserving transformations.

%
 
\section{Previous Work}\label{sec:prevWork}

\NSTUFF{Let us recall that an
\begin{definition}\label{def:axiom}
\emph{axiom} or \emph{postulate} is a proposition regarded as self-evidently true without proof.
\end{definition}
A clustering axiom is understood as a property that has to hold for a clustering algorithm in order for that algorithm to be considered as a clustering algorithm.  
\begin{definition}\label{def:axiomaticSystem}
An \emph{axiomatic system} is a set of axioms that are considered to hold at the same time. 
\end{definition}
If an axiomatic system holds for an algorithm, then one can reason about further properties of that algorithm without bothering of the details of that algorithm.  }

 Google Scholar lists about 600 citations.
of Kleinberg's axiomatic system. 
Kleinberg \cite[Section 2]{Kleinberg:2002}  defines clustering function as:
\begin{definition}\label{def:KleinbergClusteringFunction}
A \emph{clustering function} is 
a function $f$ 
that takes a distance function $d$ on [set] $S$ [of size $n\ge 2$] and
returns a partition $\Gamma$  of $S$. The sets in $\Gamma$  will 
be called its  \emph{clusters}.
\end{definition}
Given that, Kleinberg \cite{Kleinberg:2002} %
formulated axioms for 
distance-based cluster analysis (Properties \ref{ax:richness},\ref{ax:scaleinvariance}, \ref{ax:consistency}), which are rather termed properties by other  e.g.   \cite{Ackerman:2010NIPS}. 
\begin{ax} \label{ax:richness}
Let $Range(f)$ denote the set of all partitions $\Gamma$
 such that $f(d) = \Gamma$ for some distance
function $d$.
A function $f$ has the  \emph{richness} property 
if 
  $Range(f)$ is equal to the set of all partitions of $S$.
\end{ax} 
\begin{ax}  \label{ax:scaleinvariance}
A function $f$ has the 
\emph{scale-in{}variance}  property
if
for any distance function $d$ and any $\alpha  > 0$,
we have $f(d) = f(\alpha \cdot d)$.
\end{ax}

\begin{ax}  \label{ax:consistency}
Let $\Gamma$ be a partition of $S$, and $d$ and
$d'$  two distance functions on $S$. We say that $d'$ 
 is a $\Gamma$-transformation of $d$ ($\Gamma(d)=d')$ if (a) for
all $i,  j \in  S$ belonging to the same cluster of 
$\Gamma$, we have $d'(i, j) \le d(i, j)$   and (b) for
all $i,  j \in  S$ belonging to different clusters of $\Gamma$,
 we have $d'(i, j) \ge d(i, j)$.
  The clustering function 
  $f$ has the \emph{consistency} property 
if for each distance function $d$ and its $\Gamma$-transform   $d'$ the following holds: if $f(d) =\Gamma$,  then $f(d') = \Gamma$%
\end{ax} 
\Bem{
We say that $d'$ 
 is a $\Gamma$,$\delta$-transform of $d$ ($\Gamma_\delta(d)=d')$, $\delta\ge 0$) if (a) for
all $i,  j \in  S$ belonging to the same cluster of 
$\Gamma$, we have $d'(i, j) \le d(i, j)+\delta$   and (b) for
all $i,  j \in  S$ belonging to different clusters of $\Gamma$,
 we have $d'(i, j) \ge d(i, j)-\delta$.
  The clustering function 
  $f$ has the \emph{$\delta$-consistency} property 
if for each distance function $d$ and its $\Gamma$,$\delta$-transform   $d'$ the following holds: if $f(d) =\Gamma$,  then $f(d') = \Gamma$
}

For algorithms like $k$-means, Klein{}berg defined also the property of $k$-richness: in which he requires not generation of  ``all partitions of $S$'' but rather of ``all partitions of $S$ into exactly $k$ non-empty clusters''.  
 
\begin{ax} \label{ax:kRichness}
Let $Range(f)$ denote the set of all partitions $\Gamma$
 such that $f(d) = \Gamma$ for some distance
function $d$.
A function $f$ has the  \emph{$k$-richness} property 
if 
  $Range(f)$ is equal to the set of all partitions of $S$ into $k$ non-empty sets.
\end{ax}

 Klein{}berg demonstrated that 
his three \emph{axioms} (Properties of richness \ref{ax:richness}, scale-in{}variance \ref{ax:scaleinvariance} and consistency \ref{ax:consistency}) cannot be met all at once.
(but only pair-wise), see his   Impossibility Theorem \cite[Theorem 2.1]{Kleinberg:2002}.
In order to resolve the conflict, there was proposed 
a replacement of  consistency axiom with
order-consistency axiom  
\cite{Zadeh:2009},  
 refinement-consistency 
 (\cite{Kleinberg:2002}), inner/outer-consistency  \cite{Ackerman:2010NIPS},   order-in{}variance and so on (see \cite{Ackerman:2012:phd}). 
 We will be interested here in two concepts:  
\begin{definition}\label{def:innerConsistency}
\emph{Inner-consistency} is a special case of consistency in which 
the distances between members of different clusters do not change. 
\end{definition}

\begin{definition}\label{def:outerConsistency}
\emph{Outer-consistency} is a special case of consistency in which 
the distances between members of the same cluster do not change. 
\end{definition}
In this paper, we depart from the assumptions of previous papers on Klein{}berg's axiomatic system and its variants in that we introduce the assumption of continuity in the aspects: (1) the embedding of data in a fixed-dimensional space is assumed (Def. \ref{def:continuousSet}, \ref{def:embedding}), (2) the $\Gamma$-transformation has to be performed in a continuous way (Def.\ref{def:continuousTransformation}, Def.\ref{def:contconstransf}) and (3) the data set $S$ is considered as a sample from   a   probability density function over the continuous space, like discussed already by
 Pollard \cite{Pollard:1981}  in 1981 and by   Mac{}Queen \cite{MacQueen:1967}  in 1967   
 when considering the in the limit behavior of $k$-means target function, and by Klopotek   \cite{MAK:2020:consistency} for $k$-means++.  We assume that this probability density function is continuous.   
 In this paper, also, we will go beyond the classical convex-shaped structure of $k$-means clusters.

\section{Preliminaries}\label{sec:prelim}

We 
discuss embedding of the data points in a fixed-dimensional Euclidean space,
with distance between data points implied by the embedding and 
continuous versions of data transforms will be considered.  
\begin{definition}\label{def:embedding}
An \emph{embedding} of a data set $S$ into the $m$-dimensional space $\mathbb{R}^m$ is a function 
$\mathcal{E}:S \rightarrow  \mathbb{R}^m$
inducing  a distance function $d_\mathcal{E}(i,j)$ between these data points  
being the Euclidean distance between $\mathcal{E}(i)$ and $\mathcal{E}(j)$. 
\end{definition}

\NSTUFF{We consider here finite dimensional Euclidean spaces $\mathbb{R}^m$. 
The continuity in this space is understood as follows:}
\begin{definition}\label{def:continuousSet}
\NSTUFF{A space / set $S\subseteq \mathbb{R}^m$ is continuous iff 
within a distance of $\epsilon>0$ from any point $P \in S$ there exist infinitely many points $Q \in S$.} 
\end{definition}
\NSTUFF{Same applies to clusters:  within a distance of $\epsilon>0$ from any point $P$ of a cluster $\mathbf{C}$  there exist infinitely many points $Q$ of the same cluster $\mathbf{C}$. That is while a clustering algorithm like $k$-means operates on a finite sample, it actually splits the entire space  info clusters, that is it assigns class membership also to unseen data points from the space.} 
\begin{definition}\label{def:continuousTransformation}
\NSTUFF{By a continuous transformation of a (finite or infinite) data set $S$ in such a space ($S\subset \mathbb{R}^m$) we shall understand a function $\mathfrak{t}(t,S)$, where $t\in [0,1]$, for any $P\in S$ $\mathfrak{t}(0,P)=P$ and for each $t\in [0,1]$ and each $\epsilon>0$ there exist $\delta_{t,\epsilon}>0$ such that for any $0<\delta<\delta_{t,\epsilon}$ and any $t \in [0,1]$  the point 
$Q=\mathfrak{t}(t+\delta,P)$  lies within the distance of $\epsilon$ from $\mathfrak{t}(t,P)$ for each $P$.}
\end{definition}

$k$-means, kernel-$k$-means, their variants and many other clustering functions  rely on the embedding into Euclidean space which  raises the natural question on the behavior under continuous changes of positions in space. 
Let $\mathfrak{E}^S_m$ be the set of all possible embedding{}s of the data set $S$  in $\mathbb{R}^m$.
Define: 
\begin{definition}\label{def:contconstransf}
\NSTUFF{ 
  We speak about \emph{continuous consistency transformation} or
  \emph{continuous $\Gamma$-transformation}
  $\mathfrak{t}$ of a clustering $\Gamma$ if $\mathfrak{t}$  is a continuous transformation as defined in Def.\ref{def:continuousTransformation}  
  and 
   $d(\mathfrak{t}(t,P),\mathfrak{t}(t,Q))\ge d(\mathfrak{t}(t',P),\mathfrak{t}(t',Q))$ for any $0\le t<t'\le 1$ if $P,Q$ belong to the same cluster $C \in \Gamma$
and 
   $d(\mathfrak{t}(t,P),\mathfrak{t}(t,Q))\le d(\mathfrak{t}(t',P),\mathfrak{t}(t',Q))$ for any $0\le t<t'\le 1$ if $P,Q$ do not belong to the same cluster.  
}
\end{definition}
 The transform for the probability density function is understood in a natural way as a transformation of the probability density estimated from the sampling. The clustering function $f$ is also understood as partitioning the space in such a way that with increasing sample size the sample-based partition of the space approximates the partition of the space based on the probability density, as discussed in \cite{MAKRAK:2020:ICAISC2020}. 
 Continuous inner-$\Gamma$-transformation/consistency and continuous outer-$\Gamma$-transformation/consistency are understood in analogy to 
 inner/outer-$\Gamma$-transformation/consistency.

 Subsequently, whenever we talk about distance, we mean Euclidean distance. 
\begin{ax}  \label{ax:continuousconsistency}
   A clustering function  $f$ has the   \emph{continuous-consistency} property,
  if for each dataset $S$ and the induced clustering $\Gamma=f(S)$   each 
  \emph{continuous $\Gamma$-transformation}
  $\mathfrak{t}$ has the property 
  that for each $0 \le t \le 1$ the clustering of  $\mathfrak{t}(t,S)$ will result in the same clustering $\Gamma$. 
\end{ax} 

The property of continuous consistency is a very important notion from the point of view of clustering in the Euclidean space, especially in the light of research on perturbation robustness   \cite{Awasthi:2012,Balcan:2016,Moore:2016}. 

Note that a $\Gamma$-transformation cannot be in general replaced by a finite sequence of inner-consistency and outer $\Gamma$-transformations. 
But if we have to do with a continuous $\Gamma$-transformation then it can be replaced by a continuum of inner-consistency and outer $\Gamma$-transformations

Let us make the remark, that the  inner-$\Gamma$-transformation and the outer-$\Gamma$-transformation combination cannot replace the $\Gamma$-transformation, even if we use a finite sequence of such transformations.  
However we can think of a sequence of sequences of $\Gamma$-transformations approximating a continuous $\Gamma$-transformation. By a sequence $\mathfrak{t}_{i,n}$ of $\Gamma$-transformations of a data set $S$ we understand the following sequence: $\mathfrak{t}_{0,n}(S)=S$ and for each $i = 1,\dots, n$, $\mathfrak{t}_{i,n}(S)$ is a $\Gamma$-transformation of $\mathfrak{t}_{i-1,n}(S)$. 
In particular, for  a continuous $\Gamma$-transformation $\mathfrak{t}$, the sequence 
$\mathfrak{t}(i/n,S)$ for $i=0,\dots,n$ wold be such a sequence. 
We say that a sequence $\mathfrak{t'}_{i,n}$ approximates the the continuous transformation 
 $\mathfrak{t}$ with precision $\pi$, if for each $P$ in $S$
 the distance between $\mathfrak{t'}_{i,n}(P)$ and $\mathfrak{t}(t,P)$, where $t \in [i/n,(i+1)/n]$ does not exceed $\pi$. 
 \begin{theorem}\label{thm:innerouterfull}
 If $\mathfrak{t}$ is a continuous-$\Gamma$-transformation, then there exists an $n$ for which the approximating sequence of $\Gamma$-transformations  $\mathfrak{t'}_{i,n}$ can be found approximating  $\mathfrak{t}$ up to $\pi$. 
 What is more, for possibly a larger $n$, a sequence exists consisting of a mixture of only inner and outer $\Gamma$-transformations. 
 \end{theorem}

\Bem{
\begin{theorem}\label{thm:innerouterfull}
In the above definition \ref{def:contconstransf}, 
the $\Gamma$,$\delta$-$\Gamma$-transformation at each step 
may be either inner- or outer- $\Gamma$,$\delta$-$\Gamma$-transformation. 
In other words, continuous $\Gamma$-transformation can be constructed as a limit from sequences of inner- and outer- $\Gamma$,$\delta$-$\Gamma$-transformations. 
\end{theorem}
}
\begin{proof}
It can be shown by appropriate playing with the $\pi$s, $\epsilon$s at various scales of $\delta$. 
\end{proof}

The Klein{}berg's axioms reflect three important features of a clustering (1) compactness of clusters, (2) separation of clusters and (3) the balance between compactness and separation. In particular, the inner-consistency property says that the increase of compactness of a cluster should preserve the partition of data. 
The outer  consistency property says that the increase of separation of clusters should preserve the partition of data. 
The scale-in{}variance states that preserving the balance between compactness and separation should preserve partition. The richness means that variation of  balances between compactness and separation should lead to different partitions.

We will demonstrate however a number of impossibility results for continuous inner and outer-consistency. Essentially we show that in a number of interesting cases these operations are reduced to identity operations. 

Therefore we will subsequently seek replacements for continuous consistency, that will relax their rigidness, by  proposing  
centric consistency 
introduced in \cite{MAKRAK:2020:ICAISC2020} as a replacement for inner-consistency, and   motion-consistency, introduced in \cite{MAKRAK:2020:AIAI2020} as a replacement for outer-consistency. We will demonstrate that under this replacement, the (nearly) richness axiom may be fulfilled in combination with centric consistency and scale-in{}variance.

As we will refer frequently to the $k$-means algorithm, let us recall that it is aimed at minimizing  the cost function $Q$ (reflecting its quality in that the lower $Q$ the higher the quality)  of the form: 

\begin{equation} \label{eq:Q::kmeans}
Q(\Gamma)=\sum_{i=1}^m\sum_{j=1}^k u_{ij}\|\textbf{x}_i - \boldsymbol{\mu}_j\|^2
=\sum_{j=1}^k \frac{1}{n_j} \sum_{\mathbf{x}_i, \mathbf{x}_l \in C_j} \|\mathbf{x}_i - \mathbf{x}_l\|^2 
\end{equation} 
$$=\sum_{j=1}^k \frac{1}{2n_j} \sum_{\mathbf{x}_i \in C_j} 
\sum_{\mathbf{x}_l \in C_j} \|\mathbf{x}_i - \mathbf{x}_l\|^2 $$
for a data{}set $\mathbf{X}$
under some partition $\Gamma$ into the predefined number $k$ of clusters, 
where  $u_{ij}$ is an indicator of the membership of data point $\textbf{x}_i$ in the cluster $C_j$ having the cluster center at $\boldsymbol{\mu}_j$ (which is the cluster's gravity center). 
Note that \cite{Pollard:1981} modified this definition by dividing the right hand side by $m$ in order to make comparable values for samples and the population, but we will only handle samples of a fixed size so this definition is sufficient for our purposes. 
A $k$-means algorithm finding exactly the clustering optimizing $Q$ shall be refereed to as $k$-means-ideal. Realistic implementations start from a randomized initial partition and then improve the $Q$ iteratively. An algorithm where the initial partition is obtained by random assignment to clusters shall be called random-set $k$-means. For various versions of $k$-means algorithm see e.g. \cite{STWMAK:2018:clustering}. 

\section{Main results}\label{sec:main}

\subsection{Internal Contradictions of Consistency Axioms}\label{sec:consistencyProblems}

\NSTUFF{
In this section we demonstrate that inner consistency axiom alone  induces contradictions (Theorem    \ref{lem:noinnerconsistency2dim4clusters}  and Theorem \ref{thm:noinnerConsistency}).
The continuous versions of the respective consistency transformations amplify the problems. 
Theorem \ref{thm:noGeneralConsistency} shows that if concave clusters are produced, continuous consistency transform may not be applicable. 
In general settings, continuous outer consistency transform (Theorem \ref{thm:noGeneralConvexOuterConsistency}) may not be applicable to convex clusters either. 
These inner contradictions can be cured by imposing additional constraints on continuous consistency transform in form of gravitational (Def.\ref{def:gravitcons}, or homothetic (Theorem \ref{thm:homotheticgravcons}) or convergent (Property \ref{ax:convergentconsistency} transforms. 
}

We claim first of all that   
inner-consistency cannot be satisfied non-trivially for quite natural clustering tasks, see  Theorem    \ref{lem:noinnerconsistency2dim4clusters}  and Theorem \ref{thm:noinnerConsistency}. 
  
\begin{theorem} \label{lem:noinnerconsistency2dim4clusters}
In $m$-dimensional Euclidean space for a data set with no co{}hyper{}planar{}ity of any $m+1$ data points (that is when the points are in general position) the inner-$\Gamma$-transform is not applicable non-trivially, if we have a partition into more than $m+1$ clusters.    
\end{theorem}
See proof on page \pageref{lem:noinnerconsistency2dim4clustersPROOF}.
By non-trivial we mean the case of $\Gamma$-transformation different from isometric transformation. 
Note that here we do not even assume continuous inner-consistency. Non-existence of non-trivial inner-consistency makes continuous one impossible. 

As the data sets with properties mentioned in the premise of  Theorem \ref{lem:noinnerconsistency2dim4clusters} are always possible, no algorithm producing the $m+1$ or more clusters in $\mathbb{R}^m$  has non-trivially the inner-consistency property. 
This theorem sheds a different light on 
claim of    
\cite{Ackerman:2010NIPS}  that $k$-means does not possess the property of inner-consistency. 
No algorithm producing that many clusters  has this property if we consider embedding{}s into a fixed dimensional Euclidean space. Inner-consistency is a special case of Klein{}berg's consistency. 

The above Theorem can be slightly generalized:
\begin{theorem} {} \label{thm:noinnerConsistency}
In a fixed-dimensional space $\mathbb{R}^m$ under Euclidean distance 
if there are at least $m+1$  clusters 
such that we can pick from each cluster a data point so that $m+1$ points are not co-hyper{}planar (that is they are "in general position"), and at least two clusters contain  at least two data points each,
then  inner-$\Gamma$-transform is not applicable (in non-trivial way).
\footnote{This impossibility does not mean that there is an inner-contradiction 
when executing the inner-$\Gamma$-transform. 
Rather it means that considering inner-consistency is pointless because inner-$\Gamma$-transform is in general impossible except for isometric transformation.
}
\end{theorem}
See proof on page \pageref{thm:noinnerConsistencyPROOF}. Again continuous inner-consistency was not assumed.

\begin{theorem} {} \label{thm:noGeneralConsistency}
If a clustering function is able to produce concave clusters such that there are   mutual  intersections of one cluster with  convex hull of  a part of the other cluster,  then it does not have the   \emph{continuous} consistency property if  identity transformation is excluded.
\end{theorem}
See proof on page \pageref{thm:noGeneralConsistencyPROOF}. 
The Theorem \ref{thm:noGeneralConsistency} referred to clustering algorithms producing   concave clusters. 
But what when we restrict ourselves to partitions into  convex clusters? Let us consider the continuous outer-consistency. 

\begin{definition}
Consider three clusters $C_1,C_2,C_3$. 
If the points $S_i$, the separating hyper{}planes $h_{ij}$ and vectors $\mathbf{v}_{ij}$ with properties mentioned below  exist, then we say that the three clusters follow the \emph{three-cluster-consistency principle}. 
The requested properties are: 
(1) There exist  points $S_1, S_2, S_3$ in convex hulls of  $C_1,C_2,C_3$ resp. such that   $C_{i}$ is separated from $S_j$ by a hyper{}plane $h_{ij}$ such that  the line segment  ${S_iS_j}$ is orthogonal to  $h_{ij}$  for $i,j=1,2,3$. 
(2) 
Let $\mathbf{v}_{ij}$ be the speed vector of cluster $C_i$ relatively to $C_j$, $\mathbf{v}_{13}$ for $i,j=1,2,3$ such that      
$\mathbf{v}_{12}\perp h_{12}, $ $\mathbf{v}_{13}\perp h_{13}$,
 and
$\mathbf{v}_{23}\perp h_{23}$, 
and 
$\|\mathbf{v}_{12}\|/|S_1S_2| = \|\mathbf{v}_{13}\|/|S_1S_3|$. \end{definition}
Note that condition (2) implies that 
$ \mathbf{v}_{23} =  \mathbf{v}_{12}-\mathbf{v}_{13} $ and 
$\|\mathbf{v}_{12}\|/|S_1S_2| = \|\mathbf{v}_{23}\|/|S_2S_3|$. 

\begin{theorem} {} \label{thm:noGeneralConvexOuterConsistency}
If a clustering function may produce   partitioning such that there exists a sequence of cluster indices (first and last being identical) so that one is unable to find hyper{}planes separating the clusters not  violating for any sub{}sequence of three clusters the three-cluster-consistency principle and without a mismatch between the first and the last $S$ point,  then this function does not have  continuous outer-consistency property if we discard the isometric transformation. 
\end{theorem}
See proof on page \pageref{thm:noGeneralConvexOuterConsistencyPROOF}. 

Note that $k$-means clustering function does not suffer from the deficiency described by the precondition of Theorem \ref{thm:noGeneralConvexOuterConsistency} because it implies a Voronoi tessellation: if we choose always  the gravity centers of the clusters as the $S$ points, then $h$ hyper{}planes can always be chosen as hyper{}planes orthogonal to $S_jS_i$  lying half-way between $S_i$ and $S_j$. Hence
\begin{theorem}\label{thm:kmeansouterconsistencyproperty}
Given $k$-means clustering (in form of a Voronoi tessellation), there exists a non-trivial continuous outer-$\Gamma$-transform preserving the outer-consistency property in that we fix the position of one cluster $C$ and move each other cluster $C_j$ relocating each data point of cluster $C_j$ with a speed vector proportional to the vector connecting the cluster center $C$ with cluster center $C_j$.  
\end{theorem}

So, by irony, the $k$-means clustering algorithm for which Klein{}berg demonstrated that it does not have the consistency property, is the candidate for a non-trivial continuous  outer  consistency variant. Let us elaborate now on this variant. 

\begin{definition}\label{def:gravitcons}
A $\Gamma$-transform has the property of gravitational consistency, if within each cluster the distances between gravity centers of any two disjoint subsets of cluster elements do not increase.  
\end{definition}

\begin{theorem}\label{thm:gravcons}
Under gravitational $\Gamma$-transformation, the $k$-means possesses the property of Klein{}berg's consistency and continuous consistency. 
\end{theorem}
See proof on page \pageref{thm:gravconsPROOF}. 

The gravitational consistency property is not easy to verify (the number of comparisons is exponential in the size of the cluster). However, some transformations have this property, for example:

\begin{theorem}\label{thm:homotheticgravcons}
The homothetic transformation of  cluster with a ratio  in the range (0,1)  in the whole space or any subspace has the property that   the distances between gravity centers of any two disjoint subsets of cluster elements do not increase.   
\end{theorem}
See proof on page \pageref{thm:homotheticgravconsPROOF}. 

The gravitational consistency may be deemed as a property that is too $k$-means specific. 

Therefore,  we considered, in the paper  \cite{MAKRAK:2020:ICAISC2020}, the problem of Klein{}berg's consistency in its generality. We showed there that the problem of Klein{}berg's contradictions lies in the very concept of consistency because his $\Gamma$-transformation creates new clusters within existent one (departs from cluster homogeneity) and clusters existent ones into bigger ones (balance between compactness and separation is lost).   This effect can be avoided probably only via requiring the convergent consistency, as explained below.

\begin{ax}  \label{ax:convergentconsistency}
Define the convergent-$\Gamma$-transform as $\Gamma$-transform from Property \ref{ax:consistency}
in which if  $ d(i,j)\le d(k,l)$, then also 
$d'(i,j)\le d'(k,l)$ and $d(i,j)/d(k,l)\le  d'(i,j)/d'(k,l)$.
The clustering function $f$ has the \emph{convergent consistency} property 
if for each distance function $d$ and its convergent $\Gamma$-transformation, with    $f(d)=d'$ the following holds: if $f(d) =\Gamma$,  then $f(d') = \Gamma$%
\end{ax}

\begin{theorem}
The axioms of richness, scale-in{}variance and convergent-con\-sis\-ten\-cy are not contradictory.
\end{theorem}
See proof in paper \cite{MAKRAK:2020:ICAISC2020}. 

When we actually cluster data sampled from a continuous distribution, we would expect the "in the limit" condition to hold, that is that the clustering obtained via a sample converges to the clustering of the sample space. 

\begin{theorem}
If the in-the-limit condition has to hold then the convergent-$\Gamma$-transform must be an isomorphism.
\end{theorem}
See proof in paper \cite{MAKRAK:2020:ICAISC2020}. 

For this reason, under conditions of continuity, we need to weaken the constraints imposed by Klein{}berg on clustering functions via his consistency axiom. 
As stated in Theorem \ref{thm:innerouterfull}, the $\Gamma$-transformation of Klein{}berg under continuous conditions can be replaced by a limit on a sequence of inner and outer $\Gamma$-transforms. Deficiencies of both have been just listed. 
Therefore we propose here two relaxations of the two properties: centric consistency as a replacement of inner-consistency and motion consistency as a replacement of outer-consistency. 
\subsection{Centric Consistency}

Let us start  with the centric consistency. It is inspired by the result of Theorem \ref{thm:homotheticgravcons}.

\begin{definition}
Let $\mathcal{E}$ be an embedding of the data{}set $S$ with distance function $d$ (induced by this embedding). 
Let $\Gamma$ be a partition of this data{}set.
Let $C\in \Gamma$ and let $\boldsymbol\mu_c$ be the gravity center of the cluster $C$ in $\mathcal{E}$.
We say that we execute the \emph{$\Gamma^*$} transformation (or a centric  transformation, $\Gamma(d;\lambda)=d'$) if for some $0<\lambda\le 1$ a new embedding $\mathcal{E}'$ is created in which  each element $x$ of $C$, with coordinates $\mathbf{x}$ in $\mathcal{E}$, has coordinates $\mathbf{x'}$ in $\mathcal{E}$  such that   $\textbf{x'}=\boldsymbol\mu_c+\lambda(\textbf{x}-\boldsymbol\mu_c)$, all coordinates of all other data points are identical, and $d'$ is induced by $\mathcal{E}'$.  $C$ is then said to be subject of the centric transform.
\end{definition} 

Note that the set of possible centric $\Gamma$-transformations for a given partition is neither a subset nor super{}set of the set of possible Klein{}berg's $\Gamma$-transformation in general. However, if we restrict ourselves to the fixed-dimensional space and to the fact that the inner-consistency property means in this space isometry, as shown in Theorems     \ref{lem:noinnerconsistency2dim4clusters}  and   \ref{thm:noinnerConsistency},  
 then obviously the centric consistency property can be considered as much more general, in spite of the fact that  it is a $k$-means clustering model-specific adaptation of the general idea of shrinking the cluster.

\begin{ax}\label{ax:centricconsistency}
A clustering method has the property of \emph{centric consistency}
if after a $\Gamma^*$ transform it returns the same partition. 
\end{ax}

\begin{theorem}{} \label{thm:globalCentricCinsistencyFor2means}
$k$-means algorithm satisfies centric consistency property in the following way:  
if the partition $\Gamma$ of the set $S$ with distances $d$ is a global minimum of $k$-means, and $k=2$, and the partition $\Gamma$  has been subject to centric $\Gamma$-transformation  yielding distances $d'$,  then $\Gamma$ is also a global minimum of $k$-means under distances $d'$.  
\end{theorem}
See proof in paper \cite{MAKRAK:2020:ICAISC2020}. 

For the sake of brevity, let us use the following convention subsequently. By saying that a $\Gamma$ transformation produced the clustering $\Gamma'$ out of clustering $\Gamma$ we mean that the original embedding  was changed in such a way that any data point $x$ with original coordinates $\mathbf{x}$ is located now at $\mathbf{x'}$. If we say that the transformation turned elements of sets $P,Q$ into $P,Q'$, then we mean that elements of $P$ did not change coordinates while the elements of $Q$ changed coordinates under new embedding.

\begin{theorem}{} \label{thm:globalCentricConsistencyFor2meansSubsetted} 
Let a partition $\{T,Z\}$ be an optimal partition under 
$2$-means algorithm.
Let a subset $P$ of $T$ be subjected to centric transform 
yielding $P'$ (that is all points of $P$ were moved to the gravity center of $P$ by a factor $\lambda$), and $T'=(T-P)\cup P'$. Then  
partition $\{T',Z\}$ is an optimal partition of $T'\cup Z$ under $2$-means. 
    \end{theorem}

See the proof on page
\pageref{thm:globalCentricConsistencyFor2meansSubsettedPROOF}.

Let us  discuss a variant of bi{}sectional-$k$-means by \cite{Steinbach:2000}. 
The idea behind bi{}sectional versions of $k$-means is that we apply $2$-means to the entire data set and then recursively $2$-means is applied to some cluster obtained in the previous step, e.g. the cluster with the largest cardinality. 
The algorithm is terminated e.g. upon reaching the desired number of clusters, $k$. Note that in this case the $k$-means-ideal quality function is not optimized, and even the cluster borders do not constitute a Voronoi diagram. 
However, we will discuss, for the purposes of this paper,  a version of $k$-means such that the number of clusters is selected by the algorithm itself. 
We introduce a stopping criterion that a cluster is not partitioned, if for that cluster the decrease of $Q$ (Def. \ref{eq:Q::kmeans}) for the bisection relatively to the original cluster $Q$ does not rich  some threshold. 
An additional stopping criterion will be used, that the number of clusters cannot exceed $k$.
It is easily seen that such an algorithm is scale-invariant (because the stopping criterion is a relative one) and it is also rich "to a large extent", that is we exclude partitions with more than $k$ clusters only. 
If a clustering algorithm can return any clustering except one with the number of  cluster over $k$,   we shall call this property $k\downarrow$-near-richness. 
We shall call it \emph{bi{}sectional-$auto-k$-means} algorithm. 
Theorem \ref{thm:globalCentricConsistencyFor2meansSubsetted}  implies that
\begin{theorem} \label{thm:autocentricinvrich}
 Bi{}sectional-$auto-k$-means algorithm is   centric-consistent,  (scale)-invariant and $k\downarrow$-nearly-rich.  
\end{theorem}
See proof on page \pageref{thm:autocentricinvrichPROOF}.
Note that the Klein{}berg's impossibility theorem \cite{Kleinberg:2002} could be worked around so far only if we assumed clustering into exactly $k$ clusters. But one ran into contradictions already when the number of clusters could have a range of more than two values, because the so-called anti-chain property of the set of possible partitions does not hold any more (see \cite{Kleinberg:2002} for the impact of missing anti{}chain property on the contradictions of Klein{}berg axioms). So this theorem shows the superiority of the concept of centric consistency. Though centric consistency relaxes only inner- consistency, nonetheless Klein{}berg's axiomatic system with inner-consistency fails also. 

Let us now demonstrate theoretically, that $k$-means algorithm really fits 
\textit{in the limit} the centric-consistency axiom.
\begin{theorem}{} \label{thm:localCentricConsistencyForKmeans} 
$k$-means algorithm    satisfies
 centric consistency in the following way:  
if the partition $\Gamma$ is a local minimum of $k$-means, 
and the partition $\Gamma$  has been subject to centric consistency yielding $\Gamma'$, then $\Gamma'$ is also a local minimum of $k$-means.  
\end{theorem}
See proof on page \pageref{thm:localCentricConsistencyForKmeansPROOF}. 
However, it is possible to demonstrate that 
the newly defined transform preserves also the global optimum of $k$-means.
\begin{theorem}{} \label{thm:globalCentricCinsistencyForKmeans} 
$k$-means algorithm    satisfies
 centric consistency in the following way:  
if the partition $\Gamma$ is a global minimum of $k$-means, 
and the partition $\Gamma$  has been subject to centric consistency yielding $\Gamma'$, then $\Gamma'$ is also a global minimum of $k$-means.  
\end{theorem}
See the proof on page
\pageref{thm:globalCentricCinsistencyForKmeansPROOF}.

Hence it is obvious that 
\begin{theorem}{} \label{thm:KleinbergImpossibilityDeniedForKmeans} 
$k$-means algorithm    satisfies
Scale-in{}variance,
 $k$-Richness,
and
 centric Consistency.
\end{theorem}
Note that $k$-means does not have the property of inner-consistency. This means that the centric consistency truly relaxes inner-consistency. 

\NSTUFF{
Under some restrictions, the concept of "centric consistency" may be broadened to non-convex clustering methods. Consider the $k$-single link algorithm product $\Gamma$. Each cluster $C$ can be viewed as a tree $T_C$ if we consider the links in the cluster as graph edges. Define the area $A_C$ of a cluster $C$ as the union of balls centered at each node and with a radius equal to the longest edge incident with the node. A cluster may be deemed \emph{link-ball-separated} if for each node $N\not\in C$ the distance of $N$ to  $A_C$ is bigger than any edge incident with $N$. 
In such a case let move a leaf node $L$ of $T_C$   along the edge connecting it to $T_C$ by factor $1>\lambda_L>0$.   
For non-leaf nodes $M$ if a branch starting at $M$ lies entirely within original $A_C$, let move all nodes of the branch can be moved by the same factor $\lambda_M$ towards $M$. 
Both operations be called \emph{semi-centric transformation}.

\begin{theorem}\label{thm:GeneralCentricConsistency}
(a) If the cluster $C$ is \emph{link-ball-separated}, then application of semi-centric transformation 
yield a new data set that has the same cluster structure as $\Gamma$.
and $C$ remains  \emph{link-ball-separated}. 
\\
(b) If the cluster $C_1$ is \emph{link-ball-separated} from other clusters and the cluster $C_2$ is \emph{link-ball-separated} from other clusters and a just-mentioned transformation is performed on $C_1$, then both remain  \emph{link-ball-separated}
\end{theorem}

The proof relies on trivial geometric observations. 
Part (a) is valid because distances will decrease within a cluster and no node from outside will get closer to $N'$ than to nodes from its own cluster, 
though now the tree  $T'_C$  of $C$ may connect other nodes than $T_C$. The cluster area $A'_C$ will be a subarea of $A_C$ hence $C$ is still  \emph{link-ball-separated} from the other clusters. This transformation may violate Kleinberg's outer as well as inner consistency constraints, but will still yield new data sets with same clustering properties. 
Part (b) holds due to smaller decrease of between-cluster distances compared to the $\lambda$. .

Note that such "area" restrictions are not needed in case of $k$-means centric consistency. }

\subsection{Motion Consistency}

Centric consistency replaces only one side of Klein{}berg's consistency, the so-called inner-consistency. But we need also a replacement for the outer-consistency. 
Let us therefore introduce the motion-consistency  
(see \cite{MAKRAK:2020:AIAI2020}.).

\begin{definition}
\emph{Cluster area} is any solid body containing all 
cluster data points. 
\emph{Gap between two clusters} is the minimum distance between the cluster areas, i.e., Euclidean distance between the closest points of both areas.  
\end{definition}

\begin{definition}
Given a clustered data set embedded in a fixed dimensional Euclidean space, the \emph{motion-transformation} is any continuous transformation of the data set in the fixed dimensional space that (1) preserves the cluster areas (the areas may only be subject of isomorphic transformations) and (2) keeps the minimum required gaps between clusters (the minimum gaps being fixed prior to transformation). By \emph{continuous} we mean: there exists a continuous trajectory for each data point such that the conditions (1) and (2) are kept all the way. 
\end{definition}

\begin{definition} \label{def:motionConsistency}
A clustering method has the property of \emph{motion-consistency},
if it returns the same clustering 
after motion-transformation.  
\end{definition}

Note that motion-consistency is a relaxation of outer-consistency because it does not impose restrictions on increasing distances between points in different clusters but rather it requires keeping a distance between cluster gravity centers. 

\begin{theorem}\label{thm:localminskmeans}
If random-set $k$-means has a local minimum in ball form that is such that the clusters are enclosed into equal radius balls centered at the respective cluster centers, and gaps are fixed at zero, then the motion-transform preserves this local minimum. 
\end{theorem}
See proof in paper \cite{MAKRAK:2020:AIAI2020}.

However, we are not so interested in keeping the local minimum of $k$-means, but rather the global one (as required by motion-consistency property definition). 

In \cite{MAKRAK:2020:AIAI2020}, we have discussed some special cases of motion-consistency. Here, however, we are interested in the general case. 

\begin{theorem}\label{thm:GeneralMotionConsistency}
Let the partition $\Gamma_O$ be the optimal clustering for $k$-means. Let $R$ the radius such that, for each cluster $C\in\Gamma_O$, the ball centered at gravity center of $C$ and with radius $R$ contains all data points of $C$. 
Let us perform a transformation according to Theorem 
\ref{thm:kmeansouterconsistencyproperty} yielding a clustering $\Gamma$ so that the  distances between the cluster centres is equal to or greater than the distance between them under $\Gamma_O$ plus $4R$. (This transformation is by the way motion-consistent.) Then the clustering is motion consistent under any motion transformation keeping cluster center distances implied by $\Gamma$.
\end{theorem}
See proof on page \pageref{thm:GeneralMotionConsistencyPROOF}. 

So we can state that 
\begin{theorem}\label{thm:noncontradictoryaxiomaticsystem}
The axioms of $k$-richness, scale-in{}variance, centric-consistency and motion-consistency are not contradictory.
\end{theorem}
The proof is straight forward: $k$-means algorithm has all these properties. 

Note that the theorem \ref{thm:GeneralMotionConsistency} can be extended to a broader range of clustering functions. 
Let $\mathcal{F}$ be the class of distance-based clustering functions embedded into Euclidean space such that the clustering quality does not decrease   with the decrease of distances between cluster elements and changes  of distances between elements from distinct clusters do not impact the quality function.  \footnote{This property holds clearly for $k$-means, if quality is measured by inverted $Q$ function, but also we can measure cluster quality of $k$-single-link with the inverted longest link in any cluster and then the property holds.} 

\begin{theorem}\label{thm:ClassFGeneralMotionConsistency}
If we replace in Theorem \ref{thm:GeneralMotionConsistency} the phrase " $k$-means." with "a function from the class  $\mathcal{F}$, then the Theorem is still valid. 
\end{theorem}
See proof on page \pageref{thm:ClassFGeneralMotionConsistencyPROOF}.

But what about the  richness, scale-in{}variance, centric-consistency and motion-consistency? 
As already mentioned, we will not consider the full richness but rather a narrower set of possible clustering{}s that still does not have the property of anti-chain. 

Consider the Auto-means algorithm introduced in \cite{MAKRAK:2020:ICAISC2020}. It differs from the just introduced {bi{}sectional-$auto-k$-means} in that two parameters: $p, g$ are introduced such that at each step   both created clusters $A,B$  have cardinalities such that $p\le |A|/|B| \le 1/p$ for some parameter $0<p<1$, and $|A|\ge m+1, |B|\ge m+1$, where $m$ is the dimensionality of the data set, and   both $A$ and $B$ can be enclosed in a ball centered at each gravity center with a common radius $R$ such that the distance between gravity centers is not smaller than $(2+g)R$, where $g>0$ is a relative gap parameter.
 
Let us introduce the concept of radius-R-bound motion transform of two clusters $A,B$ as motion transform preserving the gravity center of the set $A \cup B$ and keeping all data points within the radius $R$ around the gravity center of $A\cup B$. 
A hierarchical motion transform for auto-means should be understood as follows: If clusters $A,B$ are at the top of the hierarchy, then the radius-R-bound motion transform is performed with $R=\infty$. If clusters $C,D$ are sub{}clusters of a cluster $A$ from the next higher hierarchy level where $A$ was included into a ball of radius $R'$, then $R'$-bound motion transform is applied to the clusters $C,D$. If the hierarchical motion transform does not change clustering, then we speak about hierarchical motion consistency. 

\begin{theorem}
The axioms of $k\downarrow$-nearly-richness, scale-in{}variance, centric-consistency and hierarchical motion-consistency are not contradictory.
\end{theorem}
The proof is straight forward: Auto-means algorithm has all these properties.

The disadvantage of using the $k$-means clustering is that it produces convex (in particular ball-shaped) clusters. There are, however, efforts to overcome this limitation via constructing clusters with multiple centers 
by modifying the $k$-means-cost function \cite{Nie:2019}, by a combined $k$-means clustering and agglomerative clustering based on data projections onto lines connecting cluster centers \cite{Liu:2009}, or by combining $k$-means-clustering with single-link algorithm \cite{Lin:2005}. 
Let us follow the spirit of the latter paper and introduce the concave-$k$-means algorithm in the following way: 
First perform the $k$-means clustering. Then construct a minimum weight spanning tree, where the edge weight is the distance between cluster centres. Then  in such a way that the weight of an edge connecting two clusters is the quotient of the actual distance between cluster centres.
Then remove $l-1$ longest edges from the tree, producing $l$ clusters. The algorithm shall be called $k$-means-$l$-MST algorithm.  
As visible in the examples in Section \ref{sec:motionconsistency}
if we perform centric transformation for such a clustering  on constituent $k$-means clusters, the clustering is preserved.

\begin{theorem}\label{thm:concavemotion}
If the areas of clusters could be moved in such a way that the distances between the closest points of these clusters would not be lower than some quantity $m_d$, then the data can be clustered using $k$-means-$l$-MST algorithm in for $k$ bigger than some $k_d$ in such a way that following some centric $\Gamma$-transform on $k$-means clusters 
a motion-$\Gamma$-transformation can be applied preserving the clustering (motion consistency property). 
\end{theorem}
See the proof on page \pageref{thm:concavemotionPROOF}.

In this way 
\begin{itemize}
    \item We have shown that  the concept of consistency introduced by Klein{}berg is not suitable for clustering algorithms operating in continuous space as a clustering preserving $\Gamma$-transform reduces generally to identity transform. 
    \item Therefore we proposed, for use with $k$-means, property of centric consistency as a replacement for inner-consistency and the property of motion consistency as a replacement of outer-consistency, that are free from the shortcomings of Klein{}berg's $\Gamma$-transform.
    \item The newly proposed transforms can be used as a method for generating new labeled data for testing of $k$-means like clustering algorithms. 
    \item In case of continuous axiomatisation, one can overcome the Klein{}berg's impossibility result on clustering by slightly strengthening his richness axiom ($k\downarrow$-nearly-richness which is much weaker than $k$-    richness because it allows for non-anti-chain partition sets) and by relaxing axioms of inner-consistency to centric-consistency and outer-consistency to hierarchical-motion-consistency and keeping the scale-in{}variance. 
\end{itemize}

\begin{figure}
\includegraphics[width=0.4\textwidth]{\figaddr{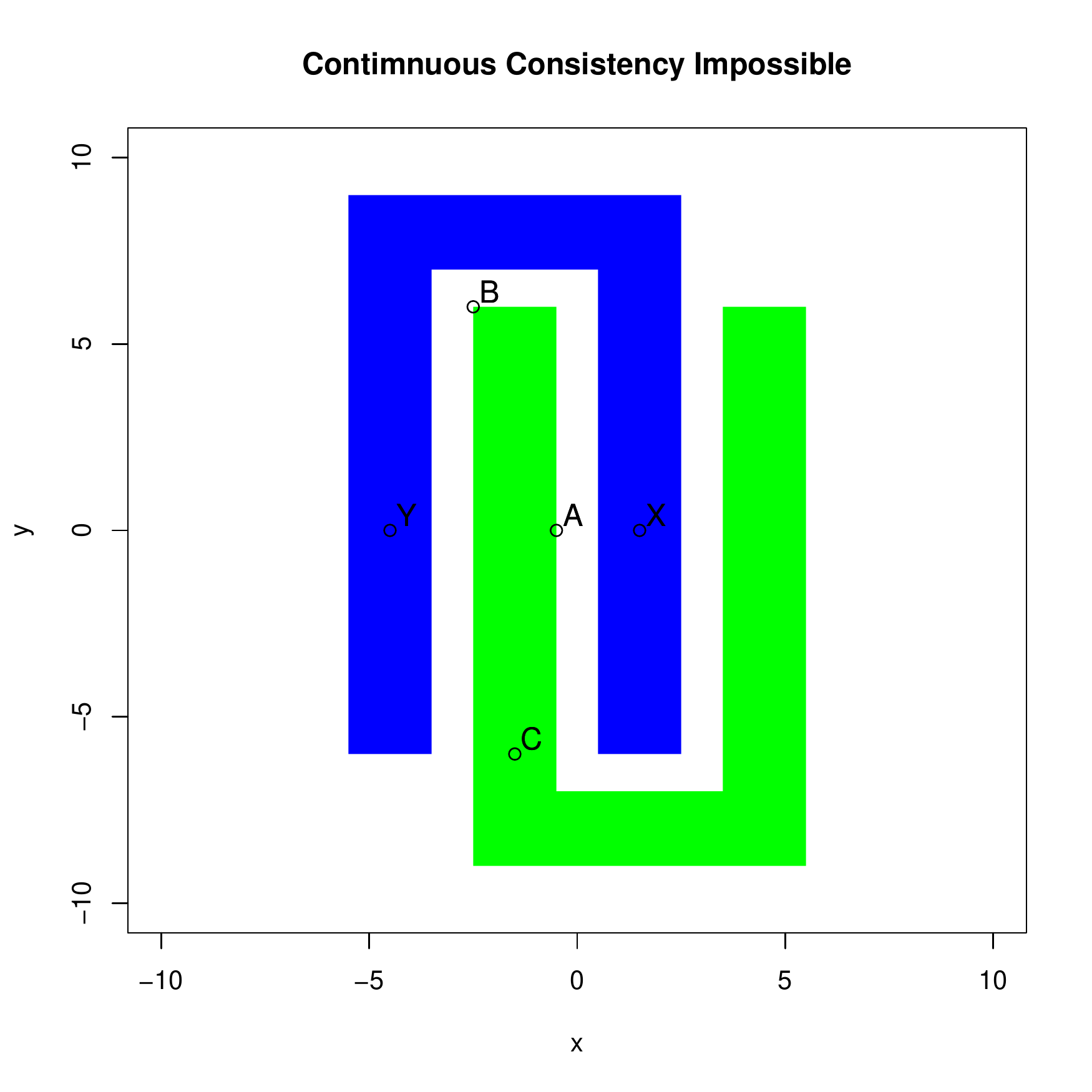}}  %
\includegraphics[width=0.4\textwidth]{\figaddr{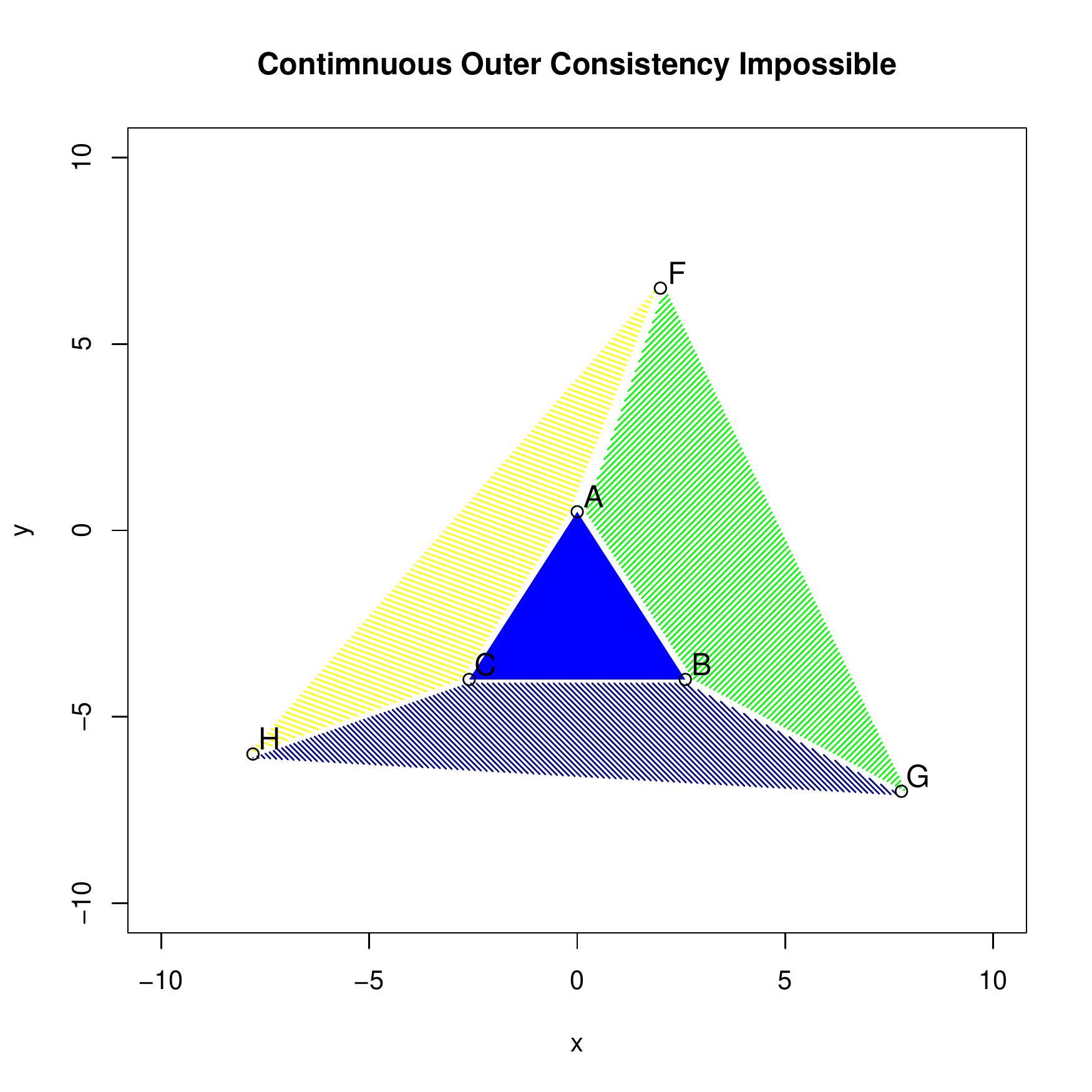}} %
\caption{Impossible continuous $\Gamma$-transform (left figure) and impossible continuous outer-$\Gamma$- Transform and impossible inner-$\Gamma$- transform (right figure)}
\label{fig:impossibleConsistency}
\end{figure}

\section{Problems with Inner-Consistency Property - Proofs}\label{sec:innerConsistency}

  \begin{proof} of Theorem  \ref{lem:noinnerconsistency2dim4clusters}. 
  \label{lem:noinnerconsistency2dim4clustersPROOF}
Let 
$\mathcal{E}_1$ and   $\mathcal{E}_2$ be two embedding{}s such that the second is an inner-$\Gamma$-transform of the first one. 
In $\mathbb{R}^m$,  the position of a point is uniquely defined by distances from $m+1$  distinct non-cohyperplanar points with fixed positions. 
Assume that the inner-$\Gamma$-transform moves closer points in cluster $C_0$, when switching from embedding $\mathcal{E}_1$ to $\mathcal{E}_2$.  
So pick $m+1$ point embedding{}s $\mathbf{p_1},\dots,\mathbf{p_{m+1}}$  from any $m+1$ other different clusters $C_1,\dots,C_{m+1}$.  
The distances between these $m+1$ points are fixed. So let their positions, without any decrease in generality, be the same under both  $\mathcal{E}_1, \mathcal{E}_2$.
Now the distances of any point $\mathbf{p_Z}$ in the  cluster $C_0$ to any of these selected points  cannot be changed under the transform. Hence the positions of points of (the embedding of) the first cluster are fixed, no non-trivial  inner-$\Gamma$-transformation applies.  
\end{proof}

\begin{proof} of Theorem \ref{thm:noinnerConsistency}
\label{thm:noinnerConsistencyPROOF}
The $m+1$ data points from different clusters have to be rigid under inner-$\Gamma$-transformation. 
Any point from outside of this set is uniquely determined in space given the distances to these points. 
So for any two points from clusters not belonging to the selected $m+1$ clusters no distance change is possible. 
So assume that these two clusters, containing at least two points each, are among those $m+1$ selected. 
If a point different from the selected points shall become closer to the selected point after inner-$\Gamma$-transformation, then it has to be on the other side of a hyper{}plane formed by $m$ points than the $m+1$st point of the same cluster before transformation and on the same side after the transformation. 
But such an effect is possible for one hyper{}plane only and not for two because the distances between the other points will be violated.

\end{proof}

\begin{figure}
\centering
\includegraphics[width=0.4\textwidth]{\figaddr{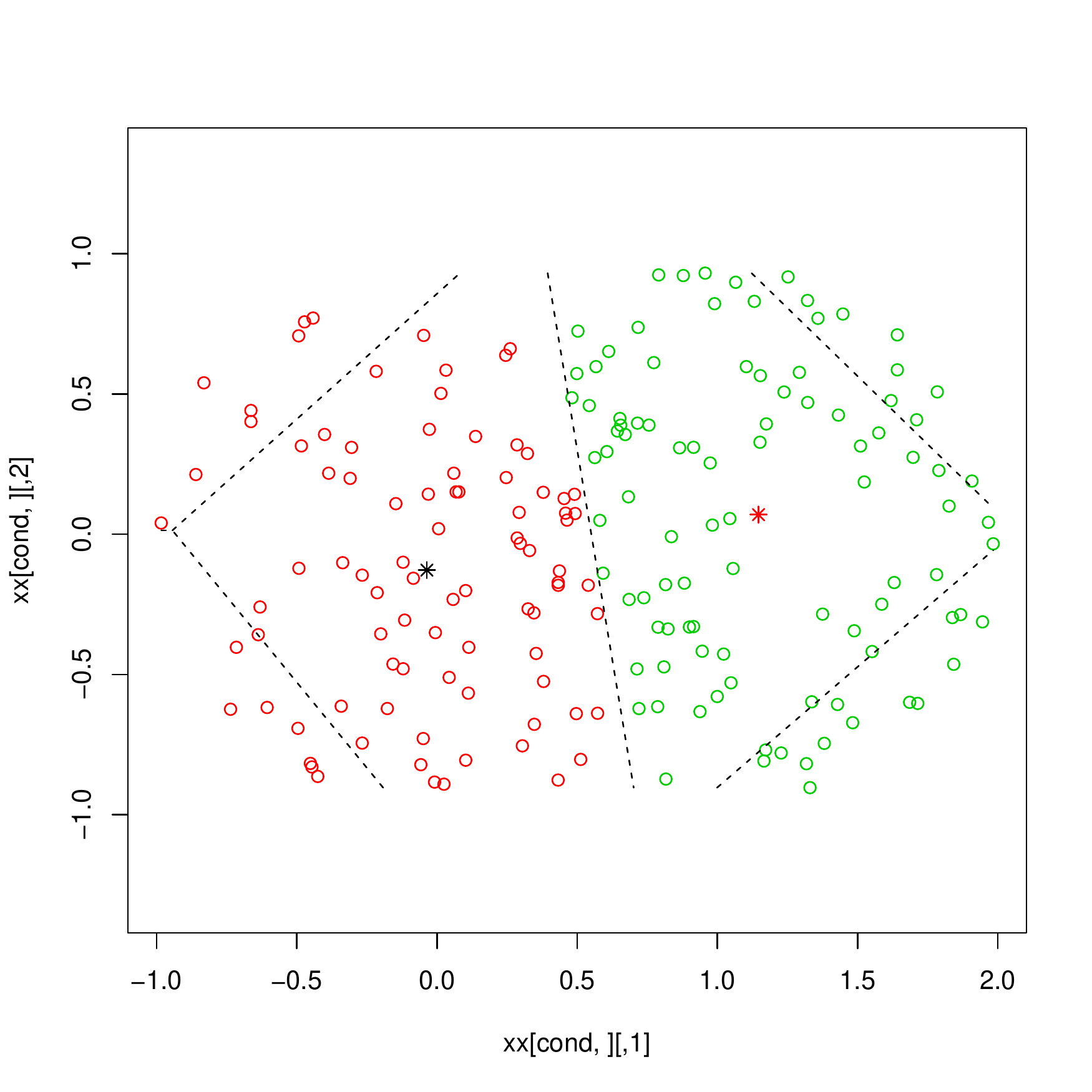}}  %
\includegraphics[width=0.4\textwidth]{\figaddr{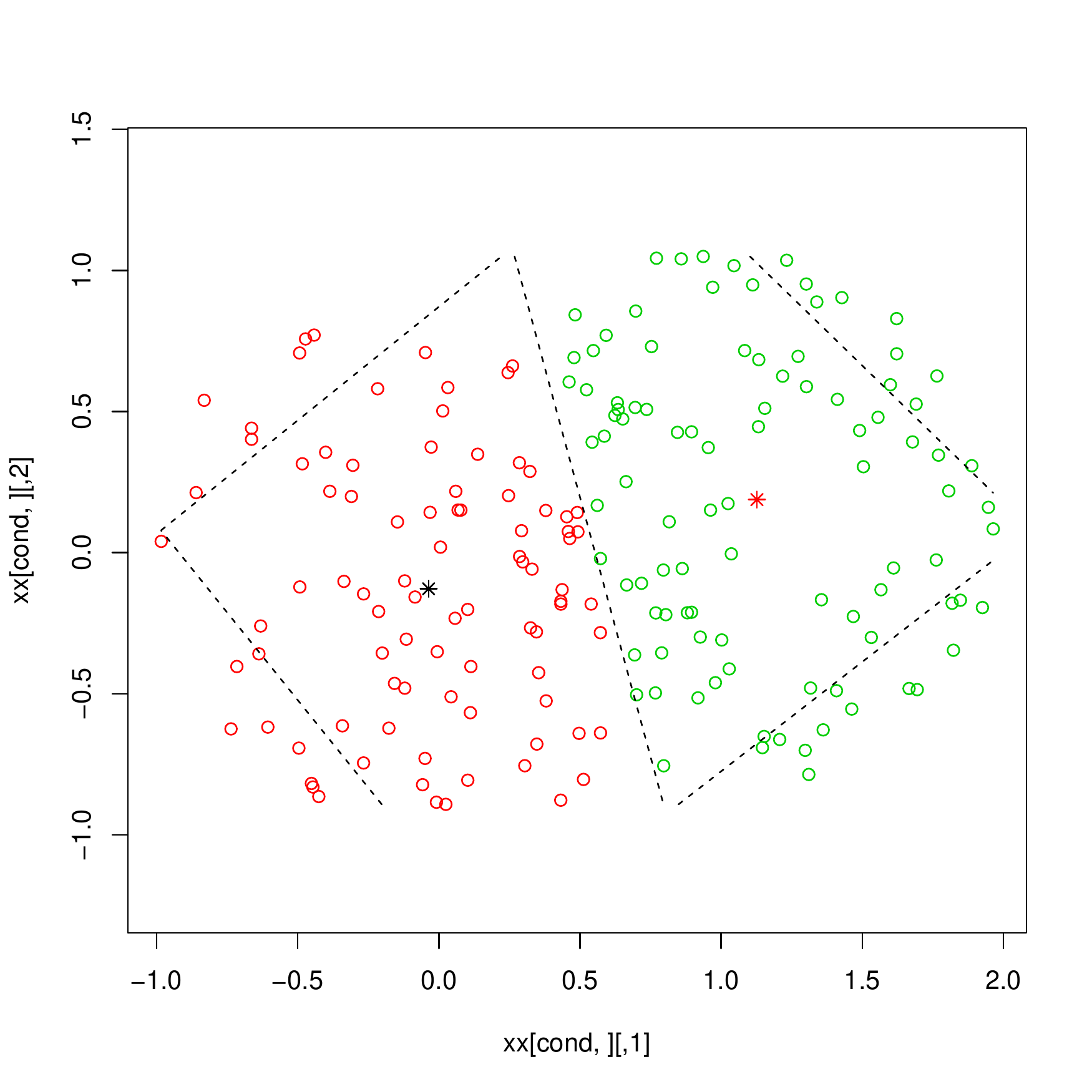}}  %
\caption{Effect of outer-$\Gamma$-transform if moving not orthogonally to cluster border. Left - the original data, clustered by 2-means.  
Right - effect of moving by a vector parallel to cluster border.
}\label{fig:danerazdwa}
\end{figure}

\section{Problems with Continuous Outer   Consistency Property - Proofs}\label{sec:outerConsistency}



\begin{proof} of Theorem \ref{thm:noGeneralConsistency}
\label{thm:noGeneralConsistencyPROOF}. 
2D case proof 

Consider two clusters, calling them  a green one and a blue one. By definition,  one of them must have points outside of the convex hull of a part of the other. For an illustrative example see Figure \ref{fig:impossibleConsistency}.   

There exists
the line segment connecting two green points $B,C$  that intersects the line segment connecting blue points $X,Y$ not at the endpoints of any of these segments. The existence is granted e.g. by choosing one green  point inside of blue convex hull and the other outside. 
For any point $A$ let $\vec{A}$ denote the vector from coordinate system origin to the point $A$. 
 Let the positions of these points after $\Gamma$-transformation be $B', C', X', Y'$ resp. and let the respective line segments $\overline{B'C'},\overline{X'Y'}$ interiors intersect at the point $M'$, with    $\vec{M'}=
\omega \vec{X'}+(1-\omega) \vec{Y'}$. 
for some $0<\omega<1$. 
(With continuous transformation this is always possible).
For this $\omega$ define the point $M$ such that $\vec{M}=\omega \vec{X} (1-\omega) \vec{Y}$. Obviously it will belong to the interior of the line segment $\overline{XY}$. 
The elementary geometry tells us that 
$\cos(\measuredangle YXB) = \frac{|YX|^2+|XB|^2 - |YB|^2}{2|YX||XB|}$.
and at the same time
$|MB|^2=|XM|^2+|XB|^2-2\cos(\measuredangle YXB)|XM||XB| $ because $\measuredangle YXB=\measuredangle MXB$. 
Hence
$$|MB|^2
=|XM|^2+|XB|^2-2|XM||XB| \frac{|YX|^2+|XB|^2 - |YB|^2}{2|YX||XB|}
$$ $$=\omega^2|XY|^2+|XB|^2-2\omega^2|XY||XB| \frac{|YX|^2+|XB|^2 - |YB|^2}{2|YX||XB|}
$$ $$
=(1-  \omega)|XB|^2+\omega   |YB|^2  -(1-\omega) \omega |XY|^2
$$

This result makes it clear that under Klein{}berg's $\Gamma$-transformation $MB$ will increase or stay the same, as $|XB|$ and $|YB|$ increase or remain the same while $|XY|$ decreases or is unchanged. 
By definition of Klein{}berg's consistency:
$|BC|\ge |B'C'|$, 
$|XY|\ge |X'Y'|$, 
$|BX|\le |B'X'|$, 
$|CX|\le |C'X'|$, 
$|BY|\le |B'Y'|$, 
$|CY|\le |C'Y'|$.
Now recall that 
$|BC|\le |BM|+|MC|$ and 
$|B'C'|= |B'M'|+|M'C'|$. We have already shown that if $|XY|>|X'Y'$ or $|BX|<|B'X'|$or $|BY|<|B'Y'|$, then $|BM|<|B'M'|$, and 
 if $|XY|>|X'Y'$ or $|CY|<|C'Y'|$or $|CX|<|C'X'|$, then $|MC|<|M'C'|$ and hence 
 $|BC|<|B'C'|$. This would be, however, a contradiction which implies that $|BC|=|B'C'|$, $|XY|=|X'Y'|$, 
 $|BX|= |B'X'|$, 
$|CX|= |C'X'|$, 
$|BY|= |B'Y'|$, 
$|CY|= |C'Y'|$.

\begin{figure}
\centering
\includegraphics[width=0.4\textwidth]{\figaddr{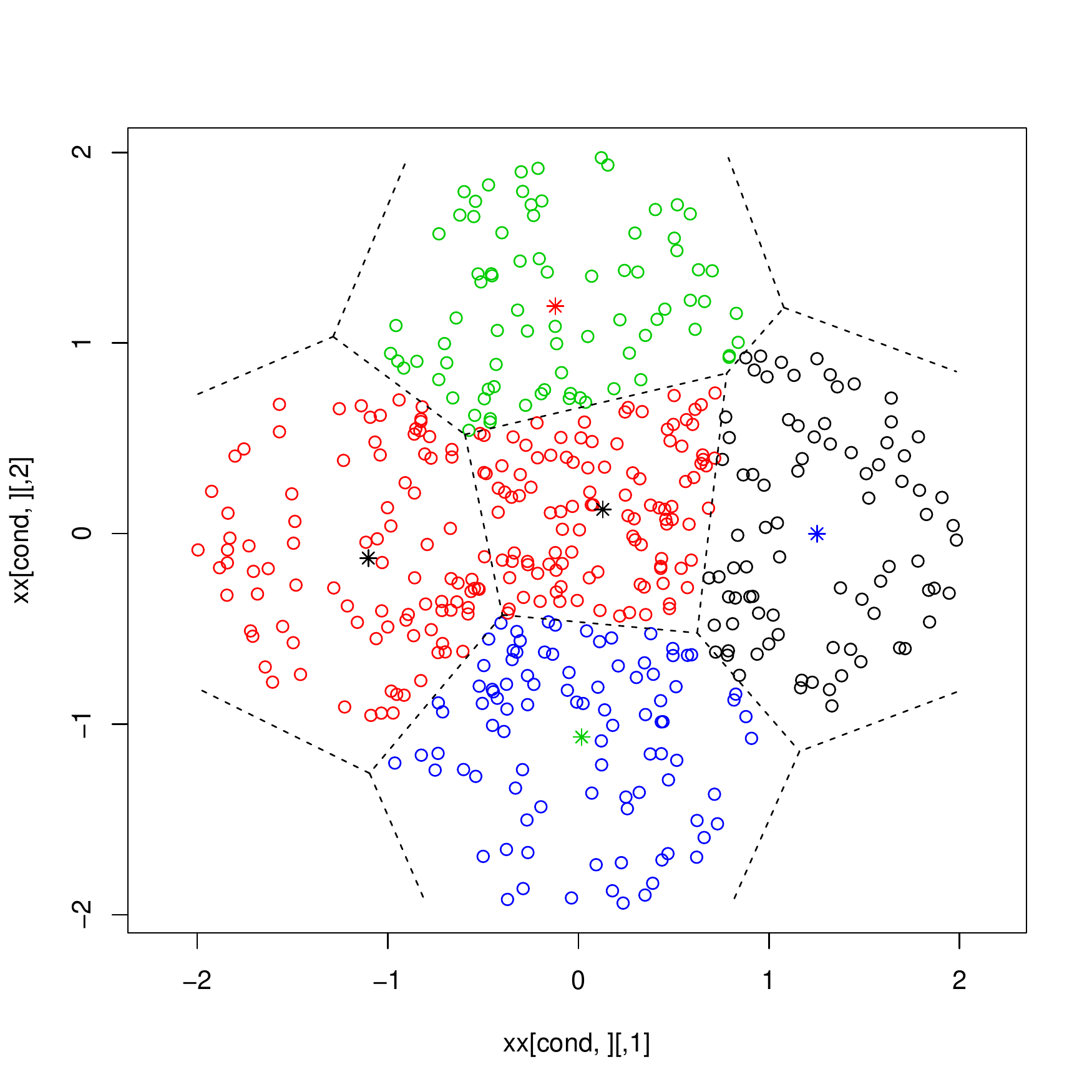}}  %
\includegraphics[width=0.4\textwidth]{\figaddr{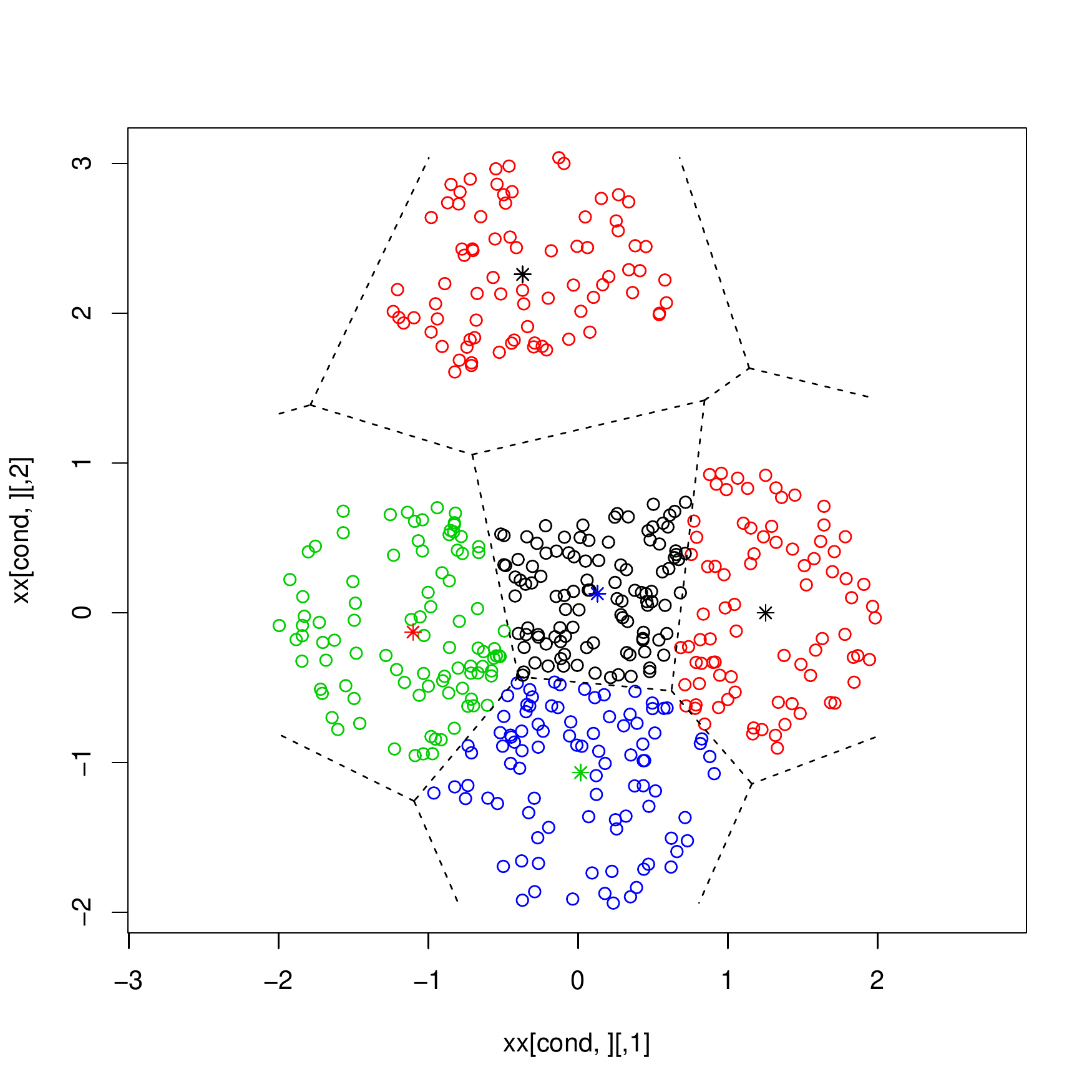}}  %
\caption{Effect of outer-$\Gamma$-transform if moving not orthogonally to all cluster borders. Left - the original data, clustered by 5-means. 
Right - effect of moving the top cluster by a vector orthogonal to only one cluster border. 
}\label{fig:danejedenpievcv}
\end{figure}

Now for any green data{}point $A$ such that there exists a blue point $Z$ (not necessarily different from $Y$) and the  line segment $\overline{XZ}$ intersects in the interior with  $\overline{BA}$, we can repeat the reasoning and state that the distances to $B,C,X,Y$ will remain unchanged. Obviously, for all points $A$ from outside of the blue convex hull it would be the case (a blue line segment must separate $B$ from any green point outside of the convex hull). Any point inside it will either lie on the other side of the straight line $XY$ than $B$ or $C$. If the respective segments intersect, we are done. If not, the green line segment passes through blue area either "behind" $X$ or $Y$. So there exists still another blue line segment intersecting with it.  So the green data set is rigid. By the same argument the blue data set is also rigid - and the distances between them are also rigid, so that all data points are rigid under Klein{}berg's $\Gamma$-transformation. 

This may be easily generalized to $n$-dimensional space.
%
 %
\end{proof}

\begin{proof} of Theorem 
\ref{thm:noGeneralConvexOuterConsistency}
\label{thm:noGeneralConvexOuterConsistencyPROOF}.
It is necessary that the relative motion direction is orthogonal to a face separating two clusters (see Section \ref{sec:experimental}) . If no such face exists, then there exists such a face orthogonal to the relative speed vector that members of both clusters are on each side of this face. Hence a motion causes decrease of distance between some elements of different clusters.

Consider now a loop $i_1,i_2,...,i_n$ of cluster indices   that  is an index sequence  such that $i_1=i_n $. We want that continuous outer-$\Gamma$-transform is applicable to them. 
So we need to determine points $S_{i_1}, S_{i_2},S_{i_3}$ following the above-mentioned principle. Then, when dealing with  clusters $C_{i_k} , C_{i_{k+1}} C_{i_{k+2}} $, the points    $S_{i_k} , S_{i_{k+1}}$ are predefined, and the point $S_{i_{k+2}} $ will be automatically defined, if it exists.  
If it turns out that $S_{i_1}\ne S_{i_n} $ then we have a problem because the vector of relative speed of $C_{i_{n-1}}$ with respect to $C_{i_1}$ is determined to be $\mathbf{v}_{i_1,i_n}=\mathbf{v}_{i_1,i_2}+\mathbf{v}_{i_2,i_3}+...+\mathbf{v}_{i_{n-1},i_n} $ would not be zero which is a contradiction - a cluster has zero relative speed to itself.

The proof follows directly from the indicated problem of non-zero relative speed of the first cluster with respect to itself. 
\end{proof}

For an example in 2D  
look at the  Figure 
\ref{fig:impossibleConsistency} to the right for an explanation, why each cluster has to have its own speed. 
Assume that we want to move only one cluster, that is $FACH$ with speed orthogonal to the edge $AC$. Imagine a point $X$  in the cluster $FABG$ close to $F$ and a point $Y$ in cluster $FACH$ close to $A$. When moving  the cluster $FACH$
 away from $ABC$, all the ponts of $FACH$ will increase their distances to all points of $ABC$, but the points $X$ and $Y$ will get closer, because the relative speed of $FACB$ with respect to $FABG$ is not orthogonal to the edge $FA$.
 So, 
in a plane,  
the differences between the speed vectors of clusters sharing an edge must be orthogonal to that edge (or approximately orthogonal, depending on the size of the gap).
The Figure 
\ref{fig:impossibleConsistency} to the right illustrates the problem. In this figure the direction angles of the lines $AF,BG,CH$ were deliberately chosen in such a way that it is impossible.(For the $\sin$ values of the respective angles see below).  The choice was as follows: 
Assume the speed of cluster $ACHF$ with respect to cluster ABC is fixed to 1. 
Then the speed of ABGF has to be  0.26794\Bem{9192431126} 
because sin(F,A,C)= 0.20048\Bem{0370272677}  and sin(F,A,B)= 0.74820\Bem{292777784} .
Therefore the speed of $BCHG$  has to be  0.26794\Bem{9192431126} 
because sin(G,B,A)= 0.5  and sin(G,B,C)= 0.5.
Therefore the speed of $ACHF$  has to be  0.15311\Bem{3824246358} 
because sin(H,C,B)= 0.35921\Bem{0604053549}  and sin(H,C,A) 0.62861\Bem{8557093712}.
But this is a contradiction because the speed of $ACHF$ was assumed to be  1  which differs from  0.15311\Bem{3824246358}.


\begin{table}
\caption{Number of data points from different clusters that the distance was reduced during moving one cluster. The move was by 0.1 of the distance between the cluster centers in the direction indicated by the rotation.}
\label{tab:Xdanerazdwa}
\begin{tabular}{|l|r|r|r|r|r|r|r|}
\hline rotation & -90$^o$ & -60$^o$ & -30$^o$ & 0$^o$ & 30$^o$ & 60$^o$ & 90\\ \hline 
\hline bad distances & 8064 & 3166 & 480 & 0 & 296 & 2956 & 7920\\ \hline 
\hline percentage of bad distances & 23.6\% & 9.3\% & 1.4\% & 0\% & 0.9\% & 8.6\% & 23.1 \%\\ \hline 
\end{tabular}
\end{table}

  \section{Experimental Illustration of Outer Consistency Impossibility Theorem \ref{thm:noGeneralConvexOuterConsistency}}\label{sec:experimental}

 In order to illustrate the need to move clusters perpendicularly to cluster border between clusters in case of continuous outer-$\Gamma$-transformation, compare e.g. Theorem \ref{thm:noGeneralConvexOuterConsistency}, we generated randomly uniformly on a circle small clusters of about 100 data points each in 2D. In the first experiment 2 clusters and in the second 5 clusters were considered, as shown in Figs \ref{fig:danerazdwa} and \ref{fig:danejedenpievcv} left. In the first experiment one of the clusters was moved in different directions from its original position. The more the motion deviated from the perpendicular direction, the move violations of the $\Gamma$-transform requirements were observed, as visible in table  \ref{tab:Xdanerazdwa}. 
  In the second experiment one of the clusters was moved orthogonally to one of its orders, but necessarily not orthogonal to other borders. The number of distance violations was the largest for the smallest motions as visible in table \ref{tab:Xdanejedenpievcv}. As soon as the data points of the cluster got out of the convex hall of other clusters, the number of violations dropped to zero. 
  
  \Bem{ 
A third experiment shows the advantage of the gravitational consistency as performed using homothetic transformation, see Theorem \ref{thm:gravcons}. 
[TODO]
}

\section{The Concept of Gravitational  Consistency Property - Proofs }\label{sec:gravitConsistency}

The problem with $k$-means in Klein{}berg's counter example on consistency relies on the fact that $k$-means is a centric algorithm and decrease in distances between data points may cause the distance to the cluster center to be increased. 
It is easily shown that this is not a problem between clusters - upon consistency transformation cluster centers become more distant. The problem is within a cluster. Therefore we propose to add an additional constraint (gravity center constraint)  on consistency that is that it is forbidden to increase the distance not only between data points but also between gravity centers of any disjoint subsets of a cluster.

\begin{proof}
of Theorem \ref{thm:gravcons}
\label{thm:gravconsPROOF}.
The centric sum of squares, $CSS(C)$
of the data set $C$ may be expressed in two ways:
$$CSS(C)=\sum_{\mathbf{x}_i\in C} ||\mathbf{x}_i \mu(C)||^2
= \frac{1/2}{|C|}\sum_{\mathbf{x}_i\in C} \sum_{\mathbf{x}_j\in C} ||\mathbf{x}_i-\mathbf{x}_j||^2 
$$ $$= \frac{1}{|C|}\sum_{\{\mathbf{x}_i, \mathbf{x}_j\}\subset C, \mathbf{x}_i\ne \mathbf{x}_j }  ||\mathbf{x}_i-\mathbf{x}_j||^2 $$
 where $\mu(c)=\frac{1}{|C|}\sum_{\mathbf{x}_i\in C}  \mathbf{x}_i  $.
 Obviously, for a partition $\mathfrak{C}$, $\sum_{ C \in \mathfrak{C}} CSS(C)$  is the $k$-means criterion function. 
 There exists, however, a different way to express it. 
 Let $\mathfrak{M}(C)$ be an operator replacing each data point in $C$ with $\mu(C)$ (forming a multiset). 
 Let $\mathfrak{C}(C)$ be a partition of $c$ (into disjoint sets). 
 Then 
 $CSS(C)=\sum_{C_i \in \mathfrak{C}(C)} CSS(C_i)+ CSS(   \cup_{C_j \in \mathfrak{C}(C)} \mathfrak{M}(C_j))   $.
 Note that the latter is just the weighted sum of distances between   centers of clusters from $\mathfrak{C}(C)$.
 Assume now we have a partition $\mathfrak{C}_0$ being the optimal one under $k$-means and a competing one (but not optimal)  $\mathfrak{C}_1$. 
 Construct a new partition  (possibly into much more than $k$ clusters) 
 $\mathfrak{C}_{10}=\{C; C=C_i \cap C_j, C_i \in \mathfrak{C}_0,C_j \in \mathfrak{C}_j \}$
 The cost function of both $\mathfrak{C}_0, \mathfrak{C}_1$ can be expressed as a sum of $CSS$ of all elements of $\mathfrak{C}_{10}$ plus connections between gravity centers of 
 $\mathfrak{C}_{10}$ lying inside of a $\mathfrak{C}_0$ cluster or outside. 
 By the virtue of the property of gravity center constraint the latter connections will decrease under consistency operation, and due to properties of the third form of $CSS$ the distances between clusters of $\mathfrak{C}_{10}$ contained in different clusters of $\mathfrak{C}_{0}$ will increase. Just look at the difference $CSS(A\cup B)-CSS(A)-CSS(B)$ -this difference, computed as sum of distances between data points will increase so the distance between gravity centers will increase too, by the equivalence of first two methods of computing $CSS$.
\end{proof}

\begin{proof} of Theorem \ref{thm:homotheticgravcons} 
\label{thm:homotheticgravconsPROOF}.
The proof follows from elementary geometry. 
\end{proof}
When homothetic transformations were applied to various clusters, the distances between them need to be increased by the biggest distance decrease in order for the transformation to be a $\Gamma$-transformation in the spirit of Kleinberg. 

\begin{table} 
\caption{Number of data points from different clusters that the distance was reduced during moving one cluster (hence violating $\Gamma$-transform condition). The top cluster was moved  away from the central cluster on the distance of $s$ times the distance between cluster centers.} \label{tab:Xdanejedenpievcv}
\begin{tabular}{|l|r|r|r|r|r|r|r|r|r|r|r|}
\hline shift $s$ & 0 & 0.1 & 0.2 & 0.3 & 0.4 & 0.5 & 0.6 & 0.7 & 0.8 & 0.9 & 1\\ \hline 
\hline bad dst  & 0 & 392 & 280 & 158 & 62 & 26 & 6 & 0 & 0 & 0 & 0\\ \hline 
\hline \% bad dst & 0\% & 0.2\% & 0.1\% & 0.1\% & 0\% & 0\% & 0\% & 0\% & 0\% & 0\% & 0 \%\\ \hline 
\end{tabular}
\end{table}


\newcommand{\CutEq}[1]{}

\section{$k$-means and the  centric-consistency axiom}
\label{sec:kmeanscentricconsistent}

\begin{figure}
\centering
\includegraphics[width=0.4\textwidth]{\figaddr{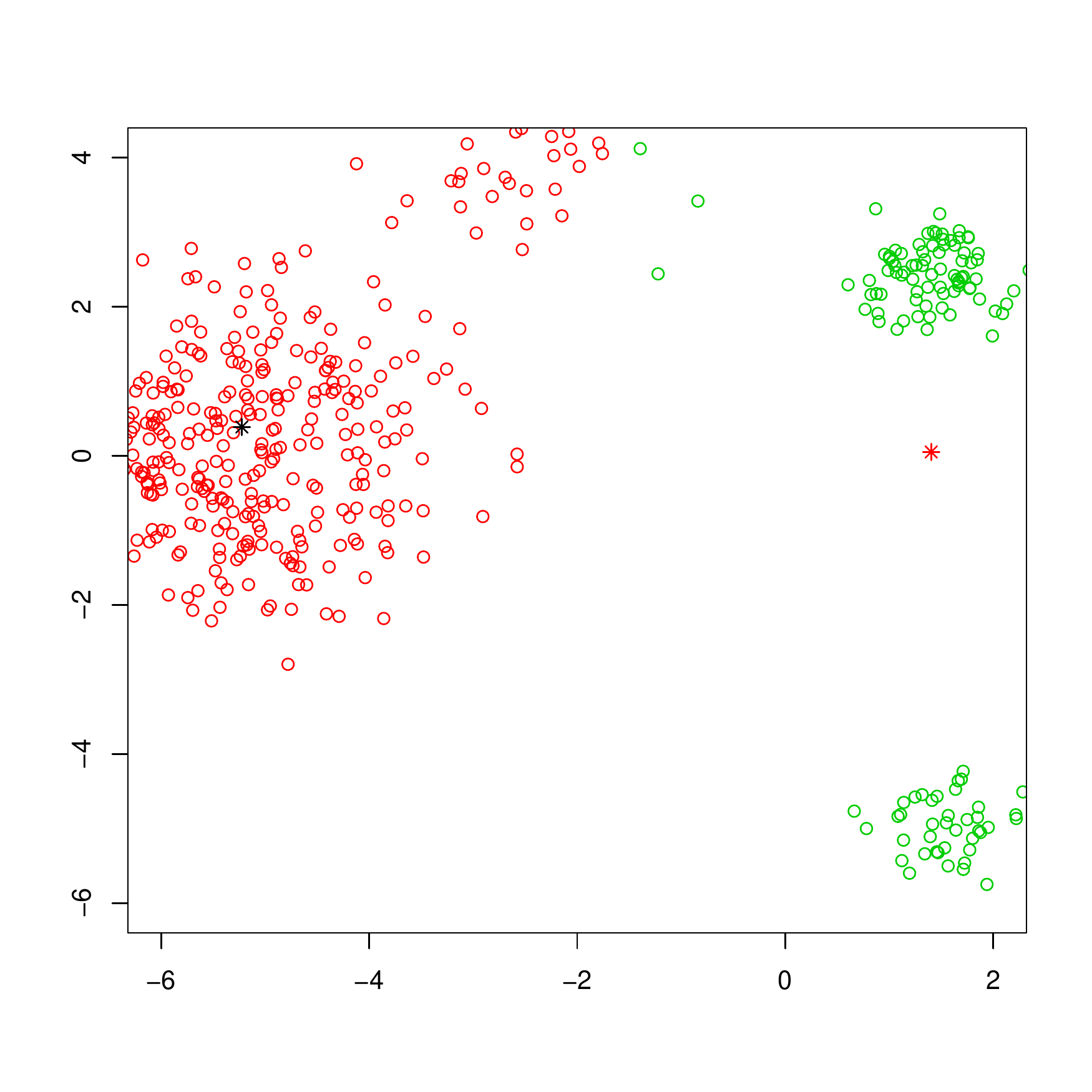}}  %
\includegraphics[width=0.4\textwidth]{\figaddr{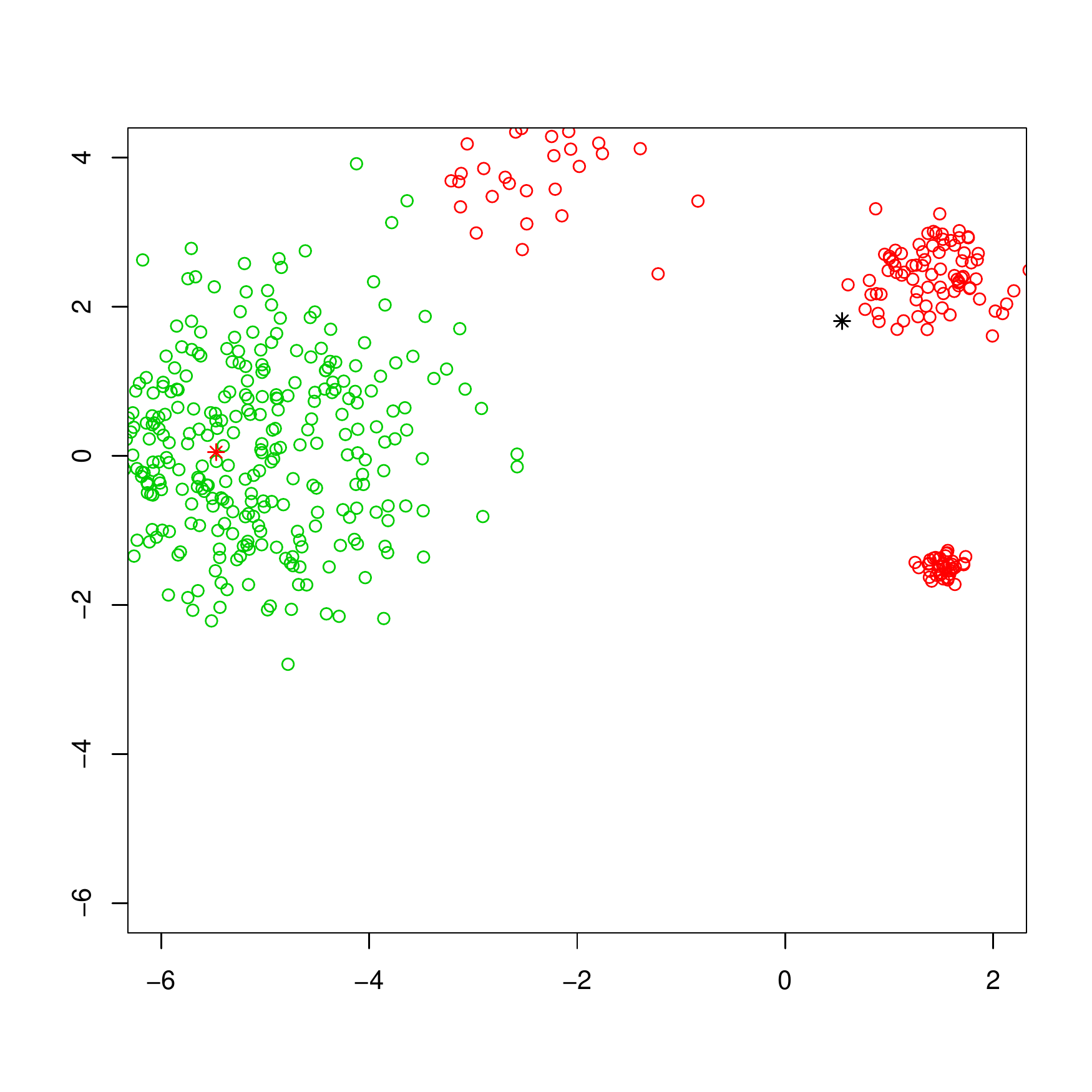}}  %
\caption{Getting some data points closer to the cluster center does not ensure cluster stability for $k$-means, because the cluster center can move. 
Left picture - data partition  before moving data closer to the cluster center. 
Right picture - data partition  thereafter. 
}\label{fig:SomeGettingCloser}
\end{figure}

\begin{figure}
\centering
\includegraphics[width=0.4\textwidth]{\figaddr{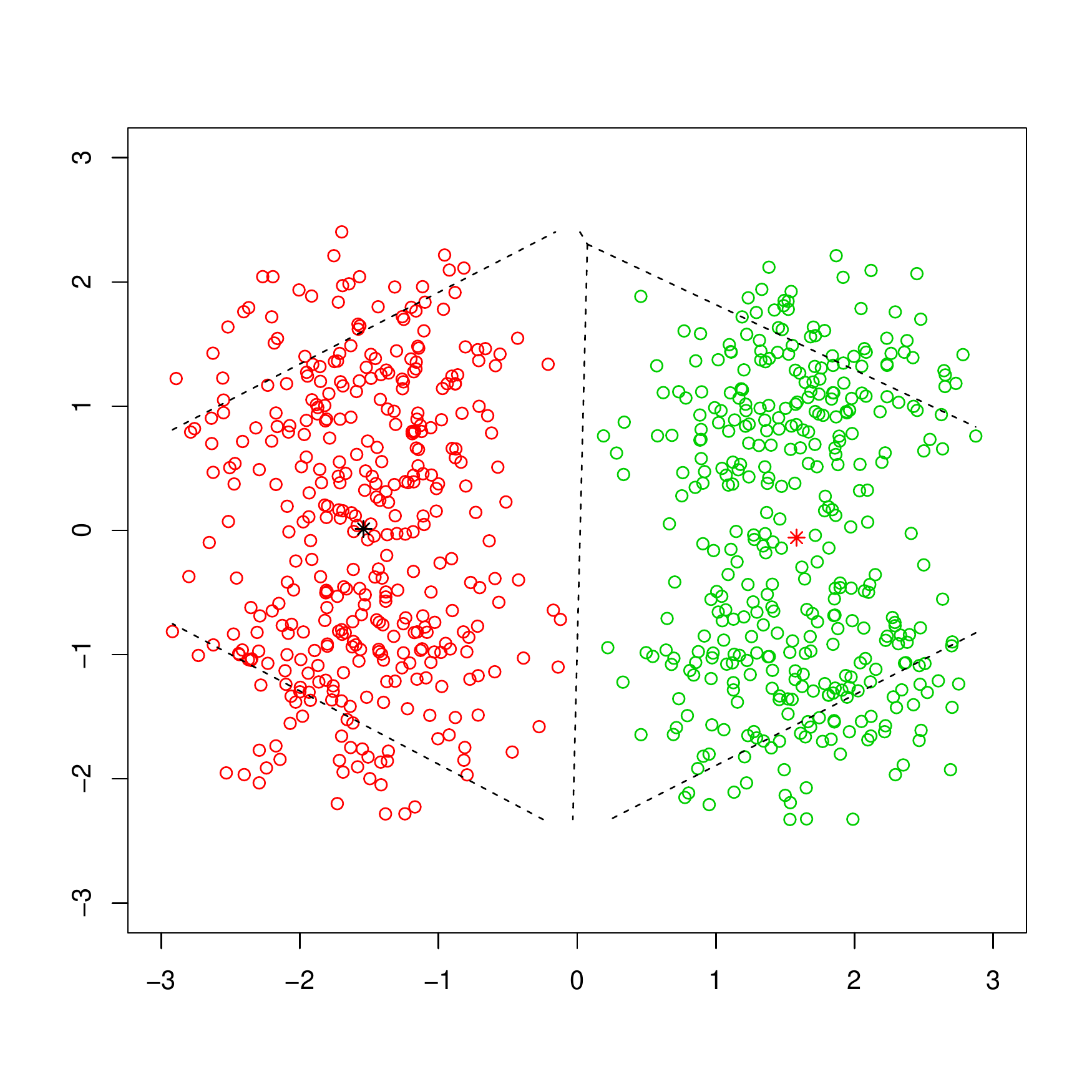}}  %
\includegraphics[width=0.4\textwidth]{\figaddr{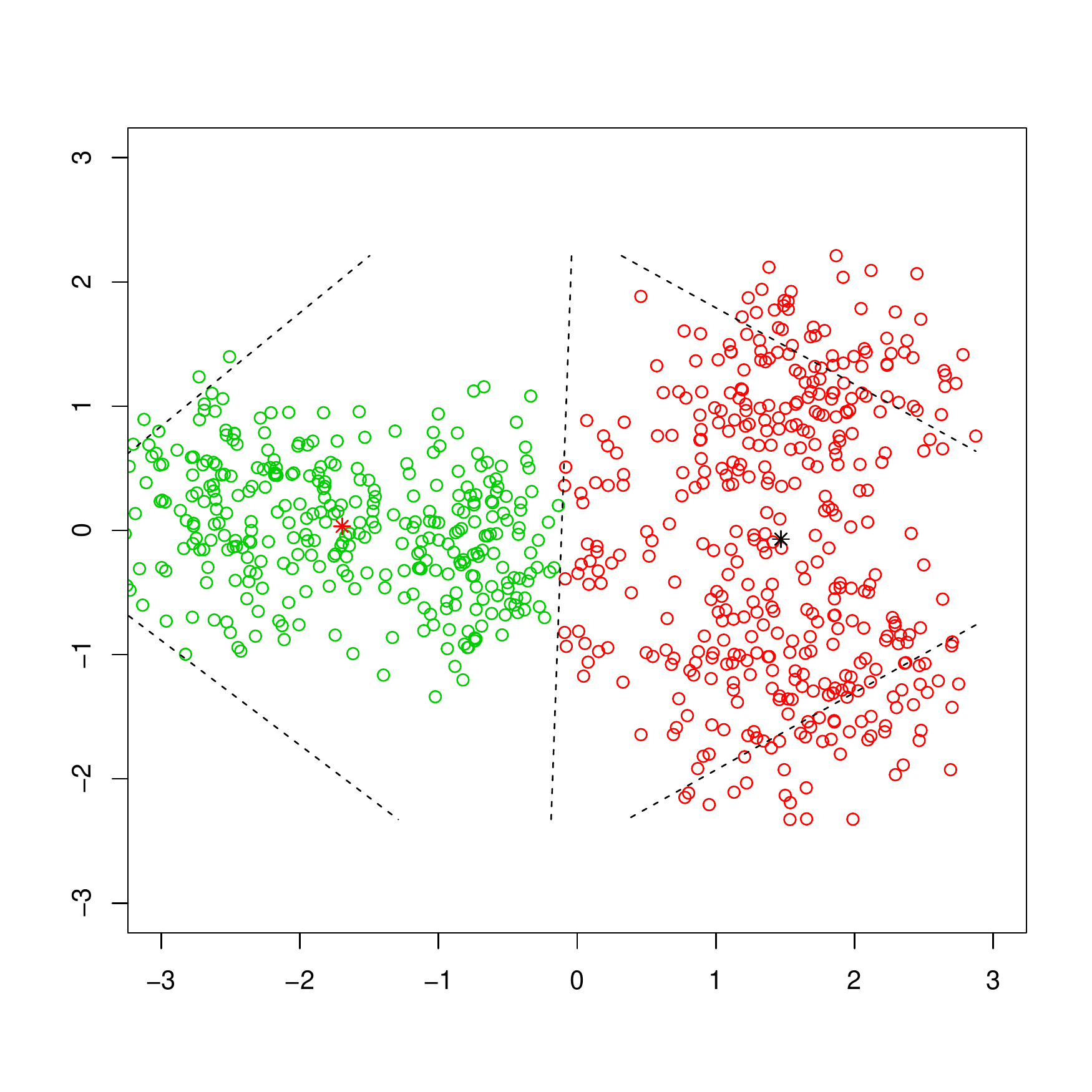}}  %
\caption{Getting all data points  closer to the cluster center without changing the position of cluster center, if we do not ensure that they move along the line connecting each with the center.
Left picture - data partition  before moving data closer to the cluster center. 
Right picture - data partition  thereafter. 
}\label{fig:AllGettingCloser}
\end{figure}

\begin{figure}
\centering
\includegraphics[width=0.8\textwidth]{\figaddr{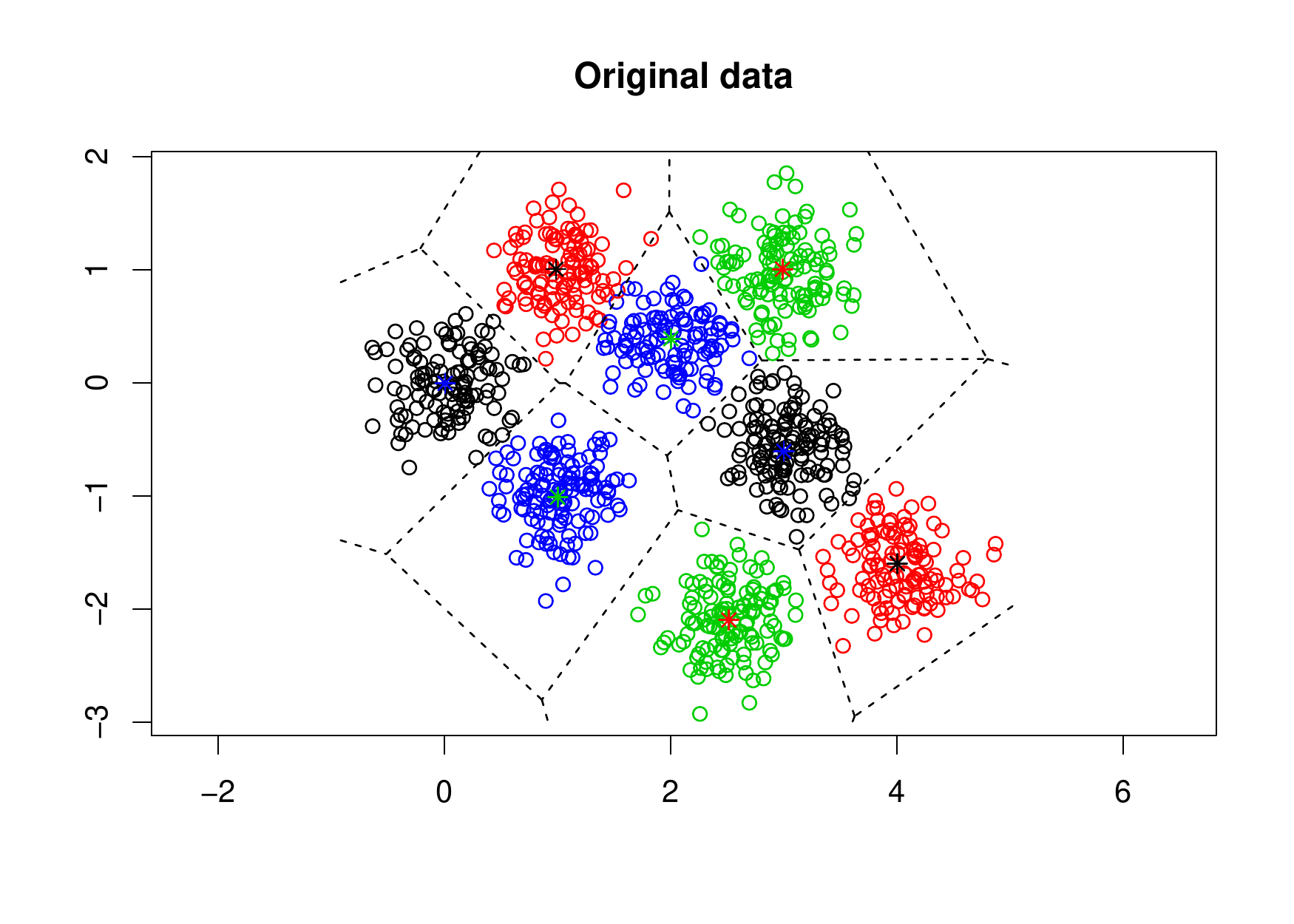}}  %
\caption{A mixture of 8 normal distributions as clustered by $k$-means algorithm (Voronoi diagram superimposed).  
}\label{fig:8clusters}
\end{figure}

\begin{figure}
\centering
\includegraphics[width=0.8\textwidth]{\figaddr{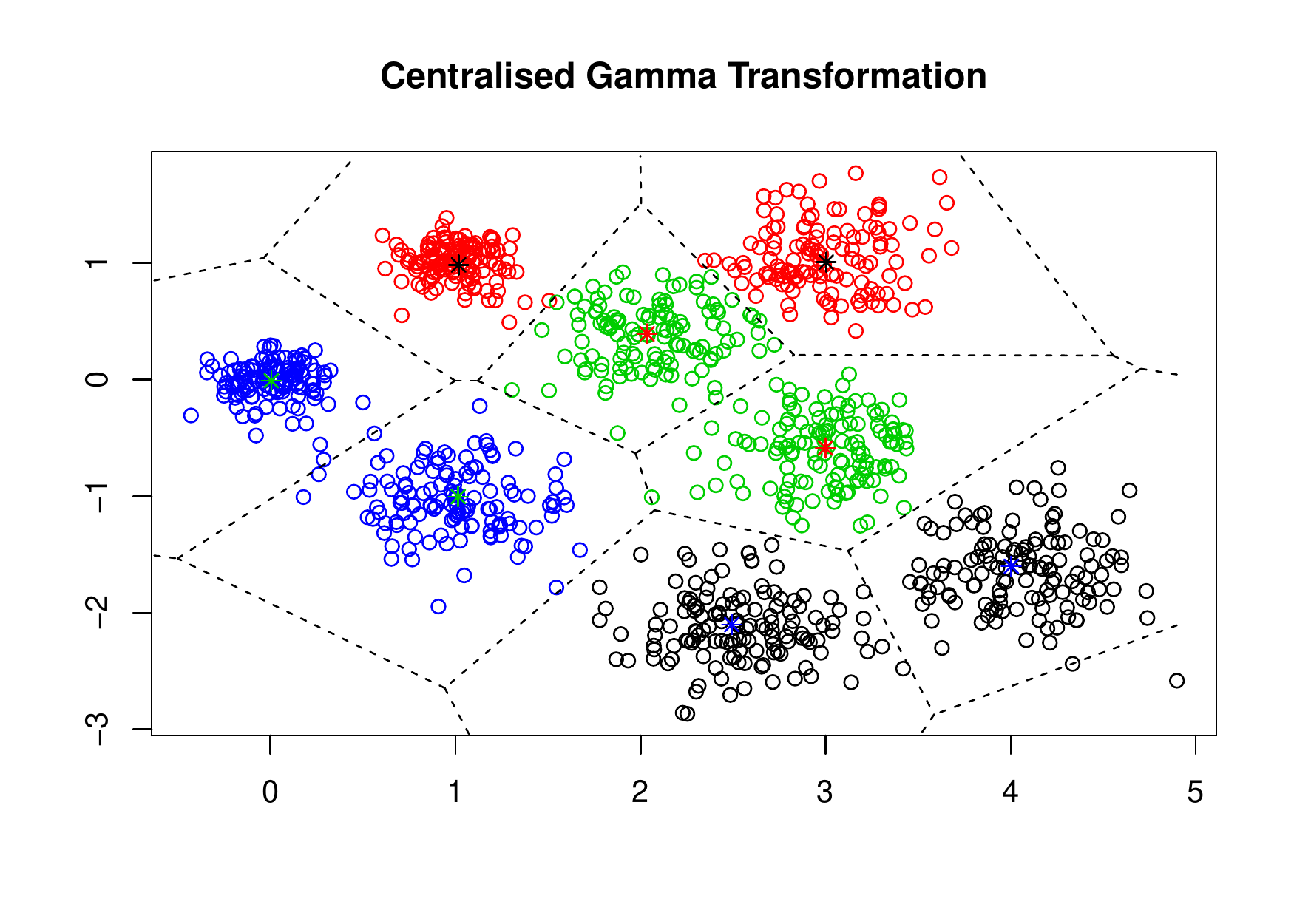}}  %
\caption{Data from 
Figure \ref{fig:8clusters} after a centralized  $\Gamma$-transformation ($\Gamma^*$ transformation),
 clustered by $k$-means algorithm  into 8 groups
}\label{fig:8clustersKlopotek}
\end{figure}

The first differentiating feature of the centric consistency is that no new structures are introduced in the cluster at any scale.
The second important feature is that the requirement of keeping the minimum distance to elements of other clusters is dropped and only cluster centers do not get closer to one another. 

Note also that the centric consistency does not suffer from the impossibility of transformation for clusters that turn out to be internal.

\Bem{
Our proposal of centric-consistency   has a practical background. 
Kleinberg proved that $k$-means does not fit his consistency axiom. 
As shown experimentally in Table \ref{tab:comparisonVariance}, $k$-means algorithm behaves properly under $\Gamma^*$ transformation. 
Figure \ref{fig:8clustersKlopotek} illustrates a two-fold application of the 
$\Gamma^*$ transform (same clusters affected as by $\Gamma$-transform in the preceding figure).
As recognizable visually and by inspecting the forth column 
of  Table \ref{tab:comparisonVariance} and  Table \ref{tab:comparisonVarianceGain}, here $k=8$ is the best choice for $k$-means algorithm, so the centric-consistency axiom is followed. 
}%

\begin{figure}
\centering
\includegraphics[width=0.45\textwidth]{\figaddr{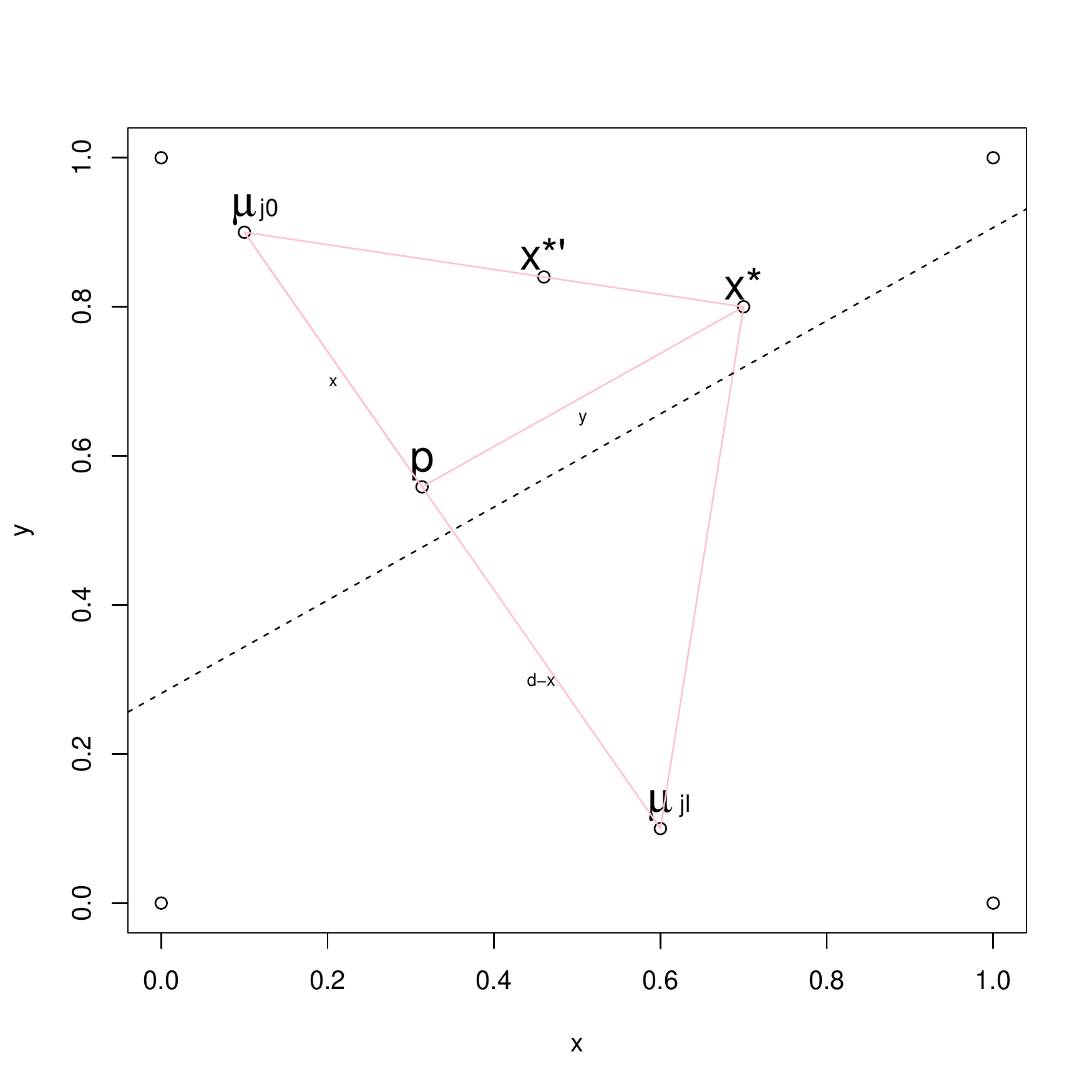}}  %
\includegraphics[width=0.45\textwidth]{\figaddr{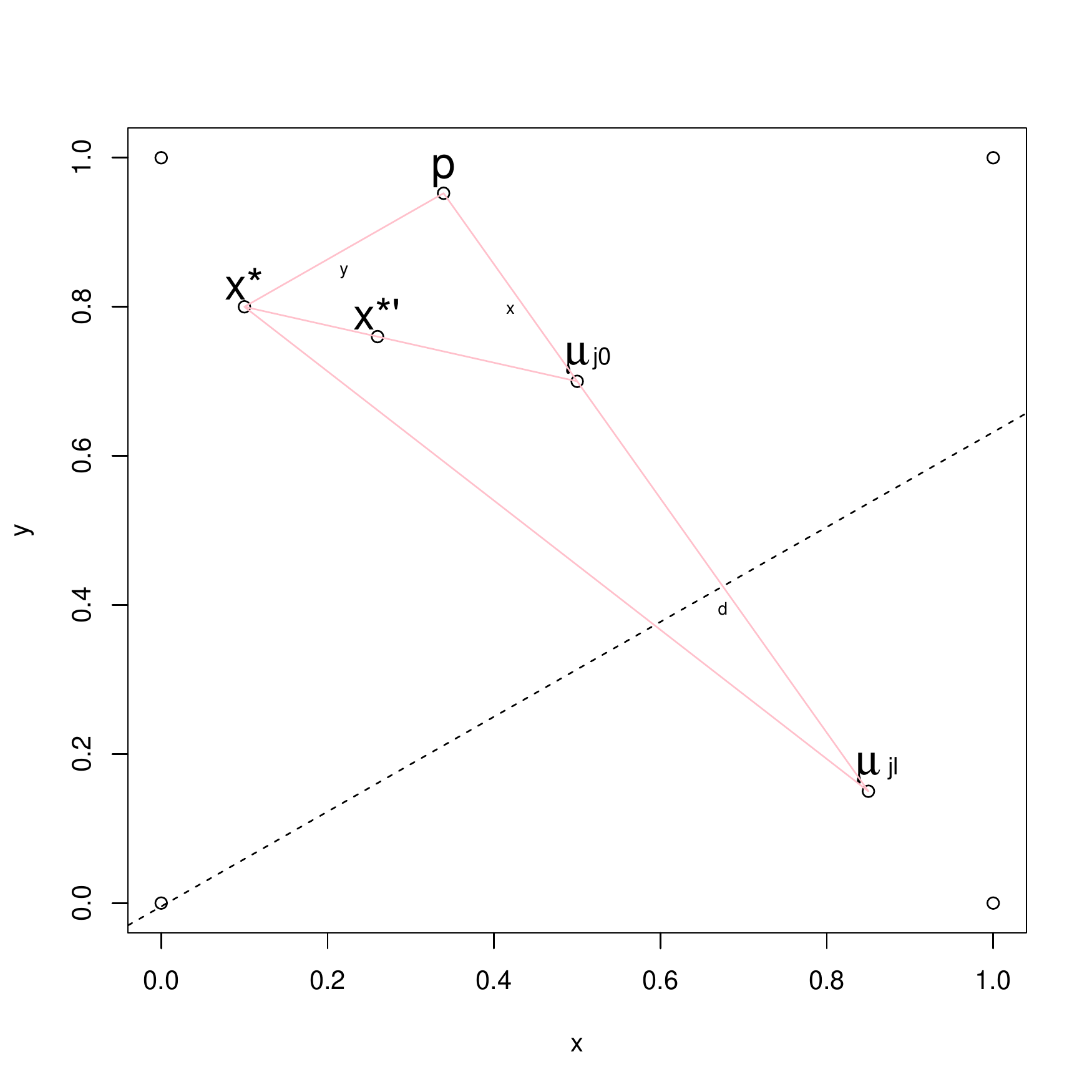}}  %
\caption{Impact of contraction 
of the data point
$\textbf{x}^*$ 
towards cluster center $\boldsymbol{\mu}_{j_0}$ by a factor $\lambda$ 
to the new location $\textbf{x}^{*'}$ 
- local optimum maintained.   
The left image illustrates the situation when 
the point $\textbf{x}^*$  is closer to cluster center 
 $\boldsymbol{\mu}_{j_0}$. The right image refers to the inverse situation. 
The   point $\textbf{p}$ is the orthogonal projection of  the point
$\textbf{x}^*$ onto the line $\boldsymbol{\mu}_{j_0}, \boldsymbol{\mu}_{j_l}$. 
}\label{fig:XBETWEENOUTSIDE}
\end{figure}

\begin{proof} of Theorem 
\ref{thm:localCentricConsistencyForKmeans} 
\label{thm:localCentricConsistencyForKmeansPROOF}

$V(C_j)$ be the sum of squares of distances of all objects of the cluster   $C_j$ from its gravity center and $Q$ be from equation (\ref{eq:Q::kmeans}). . 
Hence $Q(\Gamma)= \sum_{j=1}^k \frac{1}{n_j} V(C_j)$. 
Consider moving a data point $\textbf{x}^*$  from the cluster $C_{j_0}$ 
to cluster $C_{j_l}$ 
As demonstrated by \cite{Duda:1973},  
$V( C_{j_0} -\{\textbf{x}^*\})= V(C_{j_0}) -
  \frac{n_{j_0}}{n_{j_0}-1}\|\textbf{x}^*-\boldsymbol{\mu}_{j_0}\|^2$
and 
$V(C_{j_l}\cup\{\textbf{x}^*\})= V(C_{j_l}) +
 \frac{n_l}{n_l+1}\|\textbf{x}^*-\boldsymbol{\mu}_{j_l}\|^2 $
So it pays off to move a point from one cluster to another if
$\frac{n_{j_0}}{n_{j_0}-1}\|\textbf{x}^*-\boldsymbol{\mu}_{j_0}\|^2 > \frac{n_{j_l}}{n_{j_l}+1}\|\textbf{x}^*-\boldsymbol{\mu}_{j_l}\|^2$.
If we assume local optimality of $\Gamma$, 
this obviously did not pay off.
Now transform this data set to $\mathbf{X'}$ in that 
  we transform elements of cluster $C_{j_0}$ in such a way that 
it has now elements 
$\textbf{x}_i'=\textbf{x}_i+\lambda(\textbf{x}_i - \boldsymbol{\mu}_{j_0})$ for some $0<\lambda<1$,
see Figure \ref{fig:XBETWEENOUTSIDE}.
Consider a   partition $\Gamma'$
of $\mathbf{X'}$.  All clusters are the same as in $\Gamma$ except for 
the transformed elements that form now a cluster $C'_{j_0}$.
The question  is: 
does it pay off   to move a data point $\textbf{x'}^*\in C'_{j_0}$ between the clusters? 
Consider the plane containing 
$\textbf{x}^*, \boldsymbol{\mu}_{j_0}, \boldsymbol{\mu}_{j_l}$.
Project orthogonally the point
$\textbf{x}^*$ onto the line $\boldsymbol{\mu}_{j_0}, \boldsymbol{\mu}_{j_l}$, giving a point $\textbf{p}$.
Either $\textbf{p}$  lies  between 
$\boldsymbol{\mu}_{j_0}, \boldsymbol{\mu}_{j_l}$
or 
  $\boldsymbol{\mu}_{j_0}$  lies  between 
$\textbf{p}, \boldsymbol{\mu}_{j_l}$. Properties of $k$-means exclude other possibilities. 
Denote distances 
$y=\|\textbf{x}^*- \textbf{p}\|$,
$x=\|\boldsymbol{\mu}_{j_0}- \textbf{p}\|$,
$d=\|\boldsymbol{\mu}_{j_0}- \boldsymbol{\mu}_{j_l}\|$
In the second case 
the condition that moving the point does not pay off means:
$$\frac{n_{j_0}}{n_{j_0}-1} (x^2+y^2) \le \frac{n_{j_l}}{n_{j_l}+1}((d+x)^2+y^2) $$ 
If we multiply both sides with $\lambda^2$, we have:
\begin{align}
\lambda^2\frac{n_{j_0}}{n_{j_0}-1} (x^2+y^2) 
=& \frac{n_{j_0}}{n_{j_0}-1} ((\lambda x)^2+(\lambda y)^2) 
\nonumber \\
\le &  \CutEq{ \lambda^2 \frac{n_{j_l}}{n_{j_l}+1}((d+x)^2+y^2) 
\nonumber \\
=&
\frac{n_{j_l}}{n_{j_l}+1}( \lambda^2d^2+\lambda^2 2dx+\lambda^2x^2 +\lambda^2y^2) 
\nonumber \\
\le 
& \frac{n_{j_l}}{n_{j_l}+1}(  d^2+  2d\lambda x+\lambda^2x^2 +\lambda^2y^2) 
\nonumber \\
=& }
\frac{n_{j_l}}{n_{j_l}+1}(  (d +  \lambda x)^2+ (\lambda y)^2) 
\end{align} 
which means that it does not payoff to move the point $\textbf{x'}^*$ between clusters either.
Consider now the first case and assume that it pays off to move $\textbf{x'}^*$.
So we would have 
$$\frac{n_{j_0}}{n_{j_0}-1} (x^2+y^2) \le \frac{n_{j_l}}{n_{j_l}+1}((d-x)^2+y^2) $$ 
and at the same time
$$\frac{n_{j_0}}{n_{j_0}-1} \lambda^2(x^2+y^2) > \frac{n_{j_l}}{n_{j_l}+1}((d-\lambda x)^2+\lambda^2 y^2) $$ 
Subtract now both sides:
$$\frac{n_{j_0}}{n_{j_0}-1} (x^2+y^2) 
-\frac{n_{j_0}}{n_{j_0}-1} \lambda^2(x^2+y^2)$$
$$< \frac{n_{j_l}}{n_{j_l}+1}((d-x)^2+y^2) 
-\frac{n_{j_l}}{n_{j_l}+1}((d-\lambda x)^2+\lambda^2 y^2)$$
This  implies 
$$\frac{n_{j_0}}{n_{j_0}-1} (1-\lambda^2) (x^2+y^2) 
< 
\frac{n_{j_l}}{n_{j_l}+1}(
 (1-\lambda^2) (x^2+y^2)-2d\lambda x
)
$$ 
 It is a contradiction because 
$$\frac{n_{j_0}}{n_{j_0}-1} (1-\lambda^2) (x^2+y^2) 
>
\frac{n_{j_l}}{n_{j_l}+1} (1-\lambda^2) (x^2+y^2) 
 > 
\frac{n_{j_l}}{n_{j_l}+1}(
 (1-\lambda^2) (x^2+y^2)-2d\lambda x
)
$$
So it does not pay off to move $\textbf{x'}^*$, hence the partition $\Gamma'$ remains  locally optimal
\footnote{
$k$-means quality function is known to exhibit local minima at which the $k$-means algorithm may get stuck at. 
This claim means that after the centric $\Gamma$-transformation a partition will still be a local optimum. 
If the quality function has a unique local optimum then of course it is a global optimum and after the transform the partition yielding this global optimum will remain the global optimum. 
}
 for the transformed data set. 
\end{proof}
If the data have one stable optimum only like in case of
\textit{well separated} normally distributed 
   $k$ real clusters, then both turn to global optima. 

\begin{proof} of Theorem 
\ref{thm:globalCentricCinsistencyForKmeans} 
\label{thm:globalCentricCinsistencyForKmeansPROOF}. 
Let us consider first the simple case of two clusters only (2-means). 
Let the optimal clustering for a given set of objects $X$ consist of two clusters: $T$ and $Z$. 
The subset $T$ shall have its gravity center at the origin of the coordinate system. 
The quality of this partition 
$Q(\{T,Z\})=n_T Var(T) +n_Z Var(Z)$
where $n_T, n_Z$   denote the cardinalities of $T,Z$ and
$Var(T),Var(Z)$ their variances (averaged squared distances to gravity center). 
We will prove  by contradiction that by applying our $\Gamma$ transform we get partition that will be still optimal for the transformed data points.
We shall assume the contrary that is that we can 
transform the set $T$ by some $1>\lambda>0$ to $T'$ in such a way that optimum of $2$-means clustering is not the partition $\{T',Z\}$ but another one, say 
$\{A'\cup D,B'\cup C\}$ where $Z=C\cup D$, $A'$ and $B'$ are transforms of sets $A,B$ for which in turn $A\cup B=T$. 
It may be easily verified that
\newcommand{\w}{\mathbf{v}}  
$$Q(\{A\cup B, C\cup D\})= 
n_A Var(A) + n_A \w_A^2 
+ n_B Var(B) + n_B \w_B^2 
$$ $$
+ n_C Var(C) + n_D Var(D) +\frac{n_C n_D}{n_C+n_D} (\w_C-\w_D)^2$$  
while  
$$Q(\{A\cup C, B\cup D\})= 
n_A Var(A) + n_D Var(D)+ 
+\frac{n_A n_D}{n_A+n_D} (\w_A-\w_D)^2
$$ $$
+ n_B Var(B) + n_C Var(C)+ 
+\frac{n_B n_C}{n_B+n_C} (\w_B-\w_C)^2
$$  
and 
$$Q(\{A'\cup B', C\cup D\})= 
n_A \lambda^2Var(A) + n_A \lambda^2\w_A^2 
+ n_B \lambda^2Var(B) + n_B \lambda^2\w_B^2 
$$ $$
+ n_C Var(C) + n_D Var(D) +\frac{n_C n_D}{n_C+n_D} (\w_C-\w_D)^2$$  
while  
$$Q(\{A'\cup C, B'\cup D\})= 
n_A \lambda^2Var(A) + n_D Var(D)+ 
+\frac{n_A n_D}{n_A+n_D} (\lambda\w_A-\w_D)^2
$$ $$
+ n_B \lambda^2Var(B) + n_C Var(C)+ 
+\frac{n_B n_C}{n_B+n_C} (\lambda\w_B-\w_C)^2
$$  
The following must hold:
\begin{equation}
Q(\{A'\cup B', C\cup D\})>Q(\{A'\cup D, B'\cup C\})
\end{equation}
and 
\begin{equation}
Q(\{A\cup B, C\cup D\})<Q(\{A\cup D, B\cup C\})
\end{equation}
Additionally also 
\begin{equation}
Q(\{A\cup B, C\cup D\})<Q(\{A\cup B\cup C,  D  \})
\end{equation}
 and 
\begin{equation}
Q(\{A\cup B, C\cup D\})<Q(\{A\cup B\cup D,  C  \})
\end{equation}
These two latter inequalities imply:
\Bem{
$$Q(\{A\cup B, C\cup D\})= 
n_A Var(A) + n_A \w_A^2 
+ n_B Var(B) + n_B \w_B^2 
$$ $$
+ n_C Var(C) + n_D Var(D) +\frac{n_C n_D}{n_C+n_D} (\w_C-\w_D)^2$$  
$$<Q(\{A\cup B\cup C, D\})= 
n_A Var(A) + n_A \w_A^2 
+ n_B Var(B) + n_B \w_B^2 
$$ $$
+ n_C Var(C)+\frac{(n_A+n_B) n_C}{(n_A+n_B)+n_C} \w_C ^2 
+n_D Var(D) 
 $$ 
that is
}
$$ 
   \frac{n_C n_D}{n_C+n_D} (\w_C-\w_D)^2 < 
  \frac{(n_A+n_B) n_C}{(n_A+n_B)+n_C} \w_C ^2 
 $$   
and
$$ 
   \frac{n_C n_D}{n_C+n_D} (\w_C-\w_D)^2 < 
  \frac{(n_A+n_B) n_D}{(n_A+n_B)+n_D} \w_D ^2 
 $$  
Consider now an extreme contraction ($\lambda=0$) yielding sets $A",B"$ out of $A,B$.
Then we have 

$$Q(\{A"\cup B", C\cup D\})
- Q(\{A"\cup C, B"\cup D\})
$$ $$
= 
   \frac{n_C n_D}{n_C+n_D} (\w_C-\w_D)^2
-  
 \frac{n_A n_D}{n_A+n_D}  \w_D^2
-\frac{n_B n_C}{n_B+n_C} \w_C^2
$$ $$
= 
   \frac{n_C n_D}{n_C+n_D} (\w_C-\w_D)^2
$$ $$
-  
 \frac{n_A n_D}{n_A+n_D}
\frac{(n_A+n_B)+n_D}{(n_A+n_B) n_D}
\frac{(n_A+n_B) n_D}{(n_A+n_B)+n_D}
  \w_D^2
$$ $$
-\frac{n_B n_C}{n_B+n_C}
\frac{(n_A+n_B)+n_C}{(n_A+n_B) n_C}
\frac{(n_A+n_B) n_C}{(n_A+n_B)+n_C}
 \w_C^2
$$ $$
= 
   \frac{n_C n_D}{n_C+n_D} (\w_C-\w_D)^2
$$ $$
-  
 \frac{n_A  }{n_A+n_D}
\frac{(n_A+n_B)+n_D}{(n_A+n_B)  }
\frac{(n_A+n_B) n_D}{(n_A+n_B)+n_D}
  \w_D^2
$$ $$
-\frac{n_B  }{n_B+n_C}
\frac{(n_A+n_B)+n_C}{(n_A+n_B)  }
\frac{(n_A+n_B) n_C}{(n_A+n_B)+n_C}
 \w_C^2
$$ $$
= 
   \frac{n_C n_D}{n_C+n_D} (\w_C-\w_D)^2
$$ $$
-  
  \frac{n_A   }{ n_A+n_B   }
(1+\frac{n_B}{n_A+n_D})
\frac{(n_A+n_B) n_D}{(n_A+n_B)+n_D}
  \w_D^2
$$ $$
-\frac{n_B  }{n_A+n_B  }
(1+\frac{n_A}{n_B+n_C})
\frac{(n_A+n_B) n_C}{(n_A+n_B)+n_C}
 \w_C^2
$$ $$
< 
   \frac{n_C n_D}{n_C+n_D} (\w_C-\w_D)^2
$$ $$
-  
  \frac{n_A   }{ n_A+n_B   }
\frac{(n_A+n_B) n_D}{(n_A+n_B)+n_D}
  \w_D^2
$$ $$
-\frac{n_B  }{n_A+n_B  } 
\frac{(n_A+n_B) n_C}{(n_A+n_B)+n_C}
 \w_C^2
<0
$$
because the linear combination of two numbers that are bigger than a third yields another number bigger than this. 
Let us define a function 
$$h(x)=
  + n_A x^2\w_A^2 
  + n_B x^2\w_B^2 
    +\frac{n_C n_D}{n_C+n_D} (\w_C-\w_D)^2
$$ $$    
-\frac{n_A n_D}{n_A+n_D} (x\w_A-\w_D)^2  
-\frac{n_B n_C}{n_B+n_C} (x\w_B-\w_C)^2
$$   
It can be easily verified that $h(x)$ is a quadratic polynomial 
with a positive coefficient at $x^2$.
Furthermore
$h(1)=Q(\{A\cup B, C\cup D\})
- Q(\{A\cup C, B\cup D\})<0$, 
$h(\lambda)=Q(\{A'\cup B', C\cup D\})
- Q(\{A'\cup C, B'\cup D\})>0$, 
$h(0)=Q(\{A"\cup B", C\cup D\})
- Q(\{A"\cup C, B"\cup D\})<0$. 
But no quadratic polynomial with a positive coefficient at $x^2$ can be negative at the ends of an interval and positive in the middle.
So we have the contradiction. This proves the thesis
that the (globally) optimal $2$-means clustering remains 
(globally) optimal after transformation. 

Let us turn to the general case of $k$-means. 
Let the optimal clustering for a given set of objects $X$ consist of $k$ clusters: $T$ and $Z_1,\dots,Z_{k-1}$. 
The subset $T$ shall have its gravity center at the origin of the coordinate system. 
The quality of this partition 
$Q(\{T,Z_1,\dots,Z_{k-1}\}) =n_T Var(T) +\sum_{i=1}^{k-1}n_{Z_i} Var(Z_i)$,  
where $n_{Z_i} $ is the cardinality of the cluster ${Z_i} $.
We will prove  by contradiction that by applying our $\Gamma$ transform we get partition that will be still optimal for the transformed data points.
We shall assume the contrary that is that we can 
transform the set $T$ by some $1>\lambda>0$ to $T'$ in such a way that optimum of $k$-means clustering is not the partition $\{T',Z_1,\dots,Z_{k-1}\}$ but another one, say 
$\{T'_1\cup Z_{1,1} \cup \dots \cup Z_{k-1,1}
, T'_2\cup Z_{1,2} \cup \dots \cup Z_{{k-1},2}
\dots
, T'_k\cup Z_{1,k} \cup \dots \cup Z_{{k-1},k}
\}$ where $Z_i= \cup_{j=1}^{k} Z_{i,j}$ (where $Z_{i,j}$ are pairwise disjoint), 
$T'_1,\dots,T'_k$   are transforms of disjoint sets 
$T_1,\dots,T_k$ for which in turn $\cup_{j=1}^{k}T_j=T$. 
It may be easily verified that
$$Q(\{T,Z_1,\dots,Z_{k-1}\})= 
\sum_{j=1}^{k}n_{T_j}  Var({T_j}) + \sum_{j=1}^{k}n_{T_j} \w_{T_j}^2 
+ \sum_{i=1}^{k-1}n_{Z_i} Var({Z_i})$$  
while (denoting $Z_{*,j}= \cup_{i=1}{k-1}Z_{*,j}$) 
$$Q(\{T_1\cup  Z_{*,1},
\dots, T_k\cup  Z_{*,k}
  \})= $$ $$ = 
\sum_{j=1}^{k}
\left(
n_{T_j} Var({T_j}) + n_{Z_{*,j}}  Var(Z_{*,j})+ 
+\frac{n_{T_j} n_{Z_{*,j}}}{n_{T_j}+n_{Z_{*,j}}} (\w_{T_j}-\w_{Z_{*,j}})^2
\right)
$$ 
whereas
$$Q(\{T',Z_1,\dots,Z_{k-1}\})= 
\sum_{j=1}^{k}n_{T_j}  \lambda^2Var({T_j}) + \sum_{j=1}^{k}n_{T_j} \lambda^2\w_{T_j}^2 
$$ $$
+ \sum_{i=1}^{k-1}n_{Z_i} Var({Z_i})$$  
while
$$Q(\{T'_1\cup  Z_{*,1},
\dots, T'_k\cup  Z_{*,k}
  \})= $$ $$ = 
\sum_{j=1}^{k}
\left(
n_{T_j} \lambda^2Var({T_j}) + n_{Z_{*,j}}  Var(Z_{*,j})+ 
+\frac{n_{T_j} n_{Z_{*,j}}}{n_{T_j}+n_{Z_{*,j}}} (\lambda\w_{T_j}-\w_{Z_{*,j}})^2
\right)
$$ 
The following must hold:
\begin{equation}
Q(\{T',Z_1,\dots,Z_{k-1}\})>Q(\{
T'_1\cup  Z_{*,1},
\dots, T'_k\cup  Z_{*,k} \})
\end{equation} 
and 
\begin{equation}
Q(\{T,Z_1,\dots,Z_{k-1}\})<Q(\{
\{T_1\cup  Z_{*,1},
\dots, T_k\cup  Z_{*,k}\})
\end{equation}

Additionally also 
\begin{equation}
Q(\{T,Z_1,\dots,Z_{k-1}\})<Q(\{
\{T\cup  Z_{*,1},Z_{*,2},  \dots, Z_{*,k}
)\end{equation}
 and  
\begin{equation}
Q(\{T,Z_1,\dots,Z_{k-1}\})<Q(\{
T\cup  Z_{*,2},Z_{*,1},Z_{*,3},  \dots, Z_{*,k} \}
)\end{equation}
 and \dots and   
\begin{equation}
Q(\{T,Z_1,\dots,Z_{k-1}\})<Q(\{
T\cup  Z_{*,k},Z_{*,1},  \dots, Z_{*,k-1} \}
)\end{equation}

These latter $k$ inequalities imply that for $l=1,\dots,k$:
$$Q(\{T,Z_1,\dots,Z_{k-1}\})= 
 n_{T}  Var({T})+
\sum_{j=1}^{k}n_{T_j}  Var({T_j}) + \sum_{j=1}^{k}n_{T_j} \w_{T_j}^2 
$$ $$
+ \sum_{i=1}^{k-1}n_{Z_i} Var({Z_i})
< $$ $$
Q(\{T\cup  Z_{*,l},
Z_{*,1},\dots,Z_{*,l-1},Z_{*,l+1}
\dots,   Z_{*,k}
  \})= $$ $$ = 
 n_{T}  Var({T})+
\sum_{j=1}^{k}n_{Z_{*,j}}  Var(Z_{*,j})
+\frac{n_{T} n_{Z_{*,l}}}{n_{T}+n_{Z_{*,l}}} (\w_{T}-\w_{Z_{*,l}})^2
$$
  $$
+ \sum_{i=1}^{k-1}n_{Z_i} Var({Z_i})
< $$ $$
\sum_{j=1}^{k}n_{Z_{*,j}}  Var(Z_{*,j})
+\frac{n_{T} n_{Z_{*,l}}}{n_{T}+n_{Z_{*,l}}} (\w_{T}-\w_{Z_{*,l}})^2
$$
  $$
+  \sum_{i=1}^{k-1}n_{Z_i} Var({Z_i})
-\sum_{j=1}^{k}n_{Z_{*,j}}  Var(Z_{*,j})
< $$ $$
\frac{n_{T} n_{Z_{*,l}}}{n_{T}+n_{Z_{*,l}}} (\w_{Z_{*,l}})^2
$$

Consider now an extreme contraction ($\lambda=0$) yielding sets $T_j"$ out of $T_j$.
Then we have 

$$Q(\{T",Z_1,\dots,Z_{k-1}\})
-Q(\{T"_1\cup  Z_{*,1},
\dots, T"_k\cup  Z_{*,k}
  \})
$$ $$
= 
   \sum_{i=1}^{k-1}n_{Z_i} Var({Z_i}) 
-
\sum_{j=1}^{k}
\left(
  n_{Z_{*,j}}  Var(Z_{*,j}) 
+\frac{n_{T_j} n_{Z_{*,j}}}{n_{T_j}+n_{Z_{*,j}}} (\w_{Z_{*,j}})^2
\right)
$$ $$
= 
 \sum_{i=1}^{k-1}n_{Z_i} Var({Z_i})
-\sum_{j=1}^{k} n_{Z_{*,j}}  Var(Z_{*,j})
$$ $$
-\sum_{j=1}^{k}
\frac{n_{T_j} n_{Z_{*,j}}}{n_{T_j}+n_{Z_{*,j}}}
\frac{n_{T}+n_{Z_{*,j}}} {n_{T} n_{Z_{*,j}}}
\frac{n_{T} n_{Z_{*,j}}}{n_{T}+n_{Z_{*,j}}} 
 (\w_{Z_{*,j}})^2
$$ $$
= 
 \sum_{i=1}^{k-1}n_{Z_i} Var({Z_i})  
-\sum_{j=1}^{k} n_{Z_{*,j}}  Var(Z_{*,j})
$$ $$
-\sum_{j=1}^{k}
\frac{n_{T_j}  }{n_{T_j}+n_{Z_{*,j}}}
\frac{n_{T}+n_{Z_{*,j}}} {n_{T}  }
\frac{n_{T} n_{Z_{*,j}}}{n_{T}+n_{Z_{*,j}}} 
 (\w_{Z_{*,j}})^2
$$ $$
\le  
 \sum_{i=1}^{k-1}n_{Z_i} Var({Z_i})  
-\sum_{j=1}^{k} n_{Z_{*,j}}  Var(Z_{*,j})
-\sum_{j=1}^{k}
\frac{n_{T_j}  } {n_{T}  }
\frac{n_{T} n_{Z_{*,j}}}{n_{T}+n_{Z_{*,j}}} 
 (\w_{Z_{*,j}})^2
<0
$$
because the linear combination of  numbers that are bigger than a third yields another number bigger than this. 
Let us define a function 
$$g(x)=
 \sum_{j=1}^{k}n_{T_j} x^2\w_{T_j}^2 
+ \sum_{i=1}^{k-1}n_{Z_i} Var({Z_i})  
$$ $$
-
\sum_{j=1}^{k}
\left(
  n_{Z_{*,j}}  Var(Z_{*,j})+ 
+\frac{n_{T_j} n_{Z_{*,j}}}{n_{T_j}+n_{Z_{*,j}}} (x\w_{T_j}-\w_{Z_{*,j}})^2
\right)
$$ 
It can be easily verified that $g(x)$ is a quadratic polynomial 
with a positive coefficient at $x^2$.
Furthermore
$g(1)=Q(\{T,Z_1,\dots,z_{k-1}\})
- Q(\{T_1\cup  Z_{*,1},\dots, T_k\cup  Z_{*,k} \})<0$, 
$g(\lambda)=Q(\{T',Z_1,\dots,Z_{k-1}\})
- Q(\{T'_1\cup  Z_{*,1},\dots, T'_k\cup  Z_{*,k} \})>0$, 
$g(0)=Q(\{T",Z_1,\dots,Z_{k-1}\})
- Q(\{T"_1\cup  Z_{*,1},\dots, T"_k\cup  Z_{*,k} \})<0$. 
But no quadratic polynomial with a positive coefficient at $x^2$ can be negative at the ends of an interval and positive in the middle.
So we have the contradiction. This proves the thesis
that the (globally) optimal $k$-means clustering remains 
(globally) optimal after transformation. 
\end{proof}

So summarizing the new $\Gamma$ transformation preserves local and global optima of $k$-means for a fixed $k$.
Therefore $k$-means algorithm is consistent under this transformation.

Note that ($\Gamma^*$ based) centric-consistency 
is not a specialization of Kleinberg's consistency 
as the requirement of increased distance between all elements of different clusters is not required in
$\Gamma^*$ based  Consistency. 

\Bem{
Note also that the decrease of distance does not need to be 
equal for all elements as long as the gravity center does not relocate. 
Also a limited rotation of the cluster may be allowed for. 
But we could strengthen centric-consistency to be in concordance with Kleinberg['s consistency and under this strengthening $k$-means would of course still behave properly. 
}

\Bem{
\begin{theorem}{} \label{thm:globalCentricConsistencyFor2meansSubsetted} 
Let a partition $\{T,Z\}$ be an optimal partition under 
$2$-means algorithm.
Let a subset $P$ of $T$ be subjected to centric $\Gamma$-transformation 
yielding $P'$, and $T'=(T-P)\cup P'$. Then  
partition $\{T',Z\}$ is an optimal partition of $T'\cup Z$ under $2$-means. 
    satisfies
\end{theorem}
}

\Bem{
The above theorem means that
\begin{theorem} \label{thm:autocentricinvrich}
 Bisectional-$auto$-means algorithm is   centric-consistent,  (scale)-invariant and $2++$-nearly-rich.  
\end{theorem}
}


\section{Bi{}sectional-Auto-$X$-Means}

\begin{proof}
of Theorem \ref{thm:globalCentricConsistencyFor2meansSubsetted}\label{thm:globalCentricConsistencyFor2meansSubsettedPROOF} (Outline) 
Let the optimal clustering for a given set of objects $\mathbf{X}$ consist of two clusters: $T$ and $Z$. 
Let $T$ consist of two disjoint subsets $P$, $Y$,
$T=P \cup Y$ and let us ask the question whether or not centric transformation of the set $P$ will affect the optimality of clustering. 
Let   $T'(\lambda)=P'(\lambda) \cup Y$ with $P'(\lambda)$ being an image of $P$ under centric transformation. 
The cluster centre of $T'(\lambda)$ will be the same as that of $T$. 
 We ask if $\{T'(\lambda),Z\}$ is the globally optimal clustering of $T'(\lambda)\cup Z$.
Assume the contrary, that is that there exists 
a clustering into 
sets $K'(\lambda)=A'(\lambda)\cup B\cup C$,  $L'(\lambda)=D'(\lambda)\cup E\cup F$,
where $P=A\cup D$, and 
$A'(\lambda)$ are the points obtains from $A$ when $P$ is subjected to centric transformation, and $D'(\lambda)$ is defined analogously, hence 
$P'(\lambda)=A'(\lambda)\cup D'(\lambda), Y=B\cup E, Z=C\cup F$,
 that, for some $\lambda=\lambda^*\in (0,1)$  has lower clustering quality 
function value $Q(\{K'(\lambda),L'(\lambda)\} )$.
Define also the function 
$h(\lambda)=Q(\{T'(\lambda),Z\})-Q(\{K'(\lambda),L'(\lambda)\})$.
Due to optimality assumption, $h(1)\le 0$.  
\Bem{
1/2 |C| sum_x \in C sum_Y \in C  (x-y)^2
= in case of two 
1/2 (n1+n2)( n1*n2(mu1-mu2)^2+ n2*n1(mu2-mu1)^2
= in case of three 
1/2 (n1+n2+n3)( n1*n2(mu1-mu2)^2+ n1*n3(mu1-mu3)^2+
n2*n1(mu2-mu1)^2
n2*n3(mu2-mu3)^2
n3*n1(mu3-mu1)^2
n3*n2(mu3-mu2)^2

)

}

Let us discuss now the centric transform with   $\lambda=0$.
In this case all points from $A'(0)$ and $D'(0)$ collapse to a single point. This point can be closer to either $\boldsymbol\mu(K'(0))$ or  $\boldsymbol\mu(L'(0))$. Assume they are closer to $\boldsymbol\mu(K'(0))$. 
In this case 
$Q(\{K"(\lambda),L"(\lambda)\} ) \le
Q(\{K'(\lambda),L'(\lambda)\} )$.
where 
$K"(\lambda)=P'(\lambda)\cup B\cup C$,  $L"(\lambda)= E\cup F$ for $\lambda=0$,
As all points subject to centric consistency are contained in a single set, we get
$$Q(\{T'(0),Z\})-Q(\{K"(0),L"(0)\} )
=Q(\{T'(1),Z\})-Q(\{K"(1),L"(1)\} )
\le 0$$
because $Q(\{T'(1),Z\})=Q(\{T,Z\})$ is the optimum. 
Hence also 
$$Q(\{T'(0),Z\})-Q(\{K'(0),L'(0)\} )
\le 0$$ that is $h(0)\le 0$. 
  
It is also easily seen that $h(\lambda)$ is a quadratic function of $\lambda$. 
This can be seen as follows:
\begin{align*}
& Q(\{T'(\lambda),Z\})=
\left(\sum_{\mathbf{x}\in T'(\lambda)} \|\mathbf{x}-\boldsymbol\mu( T'(\lambda))\|^2\right)
+
\left(\sum_{\mathbf{x}\in Z} \|\mathbf{x}-\boldsymbol\mu( Z) \|^2\right)
\\=&
\left(\sum_{\mathbf{x}\in P'(\lambda)} \|\mathbf{x}-\boldsymbol\mu( P'(\lambda))\|^2\right)
+
\left(\sum_{\mathbf{x}\in Y} \|\mathbf{x}-\boldsymbol\mu( Y)\|^2\right)
\\&+\|\boldsymbol\mu(P'(\lambda))-\boldsymbol\mu(Y)\|^2\cdot \frac{1}{1/|P|+1/|Y|}
+ 
\left(\sum_{\mathbf{x}\in Z} \|\mathbf{x}-\boldsymbol\mu( Z) \|^2\right)
\\=&
\left(\sum_{\mathbf{x}\in A'(\lambda)} \|\mathbf{x}-\boldsymbol\mu( A'(\lambda))\|^2\right)
+
\left(\sum_{\mathbf{x}\in D'(\lambda)} \|\mathbf{x}-\boldsymbol\mu( D'(\lambda))\|^2\right)
\\&+|A|\|\boldsymbol\mu( A'(\lambda))-\boldsymbol\mu( P'(\lambda))\|^2
+|D|\|\boldsymbol\mu( D'(\lambda))-\boldsymbol\mu( P'(\lambda))\|^2
\\&+
\left(\sum_{\mathbf{x}\in Y} \|\mathbf{x}-\boldsymbol\mu( Y)\|^2\right)
\\&+\|\boldsymbol\mu(P'(\lambda))-\boldsymbol\mu(Y)\|^2\cdot \frac{1}{1/|P|+1/|Y|}
+
\left(\sum_{\mathbf{x}\in Z} \|\mathbf{x}-\boldsymbol\mu( Z) \|^2\right) 
\end{align*}

This expression is obviously quadratic in $\lambda$, since each point $\mathbf{x}\in P$  is transformed linearly to $\boldsymbol\mu(P)+\lambda(\mathbf{x}-\boldsymbol\mu(P))$. 
Note that $\boldsymbol\mu( P'(\lambda))=\boldsymbol\mu( P  )$
so it does not depend on $\lambda$. 
On the other hand 
\begin{align*}
&Q(\{K'(\lambda),L'(\lambda)\})=Q(\{A'(\lambda)\cup B \cup C,D'(\lambda)\cup E \cup F\})
\\&=
\left(\sum_{\mathbf{x}\in A'(\lambda)} \|\mathbf{x}-\boldsymbol\mu( A'(\lambda))\|^2\right)
+\left(\sum_{\mathbf{x}\in B} \|\mathbf{x}-\boldsymbol\mu( B)\|^2\right)
+\left(\sum_{\mathbf{x}\in C} \|\mathbf{x}-\boldsymbol\mu( C)\|^2\right)
\\&+\frac{1}{|A|+|B|+|C|}
\left(
|A||B| \|\boldsymbol\mu(A'(\lambda))-\boldsymbol\mu(B)\|^2
\right.\\&\left.
+|A||C| \|\boldsymbol\mu(A'(\lambda))-\boldsymbol\mu(C)\|^2
+|B||C| \|\boldsymbol\mu(B)-\boldsymbol\mu(C)\|^2
\right)
\\&+
\left(\sum_{\mathbf{x}\in D'(\lambda)} \|\mathbf{x}-\boldsymbol\mu( D'(\lambda))\|^2\right)
+\left(\sum_{\mathbf{x}\in E} \|\mathbf{x}-\boldsymbol\mu( E)\|^2\right)
+\left(\sum_{\mathbf{x}\in F} \|\mathbf{x}-\boldsymbol\mu( F)\|^2\right)
\\&+\frac{1}{|D|+|E|+|F|}
\left(
|D||E| \|\boldsymbol\mu(D'(\lambda))-\boldsymbol\mu(E)\|^2
\right.\\&\left.
+|D||F| \|\boldsymbol\mu(D'(\lambda))-\boldsymbol\mu(F)\|^2
+|F||E| \|\boldsymbol\mu(F)-\boldsymbol\mu(E)\|^2
\right)
\end{align*}

Then 

\begin{align*}
&h(\lambda)=Q(\{T'(\lambda),Z\})-Q(\{K'(\lambda),L'(\lambda)\})
\\=&
\left(\sum_{\mathbf{x}\in A'(\lambda)} \|\mathbf{x}-\boldsymbol\mu( A'(\lambda))\|^2\right)
+
\left(\sum_{\mathbf{x}\in D'(\lambda)} \|\mathbf{x}-\boldsymbol\mu( D'(\lambda))\|^2\right)
\\&+|A|\|\boldsymbol\mu( A'(\lambda))-\boldsymbol\mu( P )\|^2
+|D|\|\boldsymbol\mu( D'(\lambda))-\boldsymbol\mu( P )\|^2
\\&+
\left(\sum_{\mathbf{x}\in Y} \|\mathbf{x}-\boldsymbol\mu( Y)\|^2\right)
\\&+\|\boldsymbol\mu(P)-\boldsymbol\mu(Y)\|^2\cdot \frac{1}{1/|P|+1/|Y|}
+
\left(\sum_{\mathbf{x}\in Z} \|\mathbf{x}-\boldsymbol\mu( Z) \|^2\right)
\\&-
\left(\sum_{\mathbf{x}\in A'(\lambda)} \|\mathbf{x}-\boldsymbol\mu( A'(\lambda))\|^2\right)
-\left(\sum_{\mathbf{x}\in B} \|\mathbf{x}-\boldsymbol\mu( B)\|^2\right)
-\left(\sum_{\mathbf{x}\in C} \|\mathbf{x}-\boldsymbol\mu( C)\|^2\right)
\\&-\frac{1}{|A|+|B|+|C|}
\left(
|A||B| \|\boldsymbol\mu(A'(\lambda))-\boldsymbol\mu(B)\|^2
\right.\\&\left.
+|A||C| \|\boldsymbol\mu(A'(\lambda))-\boldsymbol\mu(C)\|^2
+|B||C| \|\boldsymbol\mu(B)-\boldsymbol\mu(C)\|^2
\right)
\\&-
\left(\sum_{\mathbf{x}\in D'(\lambda)} \|\mathbf{x}-\boldsymbol\mu( D'(\lambda))\|^2\right)
-\left(\sum_{\mathbf{x}\in E} \|\mathbf{x}-\boldsymbol\mu( E)\|^2\right)
-\left(\sum_{\mathbf{x}\in F} \|\mathbf{x}-\boldsymbol\mu( F)\|^2\right)
\\&-\frac{1}{|D|+|E|+|F|}
\left(
|D||E| \|\boldsymbol\mu(D'(\lambda))-\boldsymbol\mu(E)\|^2
\right.\\&\left.
+|D||F| \|\boldsymbol\mu(D'(\lambda))-\boldsymbol\mu(F)\|^2
+|F||E| \|\boldsymbol\mu(F)-\boldsymbol\mu(E)\|^2
\right)
\end{align*}
\begin{align*}
\\=&
 |A|\|\boldsymbol\mu( A'(\lambda))-\boldsymbol\mu( P )\|^2
+|D|\|\boldsymbol\mu( D'(\lambda))-\boldsymbol\mu( P )\|^2
\\&-\frac{1}{|A|+|B|+|C|}
\left(
|A||B| \|\boldsymbol\mu(A'(\lambda))-\boldsymbol\mu(B)\|^2
+|A||C| \|\boldsymbol\mu(A'(\lambda))-\boldsymbol\mu(C)\|^2\right)
\\&-\frac{1}{|D|+|E|+|F|}
\left(
|D||E| \|\boldsymbol\mu(D'(\lambda))-\boldsymbol\mu(E)\|^2
+|D||F| \|\boldsymbol\mu(D'(\lambda))-\boldsymbol\mu(F)\|^2\right)
\\&+
\left(\sum_{\mathbf{x}\in Y} \|\mathbf{x}-\boldsymbol\mu( Y)\|^2\right)
-\left(\sum_{\mathbf{x}\in B} \|\mathbf{x}-\boldsymbol\mu( B)\|^2\right)
-\left(\sum_{\mathbf{x}\in E} \|\mathbf{x}-\boldsymbol\mu( E)\|^2\right)
\\&+\|\boldsymbol\mu(P )-\boldsymbol\mu(Y)\|^2\cdot \frac{1}{1/|P|+1/|Y|}
\\&+
\left(\sum_{\mathbf{x}\in Z} \|\mathbf{x}-\boldsymbol\mu( Z) \|^2\right)
-\left(\sum_{\mathbf{x}\in C} \|\mathbf{x}-\boldsymbol\mu( C)\|^2\right)
-\left(\sum_{\mathbf{x}\in F} \|\mathbf{x}-\boldsymbol\mu( F)\|^2\right)
\\&-\frac{1}{|A|+|B|+|C|}
\left(|B||C| \|\boldsymbol\mu(B)-\boldsymbol\mu(C)\|^2
\right)
\\&-\frac{1}{|D|+|E|+|F|}
\left(
|F||E| \|\boldsymbol\mu(F)-\boldsymbol\mu(E)\|^2
\right)
\end{align*}

\begin{align*}
\\=&
 |A|\|\boldsymbol\mu( A'(\lambda))-\boldsymbol\mu( P )\|^2
+|D|\|\boldsymbol\mu( D'(\lambda))-\boldsymbol\mu( P )\|^2
\\&-\frac{1}{|A|+|B|+|C|}
\left(
|A||B| \|\boldsymbol\mu(A'(\lambda))-\boldsymbol\mu(B)\|^2
+|A||C| \|\boldsymbol\mu(A'(\lambda))-\boldsymbol\mu(C)\|^2\right)
\\&-\frac{1}{|D|+|E|+|F|}
\left(
|D||E| \|\boldsymbol\mu(D'(\lambda))-\boldsymbol\mu(E)\|^2
+|D||F| \|\boldsymbol\mu(D'(\lambda))-\boldsymbol\mu(F)\|^2\right)
\\&+c_h
\end{align*}

\noindent 
\noindent 
 where $c_h$  is a constant (independent of $\lambda$ and all the $\boldsymbol\mu$s that depend on $\lambda$, are linearly dependent on it (by definition of centric consistency). 
 Recall that  
 $\boldsymbol\mu( A'(\lambda))-\boldsymbol\mu( P )
 =1/|A|\sum_{\mathbf{x}\in A}
 \lambda(\mathbf{x}-\boldsymbol\mu( P )) 
 =
 \lambda \mathbf{v_A}
 $,
 where $ \mathbf{v_A}$ is a vector independent of $\lambda$. 
 Hence 
 $\|\boldsymbol\mu( A'(\lambda))-\boldsymbol\mu( P )\|^2=\lambda^2  \mathbf{v_A}^T \mathbf{v_A}
 $. 
 Similarly 
 \begin{align*}
 \|\boldsymbol\mu&(A'(\lambda))-\boldsymbol\mu(B)\|^2
= 
 \|(\boldsymbol\mu(A'(\lambda)
 -\boldsymbol\mu( P ))
 +(\boldsymbol\mu( P )
 -\boldsymbol\mu(B))\|^2
 \\ & = 
 (\boldsymbol\mu(A'(\lambda)
 -\boldsymbol\mu( P ))^2
 +(\boldsymbol\mu( P )
 -\boldsymbol\mu(B))^2+
 2(\boldsymbol\mu(A'(\lambda)
 -\boldsymbol\mu( P ))(\boldsymbol\mu( P )
 -\boldsymbol\mu(B))
 \\ & = 
 \lambda^2  \mathbf{v_A}^T \mathbf{v_A}
 +(\boldsymbol\mu( P )
 -\boldsymbol\mu(B))^2+
 2(\boldsymbol\mu(A'(\lambda)
 -\boldsymbol\mu( P ))(\boldsymbol\mu( P )
 -\boldsymbol\mu(B))
 \\ & = \lambda^2  \mathbf{v_A}^T \mathbf{v_A}
 + \lambda c_{ABP}
 +c_{BP}
 \end{align*}
 with $ c_{ABP},
 c_{BP}$ being constants independent of $\lambda$, 
 whereby only the first summand depends on $\lambda^2$. 
 Similarly 
$$ \|\boldsymbol\mu(A'(\lambda))-\boldsymbol\mu(C)\|^2
= \lambda^2  \mathbf{v_A}^T \mathbf{v_A}
 + \lambda c_{ACP}
 +c_{CP}
 $$
 with $ c_{ACP},
 c_{CP}$ being constants independent of $\lambda$, 
 and so on. 
 Therefore we
 can rewrite 
 the 
 $h(\lambda)$ as 
 \begin{align*}
  h(\lambda)&=   
 |A|
 \lambda^2  \mathbf{v_A}^T \mathbf{v_A}
+ |D|
 \lambda^2  \mathbf{v_D}^T \mathbf{v_D}
 \\ &  -\frac{1}{|A|+|B|+|C|}
\left(
|A||B| (
\lambda^2  \mathbf{v_A}^T \mathbf{v_A}
 + \lambda c_{ABP}
 +c_{BP}) \right.
 \\ & \left.
+|A||C| (\lambda^2  \mathbf{v_A}^T \mathbf{v_A}
 + \lambda c_{ACP}
 +c_{CP})
 \right)
 \\ & 
 -\frac{1}{|D|+|E|+|F|}
\left(
|D||E| (
\lambda^2  \mathbf{v_D}^T \mathbf{v_D}
 + \lambda c_{DEP}
 +c_{EP})\right.
\\ & \left.
+|D||F| (\lambda^2  \mathbf{v_D}^T \mathbf{v_D}
 + \lambda c_{DFP}
 +c_{FP})
 \right) +c_h
\end{align*}
  
 So the coefficient at $\lambda^2$ amounts to: 
 $$
 \left(|A|-
 -\frac{|A|(|B|+|C|)}{|A|+|B|+|C|}\right)  \mathbf{v_A}^T \mathbf{v_A}
 + \left(|D|-\frac{|D|(|E|+|F|)}{|D|+|E|+|F|}\right)
   \mathbf{v_D}^T \mathbf{v_D}
 $$
 which is bigger than 0 if only $|A|>0$ and $|D|>0$.
 which is te case by our assumption of an alternative clustering. 
 Therefore, for $\lambda$ large enough, $h(\lambda)>0$.

   As $h(\lambda)$ is a quadratic function in $\lambda$, and   $h(0)\le 0$ and $h(1)\le 0$, then also $h(\lambda)\le 0$ for any value of $\lambda$ between 0 and 1.  
This completes the proof.

 \end{proof}

\begin{proof}
Proof of Theorem \ref{thm:autocentricinvrich}
\label{thm:autocentricinvrichPROOF}.
The scale-invariance is implied by the properties of the $k$-means algorithm that serves as the subroutine and the fact that the stopping criterion is the relative decrease of $Q$, so that the stopping criterion is also not affected.

The  $2\uparrow$-nearly-richness can be achieved as follows. Each partition consists of clusters of predefined cardinalities. Within each partition distribute the data points uniformly on a line segment. If we set the stopping criterion as lower than 9-fold decrease of $Q$, then by proper manipulation of distances between groups of clusters when combining them to the cluster hierarchy will ensure that the targeted partition is restored.   

The centric consistency is implied by Theorem \ref{thm:globalCentricConsistencyFor2meansSubsetted} as follows.
At a given stage of the algorithm, when a bisection is to be performed, 
Theorem \ref{thm:globalCentricConsistencyFor2meansSubsetted} ensures that the optimal bisection is the one that was used in the original partitioning process. 
the question is only about the stopping criterion: whether  or not centric consistency would imply continuing the partitioning. 
Consider the notation from the previous proof. 
The $Q$ improvement in the original clustering would amount to:

\begin{align*}
&\frac{1}{ 
\left(\sum_{\mathbf{x}\in X} \|\mathbf{x}-\boldsymbol\mu( X) \|^2\right)
}    
\left(
\left(\sum_{\mathbf{x}\in P} \|\mathbf{x}-\boldsymbol\mu( P)\|^2\right)
+
\left(\sum_{\mathbf{x}\in Y} \|\mathbf{x}-\boldsymbol\mu( Y)\|^2\right)
 \right.\\&\left.
+\|\boldsymbol\mu(P)-\boldsymbol\mu(Y)\|^2\cdot \frac{1}{1/|P|+1/|Y|}
+ 
\left(\sum_{\mathbf{x}\in Z} \|\mathbf{x}-\boldsymbol\mu( Z) \|^2\right)
\right)
\end{align*}
and afterwards 
\begin{align*}
&\frac{1}{ 
\left(\sum_{\mathbf{x}\in X} \|\mathbf{x}-\boldsymbol\mu( X) \|^2\right)
-\left(\sum_{\mathbf{x}\in P} \|\mathbf{x}-\boldsymbol\mu( P)\|^2\right)
+\left(\sum_{\mathbf{x}\in P'(\lambda)} \|\mathbf{x}-\boldsymbol\mu( P)\|^2\right)
}    
\\& \cdot \left(
\left(\sum_{\mathbf{x}\in P'(\lambda)} \|\mathbf{x}-\boldsymbol\mu( P)\|^2\right)
+
\left(\sum_{\mathbf{x}\in Y} \|\mathbf{x}-\boldsymbol\mu( Y)\|^2\right)
 +  \frac{\|\boldsymbol\mu(P)-\boldsymbol\mu(Y)\|^2}{1/|P|+1/|Y|}
\right.\\&\left.
+ 
\left(\sum_{\mathbf{x}\in Z} \|\mathbf{x}-\boldsymbol\mu( Z) \|^2\right)
\right)
\end{align*}
This expression is as if 
$\left(\sum_{\mathbf{x}\in P} \|\mathbf{x}-\boldsymbol\mu( P)\|^2\right)
-\left(\sum_{\mathbf{x}\in P'(\lambda)} \|\mathbf{x}-\boldsymbol\mu( P)\|^2\right)$
(positive and smaller than the denominator)
were subtracted from the nominator and the denominator of the first quotient. That is the second quotient is smaller, so if the first did not stop the partitioning, so the second would not either. 
\end{proof}

Klein{}berg's three axioms/properties of consistency, scale-in{}variance and richness are contradictory. 
We obtained here three similar axioms that are not contradictory.  
If we would replace Klein{}berg's richness axiom only with $2\uparrow$-nearly-richness, this would not help to resolve the contradiction. 
The real driving force behind the conflict resolution is the centric consistency. 
Centric consistency may appear as more rigid than Klein{}berg's consistency, but on the other hand it is broader than Klein{}berg's, as already mentioned (not all distances between elements of distinct clusters need to increase).   
\Bem{
For example, in Figure \ref{fig:12internalClusters} continuous $\Gamma$-transformation is impossible for a single cluster, but continuous centric $\Gamma$-transformation is
possible. In Section \ref{sec:convergentConsistency} we show that centric consistency is less restrictive  than Klein{}berg's contradiction-free consistency variant. 
}

\Bem{
Obviously, 
convergent-$\Gamma$-transform 
is equivalent to applying 
multiple centric $\Gamma$-transformations and scale in{}variance transformation. This implies:
%
\begin{theorem}
Richness, invariance and convergent consistency are not contradictory properties 
for a clustering function. 
\end{theorem}
\begin{proof}
The Theorem is easily shown as 
according to the Theorem \ref{thm:autocentricinvrich},
the $auto$-means algorithm  preserves the clustering 
under centric consistency operator,
in{}variance operator and by itself it is rich.   
\end{proof}
}

So at this point we know that a reasonable version of Klein{}berg's $\Gamma$-transform 
is no more general than centric $\Gamma$-transform.
Let us briefly demonstrate that centric consistency 
raises claims of equivalent clustering also in those cases when Klein{}berg's consistency does not. 

Imagine points $A[-11], B[-9], C[-2],D[2], E[9], F[11]$ in 1d, constituting three clusters $\{A,B\},\{C,D\}, \{E,F\}$. 
Imagine centric-$\Gamma$-transform of the cluster $\{C,D\}$ into
$\{ C'[-1],D'[1]\}$.  
Note that hereby the quotient $\frac{|CB|}{|AB|}$ increases while  $\frac{|CB|}{|AB|}$
 $\frac{|CE|}{|EF|}$ decreases, 
and at the same time  $\frac{|DB|}{|AB|}$ decreases while  $\frac{|CB|}{|AB|}$
 $\frac{|DE|}{|EF|}$ increases.
No sequence of Kleinberg's $\Gamma$-transforms and invariance transforms  on these data 
can achieve such a transformation 
because all the mentioned quotients are non-decreasing upon consistency and invariance transforms. 
This implies that 
convergent-$\Gamma$-transform (the reasonable version of Kleinberg's $\Gamma$-transform) 
is a special case of  
centric-$\Gamma$-transform.

\section{Motion Consistency Proofs}\label{sec:motionconsistency}

\begin{proof}
of Theorem \ref{thm:GeneralMotionConsistency}
\label{thm:GeneralMotionConsistencyPROOF}. 
Given an optimal solution $\Gamma_O$ to $k$-means problem, where each cluster is enclosed in a (hyper)ball with radius $R$ centered at its gravity center, if we rotate the clusters around their gravity centers, then the position of any data point can change by at most $2R$. That is, the distances between points from different clusters can decrease by at most $4R$. So if we move away the clusters by the distance of $4R$, then (1) $\Gamma$ clustering gives the same value of $k$-means cost function as $\Gamma_O$ (as the distances within clusters do not change) and (2) upon rotation in the new position no clustering different from $\Gamma$ can have a lower cost function value than any competing clustering in the initial position of data points. Hence the optimum clustering is preserved upon any rotating  transformation, and when moving the clusters keeping the center distances as prescribed, the optimal clustering is preserved also. . 
\end{proof}

\begin{proof}
of Theorem \ref{thm:concavemotion}\label{thm:concavemotionPROOF}.
The assumption of the Theorem \ref{thm:concavemotion} implies that when applying the $k$-means-$l$-MST algorithm, the $k$-means cluster centers from different $l$-MST clusters will lie at distance of at least $m_d$. Therefore we can increase $k$ to such an extent that the distances between $k$-means clusters in different $l$-MST clusters are smaller than $m_d$. This means that the clustering will be preserved upon centric $\Gamma$-transform applied to $k$-means clusters. 
Let $R_M$ be the maximum radius of a hyper{}ball enclosing any $k$-means cluster. Let $d_M$ be the maximum distance between $k$-means clusters from different $l$-MST clusters. 
So apply centric $\Gamma$-transform to $k$-means clusters with 
$\lambda<\frac{m_d}{4R_M+d_M}$. 
If we move now the clusters, the motion consistency will be preserved, as implied by 
Theorem \ref{thm:GeneralMotionConsistency}.

\end{proof}

\begin{figure}
\begin{center}
(a)
\includegraphics[width=0.4\textwidth]{\figaddr{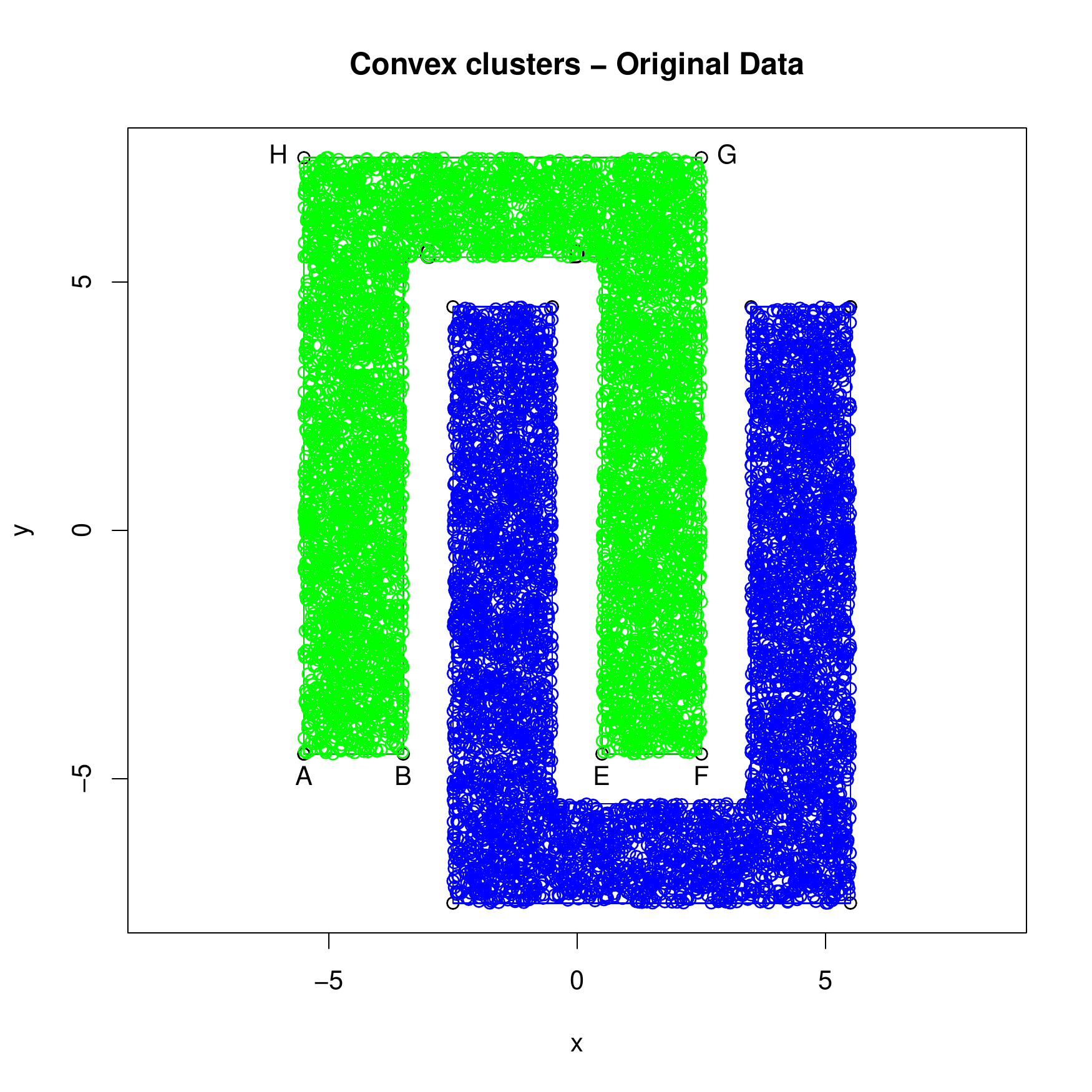}}  %
(b)
\includegraphics[width=0.4\textwidth]{\figaddr{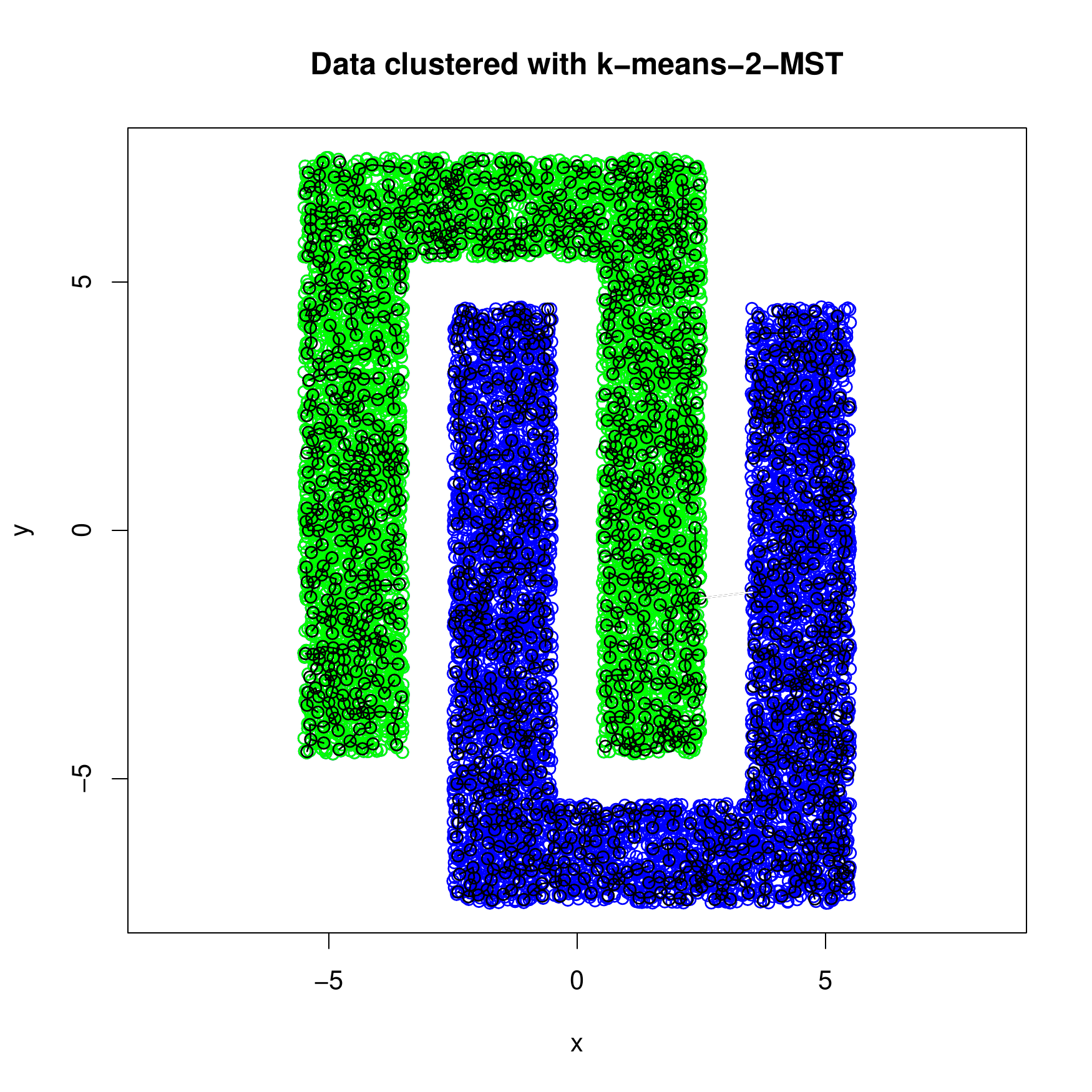}}\\  %
(c)
\includegraphics[width=0.4\textwidth]{\figaddr{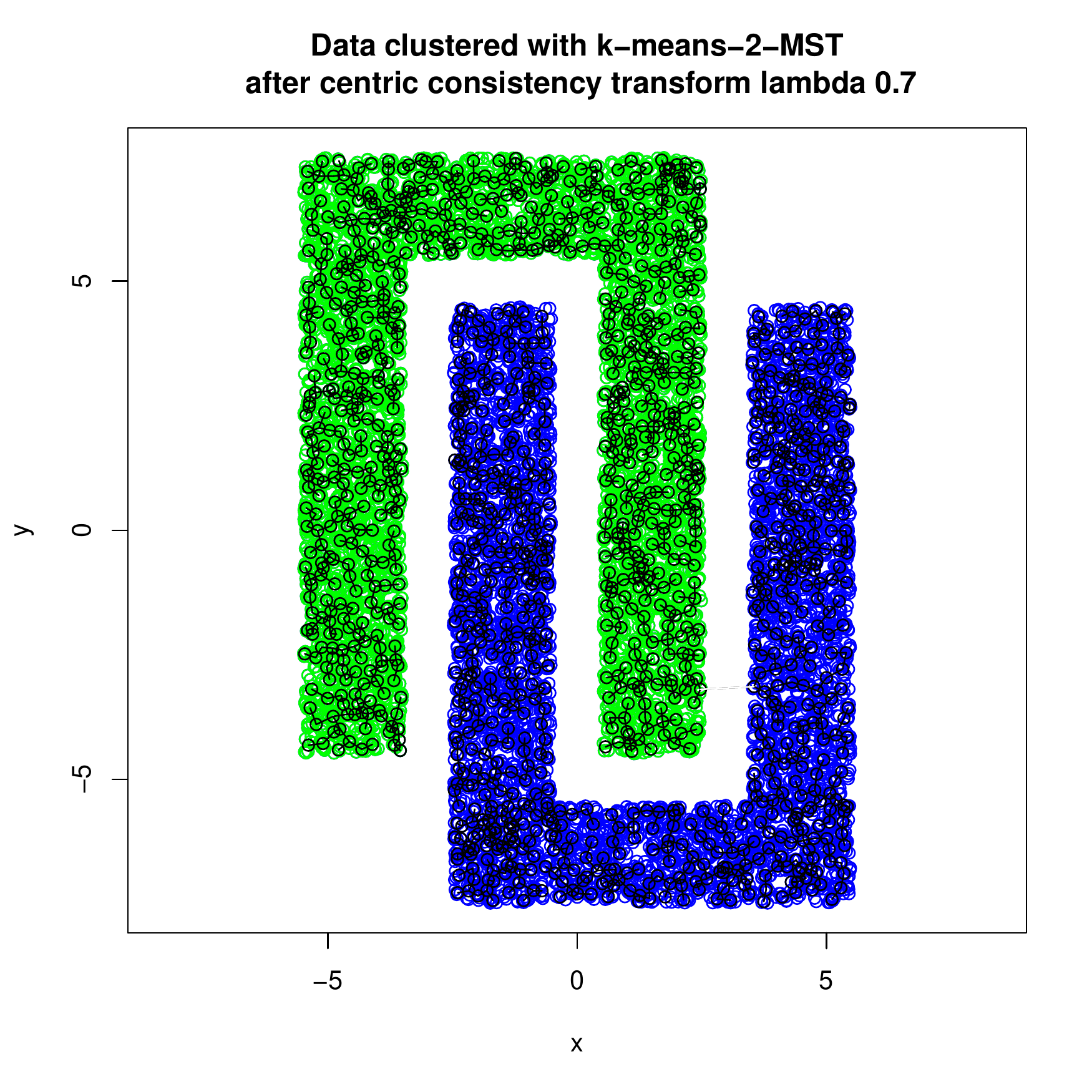}}  %
(d)
\includegraphics[width=0.4\textwidth]{\figaddr{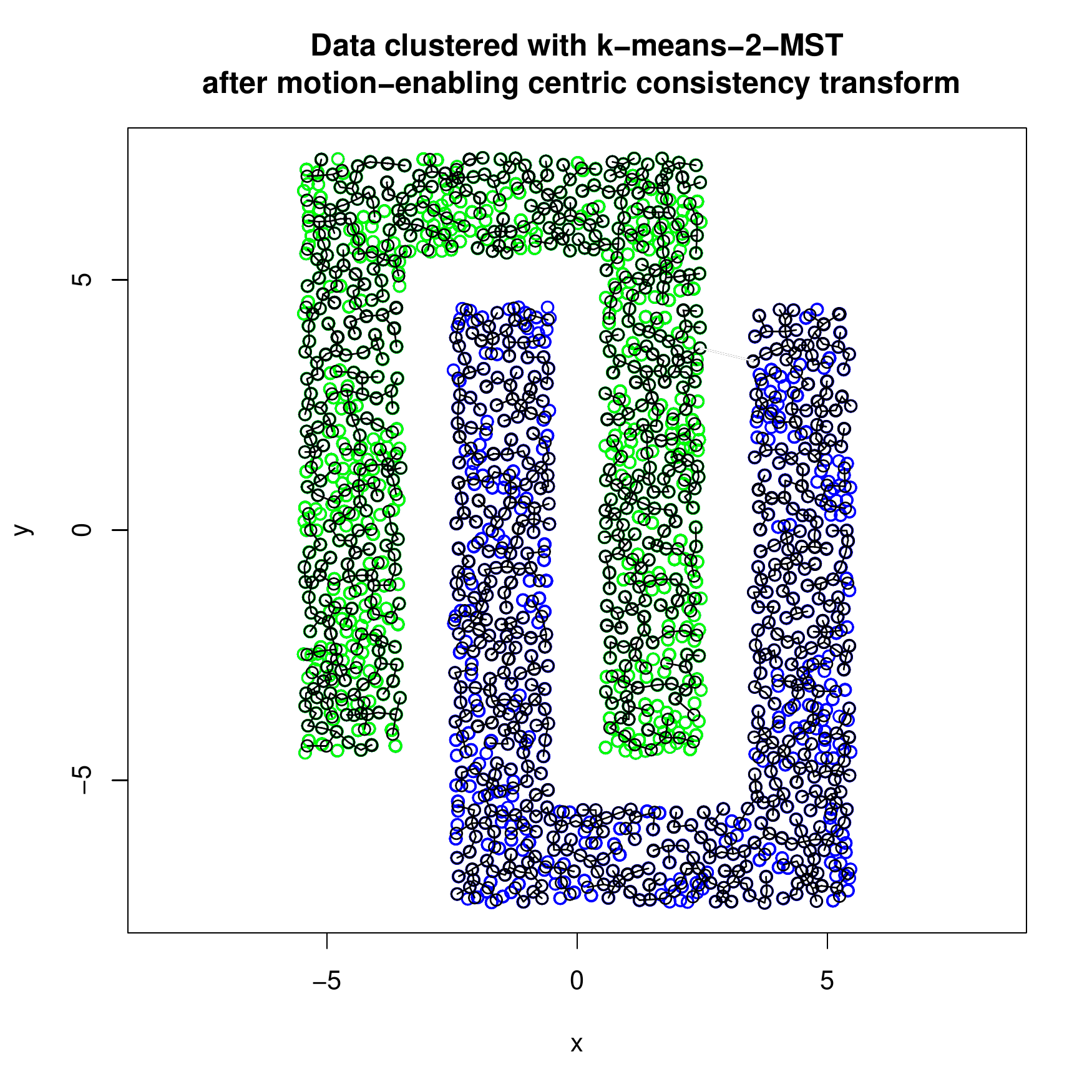}} \\ %
(e)
\includegraphics[width=0.4\textwidth]{\figaddr{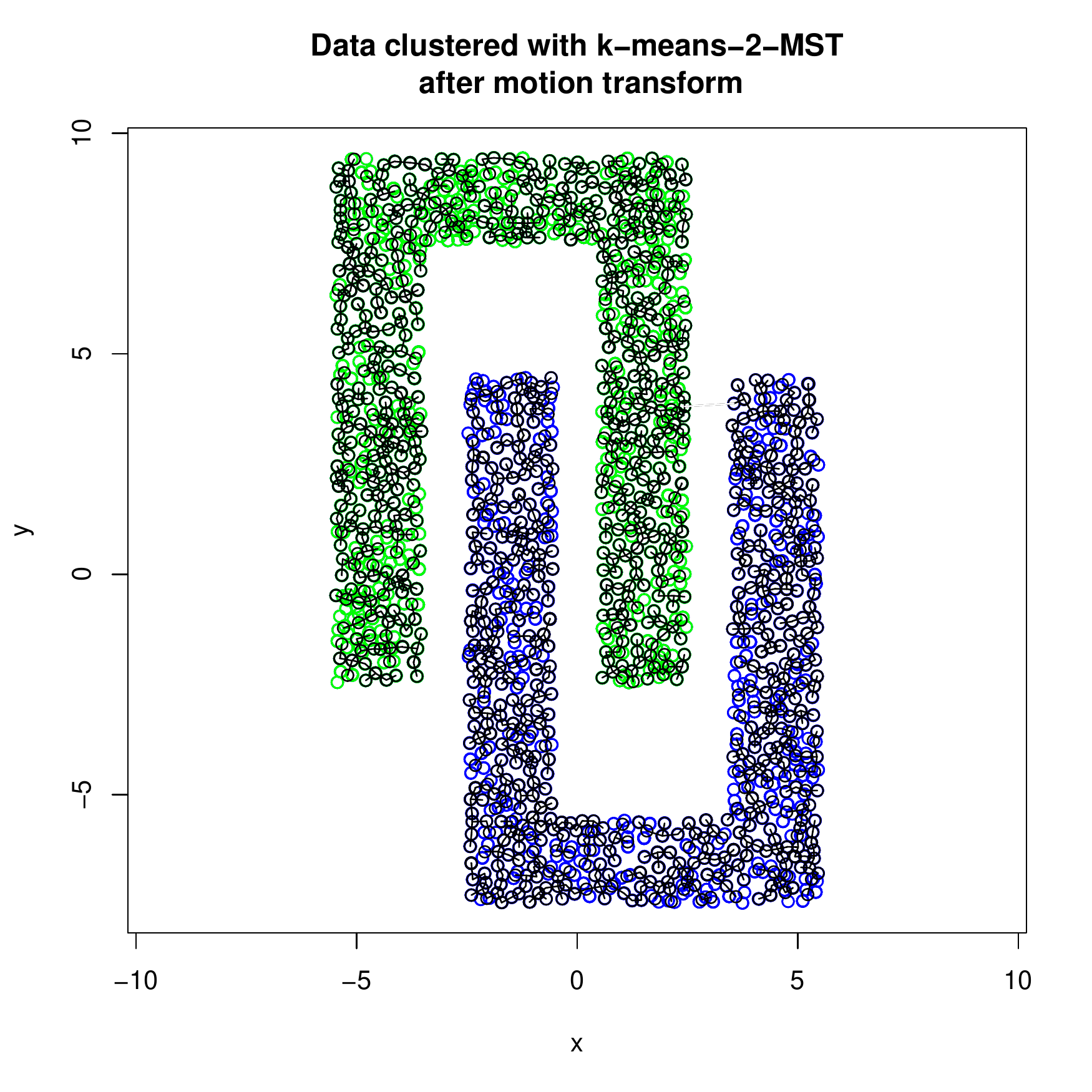}} %
(f)
\includegraphics[width=0.4\textwidth]{\figaddr{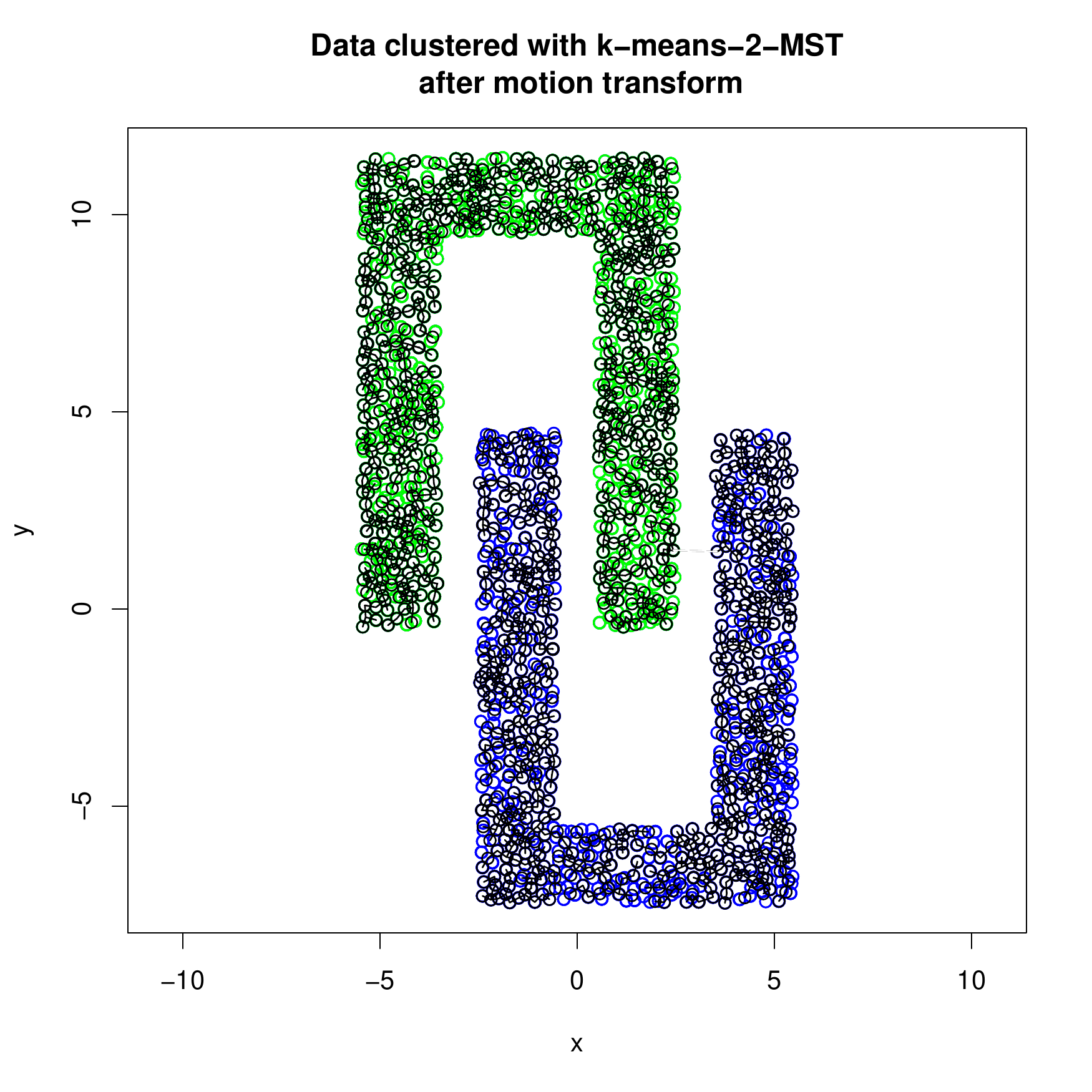}}  %
\end{center}%
\caption{Illustration of continuous motion consistency for the $k$-means-$2$-MST clustering algorithm.  The presented sample consists of 11200  data points, $k$=1500.  
(a) original data, (b) clustering using $k$-means-2-MST, spanning trees shown (c) clustering from (b) subjected to centric $\Gamma$-transform with $\lambda=0.7$, re{}clustered,
(d) clustering from (b) subjected to centric $\Gamma$-transform with $\lambda=0.052$, enabling motion with motion consistency, re{}clustered, 
(e) clustering from (d) subjected to motion (perpendicular vector with length 2, re{}clustered after motion, 
(f) clustering from (d) subjected to motion (perpendicular vector with length 4, re{}clustered after motion, 
}\label{fig:MotionConsistency}
\end{figure}

\begin{proof}
of Theorem \ref{thm:ClassFGeneralMotionConsistency}\label{thm:ClassFGeneralMotionConsistencyPROOF}.
 The proof follows the general idea behind the case of $k$-means. 
 If the solution $\Gamma_o$ is optimal, then the operation performed will induce no 
 change in the distances within the cluster and the distances between elements from different clusters will increase and this will cause an increase in the quality of the clustering function. 
 For any alternative clustering, that is a non-optimal one $\Gamma_n$,
 we can divide the distances between 
 data points in the following four categories:
 $K_1$ - intra-cluster distances both under $\Gamma_o$ and $\Gamma_n$, 
 $K_2$ - intra-cluster distances   under $\Gamma_o$ and inter-cluster distances under $\Gamma_n$, 
 $K_3$ - inter-cluster distances   under $\Gamma_o$ and intra-cluster distances under $\Gamma_n$, 
 $K_4$ - inter-cluster distances both under $\Gamma_o$ and $\Gamma_n$.
 Only $K_3$  is interesting for our considerations, as neither $K_1$ nor $K_2$ nor $K_4$ change the quality function.  
 Under $K_2$, the quality of $\Gamma_o$ does not change, while that of $\Gamma_n$ may decrease. This completes the proof. 
 \end{proof}
\section{Experiments}\label{sec:experRealData}

In order to validate the clustering preservation by the consistency transformations and motion transformations proposed in this paper also in cases when Kleinberg's consistency violation occurs, we investigated application of $k$-means to the data{}sets described in  table \ref{tab:info}, and applying to hem transformations from theorems 
\ref{thm:globalCentricCinsistencyForKmeans} and 
\ref{thm:GeneralMotionConsistency}. 

\begin{table} 
\caption{Data sets information}
\label{tab:info}
\begin{tabular}{|r|r|r|r|r|r|}
\hline Data{}set name & recs& cols& clusters& NSTART& time.sec\\
\hline  MopsiLocations2012-Joensuu.txt& 6014& 2& 5& 1000& 53\\
\hline  MopsiLocationsUntil2012-Finland.txt& 13467& 2& 5& 1000& 80\\
\hline  t4.8k ConfLong Demo.txt& 8000& 2& 5& 1000& 68\\
\hline  dim032.txt& 1024& 32& 16& 1000& 243\\
\hline  ConfLongDemo\_JSI\_164860.txt& 164860& 3& 5& 1000& 2088\\
\hline  dim064.txt& 1024& 64& 16& 977& 472\\
\hline  KDDCUP04Bio.txt& 145751& 74& 8& 54& 6119\\
\hline  kddcup99\_csv.csv& 494020& 5& 10& 640& 3488\\
\hline
\end{tabular}

\end{table} 

In Table \ref{tab:info}, the column \emph{recs}  contains the number of data records,  \emph{cols} - the number of columns used n the experiments (numeric columns only),  \emph{clusters} - the number of clusters into which the data was clustered in the experiments, \emph{NSTART} the number of restarts used for $k$-means algorithm (R implementation), \emph{time.sec} is the execution time of the experiment (encompassing not only clustering, but also time for computing various quality measures). 

The data{}sets were downloaded from (or from links from) the Web page
\url{http://cs.joensuu.fi/sipu/datasets/}.

\subsection{Experiments Related to Theorem \ref{thm:globalCentricCinsistencyForKmeans}}\label{sec:exp:globalCentricCinsistencyForKmeans}

The experiments were performed in the following manner: 
For each data{}set, the $k$-means with $k=clusters$ was performed with the number of restarts equal to $NSTART$, as mentioned in Table \ref{tab:info}. 
The result served as a "golden standard" that is the "true" clustering via $k$-means. 

Then the cluster was selected that was lying "most centrally" among all the clusters, so that the violation of Kleinberg's consistency upon centric consistency transformation would have the biggest effect.
Thereafter four experiments of centric consistency transformation on that cluster were run 
with $\lambda=0.8, 0.6, 0.4, 0.2$ resp. 
Then the resulting set was re-clustered using same $k$-means parameters. 
The number of disagreements in cluster membership is reported in Table \ref{tab:erclu}. As visible, only \url{KDDCUP04Bio.txt} exhibits problems, due to the fact that the number of restarts was kept low in order to handle the large number of records.

\begin{table} 
\caption{Clustering errors upon centric consistency transformation - see text for details }
\label{tab:erclu}
\begin{tabular}{|r|r|r|r|r|}
\hline  $\lambda=$ & 0.8& 0.6& 0.4& 0.2\\
\hline  MopsiLocations2012-Joensuu.txt& 0& 0& 0& 0\\
\hline  MopsiLocationsUntil2012-Finland.txt& 0& 0& 0& 0\\
\hline  t4.8k ConfLong Demo.txt& 0& 0& 0& 0\\
\hline  dim032.txt& 2& 2& 3& 1\\
\hline  ConfLongDemo\_JSI\_164860.txt& 0& 0& 0& 0\\
\hline  dim064.txt& 0& 0& 1& 0\\
\hline  KDDCUP04Bio.txt& 610& 610& 0& 610\\
\hline  kddcup99\_csv.csv& 0& 0& 0& 0\\
\hline
\end{tabular}

\end{table} 

In order to demonstrate that the centric consistency allows for cluster preserving transformation in spite of Kleinberg consistency conditino violations, the following statistics was performed: 100 data points were randomly picked from the "central" cluster (if there were fewer of them, all were taken) and 100 data points from other clusters were randomly picked. Then the distances prior and after the centric consistency transformations were computed between each element of the first and the second set and compared. The percentage of cases, when the Kleinberg consistency condition was violated, was computed (for each aforementioned $\lambda$)
The results are presented in Table \ref{tab:errpc}.

\begin{table} 
\caption{Kleinberg consistency violation percentage}
\label{tab:errpc}
\begin{tabular}{|r|r|r|r|r|}
\hline $\lambda=$ & 0.8& 0.6& 0.4& 0.2\\
\hline  MopsiLocations2012-Joensuu.txt& 1.32& 3.16& 54.6& 65.85\\
\hline  MopsiLocationsUntil2012-Finland.txt& 18.75& 1.965& 4& 0\\
\hline  t4.8k ConfLong Demo.txt& 17.59& 57.18& 56.67& 53.66\\
\hline  dim032.txt& 0& 0& 0.49& 0\\
\hline  ConfLongDemo\_JSI\_164860.txt& 72.55& 68.98& 19.22& 8.73\\
\hline  dim064.txt& 6& 4& 1& 6\\
\hline  KDDCUP04Bio.txt& 1& 0& 0& 2\\
\hline  kddcup99\_csv.csv& 3.62& 3.02& 3.12& 60.74\\
\hline
\end{tabular}

\end{table} 

As visible, as many as $70\%$ of distances may violate Kleinberg's condition. The seriousness of the problem depends both on the data{}set and the $\lambda$ considered. 
 
 The conclusion is that the centric consistency transformation provides with derived labelled datasets for investigation of $k$-means like algorithms even in cases when Kleinberg axiomatic system is violated.

\subsection{Experiments Related to Theorem \ref{thm:GeneralMotionConsistency}}\label{sec:exp:GeneralMotionConsistency}

The experiments were performed in the following manner: 
For each dataset, the $k$-means with $k=clusters$ was performed with the number of restarts equal to $NSTART$, as mentioned in Table \ref{tab:info}. As previously, the result served as a "golden standard" that is the "true" clustering via $k$-means. 

Then the motion experiment was initiated. 
Theorem \ref{thm:GeneralMotionConsistency} indicates that a "jump" of the clusters would have to be performed, but we insisted on "continuous" transformation. 
So first a centric consistency transformation was performed on each cluster so that the radii of all balls enclosing clusters are equal (the the radius of the smallest ball). Then centric consistency transformation was applied to each cluster so that the conditions of the Theorem \ref{thm:GeneralMotionConsistency} hold. 
After these two transformations, the correctness of clustering and the percentage of Kleinberg consistency violations was computed following the guidelines of the previous subsection. The results are reported in columns   $center.clu$ and $center.dst$ of Table \ref{tab:errmotionpc} resp. 

Afterwards twenty times  motion by a small step (randomly picking the length of the step) in random direction was performed. It was checked if conditions of Theorem  \ref{thm:GeneralMotionConsistency}  hold. If so, the correctness of clustering and the percentage of Kleinberg consistency violations was computed following the guidelines of the previous subsection. The maximal values of both are reported in columns  
  $motion.clu$ and $motion.dst$ of Table \ref{tab:errmotionpc} resp. with respect to the clustering after the two centric consistency transformations and in columns $motion.clu$ and $motion.dst$ of Table \ref{tab:errmotionpc} resp. with respect to the clustering in the previous step.

\begin{table} 
\caption{Kleinberg consistency distance errors percentage upon motion}
\label{tab:errmotionpc}
{\tiny
\begin{tabular}{|r|r|r|r|r|r|r|r|}
\hline  & center.dst& center.clu& motion.dst& motion.clu& step.dst& step.clu& time\\
\hline  MopsiLocations2012-Joensuu.txt& 65.92& 0& 74.23& 0& 98.08& 0& 384.0906\\
\hline  MopsiLocationsUntil2012-Finland.txt& 60.59& 0& 60.86& 0& 97& 0 &  392\\
\hline  t4.8k ConfLong Demo.txt& 61.07& 0& 64.46& 0& 100& 0&408\\
\hline  dim032.txt& 64.12& 4& 68.15& 5& 92& 6& 2215.708\\
\hline  ConfLongDemo\_JSI\_164860.txt& 57.22& 0& 72.05& 0& 100& 0& 498.239\\
\hline  dim064.txt& 58.89& 2& 64.46& 5& 100& 8& 4473.421\\
\hline  KDDCUP04Bio.txt& 68.41& 155& 80.93& 5393& 98& 16679& 6737.905\\
\hline  kddcup99\_csv.csv& 52.87& 0& 65.78& 0& 95& 0& 1084.221\\
\hline
\end{tabular}
}
\end{table} 

Again, the data{}set  \url{KDDCUP04Bio.txt} proves to be most problematic, due to the low  number of restarts with $k$-means clustering, even 16,000 violations of cluster membership were observed. 
Note however that this data{}set is huge compared to the other ones 
14,5751 data{}points in over 70 dimensions. Such data are problematic for the $k$-means itself (difficulties in proper seeding and ten recovery from erroneous seeding). There were also minor problems with the artificial benchmark 
data{}sets dim064.txt and dim032.txt. Otherwise the even non-optimal $k$-means algorithm approximated the true optimum sufficiently well to show that the motion consistency transformation really preserves the clustering. 

At the same time it is visible that 
 the motion consistency transformation provides with derived labelled data{}sets for investigation of $k$-means like algorithms even in cases when Kleinberg axiomatic system is seriously violated.

\section{Discussion}\label{sec:discussion}

The attempts to axiomatize the domain of clustering have a rich history. 
Van Laarhoven and Marchiori
\cite{vanLaarhoven:2014} and Ben-David and Ackerman  \cite{Ben-David:2009}
 the clustering axiomatic frameworks address either:
(1) required properties of  clustering  functions, or
(2) 
required properties of the   values of a clustering quality function,
or
(3) 
required properties of the relation between qualities of different partitions.

The work of Kleinberg\cite{Kleinberg:2002}, which we discussed extensively, fits into the first category. 

  Van Laarhoven and Marchiori \cite{vanLaarhoven:2014}
  developed an axiomatic system for graphs, fitting the same category. Though it is possible to view the data{}points in $\mathbb{R}^m$ as nodes of a graph connected by edges with appropriate weights, their axioms essentially reflect the axioms of Kleinberg with all the problems for $k$-means like algorithms.  

To overcome the problems with the consistency property, 
Ackerman et al.  \cite{Ackerman:2010NIPS}  
 propose 
splitting of it into 
the concept of \emph{outer-consistency} and of \emph{inner-consistency}.
However, 
the $k$-means algorithm is  not inner-consistent, see \cite[Section 3.1]{Ackerman:2010NIPS}. 
We show that the problem with inner-consistency is much deeper, as  
it may be not applicable  at all Theorem    \ref{lem:noinnerconsistency2dim4clusters}  and Theorem \ref{thm:noinnerConsistency}).

The $k$-means algorithm is said to be in this sense outer-consistent \cite[Section 5.2]{Ackerman:2010NIPS}.
But we have demonstrated in this paper that outer-consistency constitutes a problem in the Euclidean space, if continuous transformations are required, see Theorem \ref{thm:noGeneralConvexOuterConsistency}.  %
They claim also that  
$k$
-means-ideal has the properties of outer-consistency and locality\footnote{
A clustering function clustering into $k$ clusters has the locality property, 
if whenever a set $S$ for a given $k$ is clustered by it into the partition   $\Gamma$,
and we take a subset $\Gamma'\subset \Gamma$ with $|\Gamma'|=k'<k$,
then clustering of $\cup_{C\in \Gamma'}$ into $k'$ clusters will yield exactly $\Gamma'$. 
}. 
The property of locality would be useful for generation of testing sets for clustering function implementation. 
Ackerman at el \cite{Ackerman:2010NIPS} show that these properties are   satisfied neither by 
  $k$-means-random nor by  a $k$-means with furthest element initialization.

The mentioned properties cannot be fulfilled by any algorithm under fixed-dimensional settings.
 So they do so   for $k$-means-ideal.  

Zadeh   Ben-David \cite{Zadeh:2009} propose  instead the 
\emph{order-consistency}, satisfied by  some versions of single-linkage algorithm, providing with an elegant way of creating multitude of derived test sets.  
$k$-means is not order-consistent so that the property is not useful for continuous test data transformations for $k$-means.

Ackerman and Ben-David
\cite{Ben-David:2009} pursue the second category of approaches to axiomatization. 
Instead of  axiomatizaing  the clustering function, they create axioms for cluster quality function. 

\begin{definition} 
Let $\mathfrak{C}(\mathbf{X})$ be the set of all possible partitions 
over the set of objects $\mathbf{X}$,
and let $\mathfrak{D}(\mathbf{X})$ be the set of all possible distance functions  
over the set of objects $\mathbf{X}$.
A clustering-quality measure (CQM) $J: 
 \mathbf{X} \times \mathfrak{C}(\mathbf{X}) 
\times \mathfrak{D}(\mathbf{X})  \rightarrow  \mathbb{R}^+\cup\{0\}$
  is a function that, given a data set (with a  distance function) and its partition
into clusters, returns a non-negative real number representing how strong or
conclusive the clustering is. 
\end{definition} 

Ackerman and Ben-David propose among others  the following axioms:

\begin{ax}\label{ax:CQMscaleinvariance} (CQM-scale-invariance) A quality measure $J$ satisfies scale invariance if for every clustering
$\Gamma$ of $(X, d)$   and every positive $\beta$, $J(X,\Gamma, d) = J( X,\Gamma, \beta d)$.
\end{ax}

\begin{ax}\label{ax:CQMconsistency}
  (CQM-consistency) A quality measure $J$ satisfies consistency if for every clustering $\mathcal{C}$ over
$(X, d)$, whenever $d'$ is  consistency-transformation   of $d$, then $J(X,\Gamma, d') \ge J(X,\Gamma, d)$.
\end{ax}

If we define a clustering function $f$ to maximize the quality function, then 
the clustering function is (scale)-invariant, but 
the clustering function \emph{does not need to   be consistent} in Kleinberg's sense. 
E.g.  $k$-means is CQM-consistent, not being consistent in Kleinberg's sense. . 

The basic problem with this axiomatic set is that the CQM-consistency does  not tell anything about the (optimal) partition being the result of the consistency-transform, while Kleinberg's axioms make a definitive statement: the partition before and after consistency-transform has to be the same. 
So $k$-means could be in particular rendered to become 
  CQM-consistent, CQM-scale-invariant, and CQM-rich, if one applies 
a bi-sectional version
(bisectional-$auto$-means).    

A number of further characterizations of clustering functions has been proposed  
e.g. Ackerman's at al.  \cite{Ackerman:2010} for linkage algorithms, 
Carlsson's \cite{Carlsson:2008} for multiscale clustering.
None of the transforms proposed there seems however to fit the purpose of test set derivation for $k$-means testing by a continuous derivation. 

An interesting axiomatic system for hierarchical clustering was developed in \cite{Thomann:2015}. It is defined not for data{}points but for probability distributions over some support(s). It includes a scale invariance property, but provides nothing that can be considered as corresponding to Kleinberg's consistency. This clearly limits the usability for test set generation for (hierarchical) clustering algorithms. Nonetheless this approach is worth noting and worth pursuing for non-hierarchical clustering functions like $k$-means for two reasons: (1) $k$-means not only splits the data themselves but also the Euclidean spce as such, (2) $k$-means is probabilistic in nature and rather than exact re-clustering of data one should pay attention to recovery of appropriately defined probability distributions over the space (3) the concepts of centric consistency and that of motion consistency are applicable to probability distributions over finite support. So it may constitute a research path worthy further exploration.


Puzicha et al. \cite{Puzicha:2000} propose an axiomatic system in which the data can be transformed by shifting the entire data set (and not the individual clusters). In our approach, individual clusters may be subject to shifts. 

Strazzeri et al. \cite{Strazzeri:2021} introduced the axiom of monotonic consistency to replace Kleinberg's consistency axiom. He requires that, for an expansion function (monotonically increasing)  $\nu: : [0;\infty1) \rightarrow  [0;\infty)$,
each intercluster distance $d$ is changed to $d'=\nu(d)$ and each intracluster distance  $d$ is changed to $d'=\nu^{-1}(d)$. 
This transform, though it is non-linear on the one hand, is more rigid than our centric consistency and motion consistency transforms, as it has to apply globally, and not locally to a cluster, and on the other hand, it is designed and applicable for graphs, but not for $\mathbb{R}^m$ except for special cases.


Addad \cite{Addad:2018} proposed the following  modification of the Kleinberg's consistency axiom. 
Let $OPT(k)$ be the optimal value of a cost function when clustering into $k$ parts. Let $k^*$ the $k$ maximizing  quotient $OPT(k)/OPT(k-1)$ 
Then their \emph{refined consistency} requires that the Kleinberg's consistency transform yields the same clustering only if $k^*$ prior and after the transform agree. 
They prove that single link and $k$-means fulfil the refined consistency axiom (the latter under some "balance" requirements). 
While the refined consistency transform (identical to Kleinberg's consistency transform) is much more flexible in generating derived test sets, it is in practice hard to test whether or not a given transform produces identically cluster{}able data{}sets. 

Ackerman et al. \cite{Ackerman:2021} consider a different way of deriving new samples from old ones, re-weighting of data{}points. It turns out that some algorithms are sensitive to re-weighting, other are robust. Regrettably, $k$-means is sensitive so that the method cannot be applied to derive new clustering  test data{}sets from existent ones.

A discussion and critics of various approaches to axiomatic systems for clustering can be found in \cite{Hennig:2015}.


\section{Conclusions}

We showed that the consistency axiom of Kleinberg \cite{Kleinberg:2002} constitutes a problem: 
neither consistency, nor inner-consistency nor outer-consistency can be executed continuously in Euclidean space of fixed dimension. 
Hence the application for generating of new labelled data from existent ones for clustering algorithm testing cannot be based on the respective transformations. 
\NSTUFF{
Hence we proposed alternative axioms suitable for continuous transformations in Euclidean space matching the spirit of Klein{}berg's consistency axioms but free of their induced contradictions. 
The respective transformations can therefore be applied for the mentioned test data generation. }

\NSTUFF{
This research was restricted to embedding{}s in Euclidean space. In a separate study \cite{RAKMAKSTW:2020:trick}  we showed that kernel-$k$-means algorithm under non-euclidean distances between data points may be deemed as working under Euclidean space after adding a specific constant to distance squares. Hence, for kernel-$k$-means testing still another way of new test sample generation is possible, that is via adding or subtracting constants from squared distances between data elements. 
}

\vskip 0.2in
\bibliographystyle{plain}
\bibliography{bib_Kleinberg_related_RAK_MAK_bib,bib_generalRAKMAK_bib}


\end{document}